\RequirePackage{fix-cm}

\documentclass[twocolumn]{svjour3}
\smartqed

\usepackage{tikz}
\usetikzlibrary{spy}
\usepackage{subcaption}
\usepackage{graphicx}
\usepackage{epstopdf}
\usepackage{amsmath}
\usepackage{amssymb}
\usepackage{placeins}
\usepackage{booktabs}
\usepackage{xifthen}
\usepackage[pagebackref=true,breaklinks=true,letterpaper=true,colorlinks,bookmarks=false]{hyperref}
\usepackage{url}
\usepackage{cleveref}
\usepackage{adjustbox}
\usepackage{xstring}
\usepackage{bm}
\usepackage[space]{grffile}

\makeatletter
\def\BState{\State\hskip-\ALG@thistlm}
\makeatother

\newcommand{\feat}{\Phi}
\newcommand{\deflen}[2]{%
    \expandafter\newlength\csname #1\endcsname
    \expandafter\setlength\csname #1\endcsname{#2}%
}

\newcommand\makespy[1]{}

\usetikzlibrary{calc}
\newcommand*{\ExtractCoordinate}[1]{\path (#1); \pgfgetlastxy{\XCoord}{\YCoord}; }
\newcommand*{\ExtractImgDims}[1]{
    \ExtractCoordinate{$(#1.south west)$};
    \pgfmathsetmacro{\imgx}{\XCoord}
    \pgfmathsetmacro{\imgy}{\YCoord}
    \ExtractCoordinate{$(#1.north east)$};
    \pgfmathsetmacro{\imgw}{\XCoord - \imgx}
    \pgfmathsetmacro{\imgh}{\YCoord - \imgy}
}
\newcommand*{\RelativeSpy}[5]{
    \ExtractImgDims{#2};
    \begin{scope}[x=\imgw,y=\imgh,xshift=\imgx,yshift=\imgy]
        \coordinate (spyroi-#1) at #3;
        \coordinate (spypos-#1) at #4;
        \spy[anchor=center,color=#5] on (spyroi-#1) in node[anchor=center] at (spypos-#1);
    \end{scope}
}

\newcommand\blfootnote[1]{%
  \begingroup
  \renewcommand\thefootnote{}\footnote{#1}%
  \addtocounter{footnote}{-1}%
  \endgroup
}

\journalname{IJCV}
\begin{document}

\title{Deep Image Prior}

\author{
Dmitry Ulyanov \and
Andrea Vedaldi \and
Victor Lempitsky
}


\institute{
D. Ulyanov \at
Skolkovo Institute of Science and Technology \\
\email{dmitry.ulyanov@skoltech.ru}
\and
A. Vedaldi \at
Oxford University \\
\email{vedaldi@robots.ox.ac.uk}
\and
V. Lempitsky \at
Skolkovo Institute of Science and Technology \\
\email{lempitsky@skoltech.ru}
}

\vspace{-5mm}
\date{Accepted: 04 February 2020}

\maketitle

\hbadness=10000
\hfuzz=\maxdimen
\newcount\hbadness
\newdimen\hfuzz

\begin{abstract}
Deep convolutional networks have become a popular tool for image generation and restoration. Generally, their excellent performance is imputed to their ability to learn realistic image priors from a large number of example images. In this paper, we show that, on the contrary, the \emph{structure} of a generator network is sufficient to capture a great deal of low-level image statistics \emph{prior to any learning}. In order to do so, we show that a randomly-initialized neural network can be used as a handcrafted prior with excellent results in standard inverse problems such as denoising, super-resolution, and inpainting. Furthermore, the same prior can be used to invert deep neural representations to diagnose them, and to restore images based on flash-no flash input pairs.

Apart from its diverse applications, our approach highlights the inductive bias captured by standard generator network architectures. It also bridges the gap between two very popular families of image restoration methods: learning-based methods using deep convolutional networks and learning-free methods based on handcrafted image priors such as self-similarity.\blfootnote{Code and supplementary material are available at \url{https://dmitryulyanov.github.io/deep_image_prior}}
\end{abstract}


\begin{figure}
    \centering
    \deflen{twolensplash}{0.49\linewidth}
    \renewcommand\makespy[1]{%
        \begin{tikzpicture}[spy using outlines={rectangle,magnification=3, height=4.7cm, width=10.2cm, every spy on node/.append style={line width=2mm}}]
                \node (nd1){\includegraphics{#1}};
                \RelativeSpy{nd1-spy1}{nd1}{(0.24,0.595)}{(0.253,-0.18)}{red}
                \RelativeSpy{nd1-spy4}{nd1}{(0.87,0.494)}{(0.750,-0.18)}{blue}
        \end{tikzpicture}%
    }

    \begin{subfigure}[b]{\twolensplash}
        \resizebox{1.02\textwidth}{!}{
            \makespy{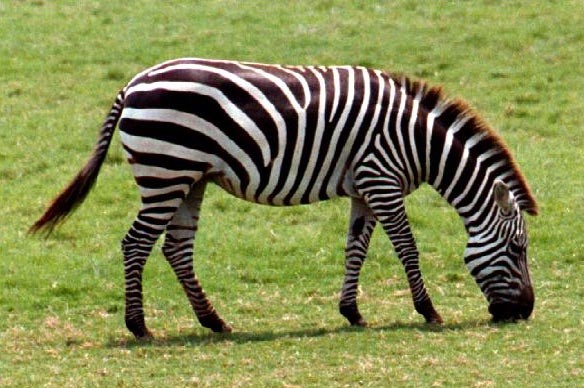}
        }
        \vspace*{-3mm}\caption{Ground truth}
    \end{subfigure}
    \begin{subfigure}[b]{\twolensplash}
        \resizebox{1.02\textwidth}{!}{
            \makespy{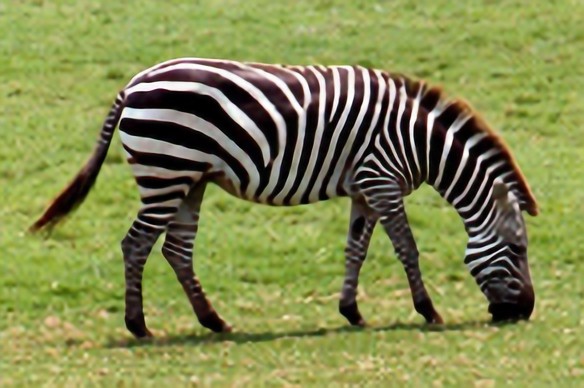}
        }
        \vspace*{-3mm}\caption{SRResNet~\cite{Ledig17sr}, \textbf{Trained}}
    \end{subfigure}
    \\ 
    \begin{subfigure}[b]{\twolensplash}
        \resizebox{1.02\textwidth}{!}{
            \makespy{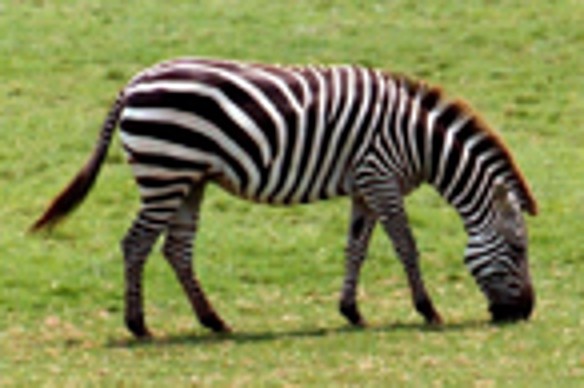}
        }
        \vspace*{-3mm}\caption{Bicubic, \textbf{Not trained}}
    \end{subfigure}
    \begin{subfigure}[b]{\twolensplash}
        \resizebox{1.02\textwidth}{!}{
            \makespy{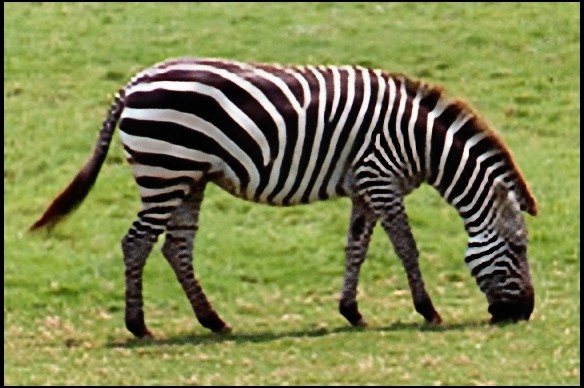}
        }
        \vspace*{-3mm}\caption{Deep prior, \textbf{Not trained}}
    \end{subfigure}
    \caption{\textbf{Super-resolution using the deep image prior.} Our method uses a randomly-initialized ConvNet to upsample an image, using its structure as an image prior; similar to bicubic upsampling, this method does not require learning, but produces much cleaner results with sharper edges. In fact, our results are quite close to state-of-the-art super-resolution methods that use ConvNets learned from large datasets. The deep image prior works well for all inverse problems we could test.}\label{fig:splash}
\end{figure}

\section{Introduction}\label{s:intro}

State-of-the-art approaches to image reconstruction problems such as denoising~\cite{burger2012image,lefkimmiatis2016non} and single-image super-resolution~\cite{Ledig17sr,Tai17sr,Lai17sr} are currently based on
deep convolutional neural networks (ConvNets).
ConvNets also work well in ``exotic'' inverse problems such as reconstructing an image from its activations within a deep network or from its HOG descriptor~\cite{dosovitskiy16inverting}.
Popular approaches for image generation such as
generative adversarial networks~\cite{goodfellow2014generative}, variational autoencoders~\cite{kingma2013auto} and direct pixel-wise error minimization~\cite{dosovitskiy2015learning,Bojanowski17} also use ConvNets.

ConvNets are generally trained on large datasets of images, so one might assume that their excellent performance is due to the fact that they learn realistic data priors from examples, but this explanation is insufficient.
For instance, the authors of~\cite{zhang16understanding} recently showed that the same image classification network that generalizes well when trained on a large image dataset can \emph{also} overfit the same images when labels are randomized.
Hence, it seems that obtaining a good performance also requires the \emph{structure} of the network to ``resonate'' with the structure of the data.
However, the nature of this interaction remains unclear, particularly in the context of image generation.

In this work, we show that, in fact, not all image priors must be learned from data; instead, a great deal of image statistics are captured by the \emph{structure} of generator ConvNets, independent of learning.
This is especially true for the statistics required to solve certain image restoration problems, where the image prior must supplement the information lost in the degradation processes.

To show this, we apply \textit{untrained} ConvNets to the solution of such problems (\cref{fig:splash}).
Instead of following the standard paradigm of training a ConvNet on a large dataset of example images, we fit a generator network to a single degraded image.
In this scheme, the network weights serve as a parametrization of the restored image.
The weights are randomly initialized and fitted to a specific degraded image under a task-dependent observation model.
In this manner, the only information used to perform reconstruction is contained in the single degraded input image \emph{and} the handcrafted structure of the network used for reconstruction.

We show that this very simple formulation is very competitive for standard image processing problems such as denoising, inpainting, super-resolution, and detail enhancement. This is particularly remarkable because \emph{no aspect of the network is learned from data} and illustrates the power of the image prior implicitly captured by the network structure.
To the best of our knowledge, this is the first study that directly investigates the prior captured by deep convolutional generative networks independently of learning the network parameters from images.

In addition to standard image restoration tasks, we show an application of our technique to understanding the information contained within the activations of deep neural networks trained for classification.
For this, we consider the ``natural pre-image'' technique of~\cite{mahendran15understanding}, whose goal is to characterize the invariants learned by a deep network by inverting it on the set of natural images.
We show that an untrained deep convolutional generator can be used to replace the surrogate natural prior used in~\cite{mahendran15understanding} (the TV norm) with dramatically improved results.
Since the new regularizer, like the TV norm, is not learned from data but is entirely handcrafted, the resulting visualizations avoid potential biases arising form the use of  learned regularizers~\cite{dosovitskiy16inverting}.
Likewise, we show that the same regularizer works well for ``activation maximization'', namely the problem of synthesizing images that highly activate a certain neuron~\cite{erhan09visualizing}.

\section{Method}\label{s:method}

\begin{figure*}
     \centering
     \includegraphics[width=1.0\linewidth]{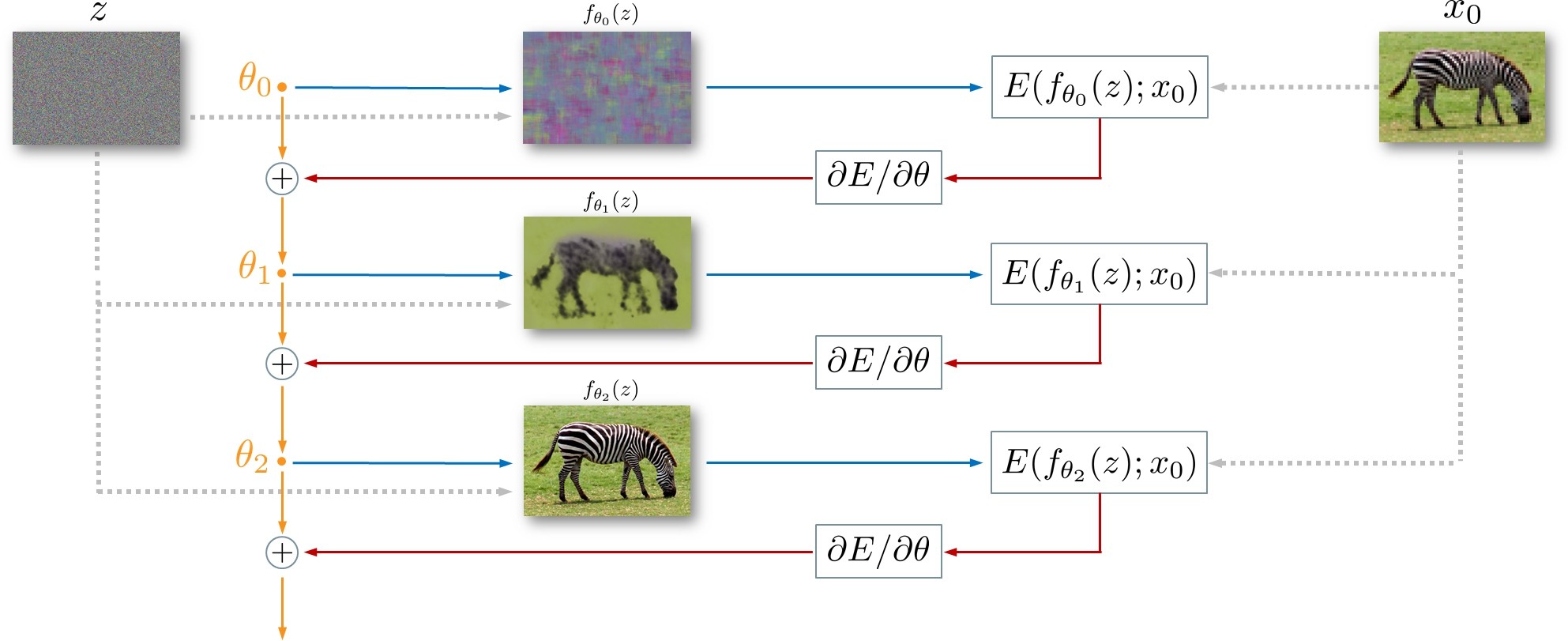}
     \caption{\textbf{Image restoration using the deep image prior.} Starting from a random weights $\theta_0$, we iteratively update them in order to minimize the data term~\cref{eq:reparametrization}. At every iteration $t$ the weights $\theta$ are mapped to an image $x = f_\theta(z)$, where $z$ is a fixed tensor and the mapping $f$ is a neural network with parameters $\theta$. The image $x$ is used to compute the task-dependent loss $E(x, x_0)$. The gradient of the loss w.r.t.~the weights $\theta$ is then computed and used to update the parameters.}\label{fig:pipeline}
\end{figure*}

 \begin{figure*}
    \centering
    \includegraphics[width=\linewidth]{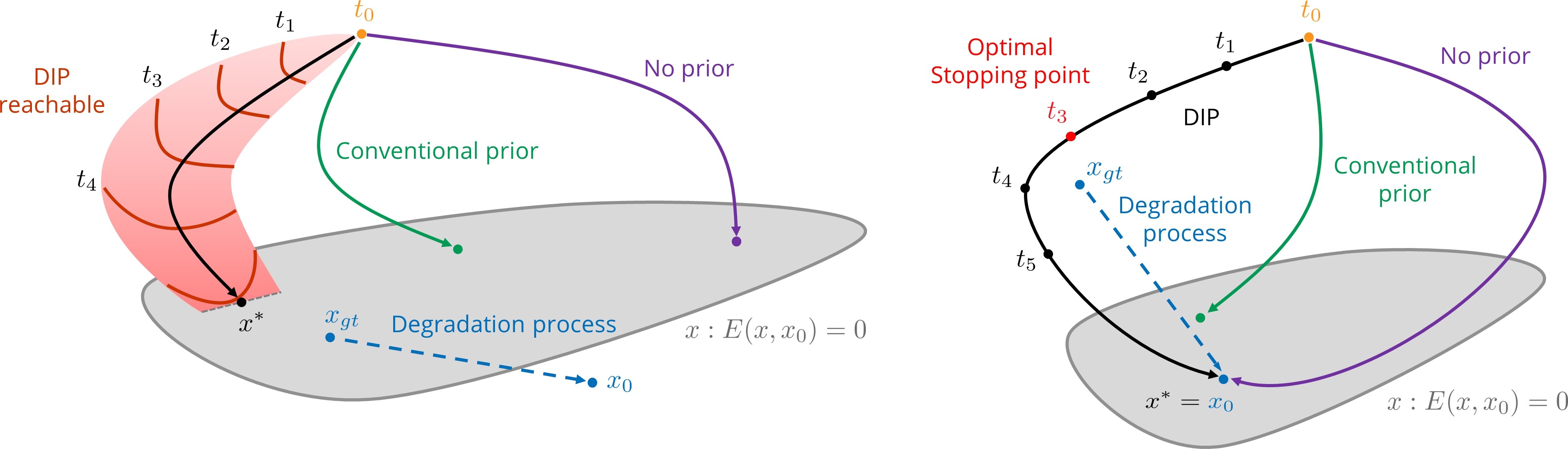}

\caption{\textbf{Restoration with priors --- image space visualization.} We consider the problem of reconstructing an image $x_{\text{gt}}$ from a degraded measurement $x_0$. We distinguish two cases. \textbf{Left} --- in the first case, exemplified by super-resolution, the ground-truth solution $x_{\text{gt}}$ belongs to a manifold of points $x$ that have null energy $x: E(x ,x_0)=0$ (shown in gray) and optimization can land on a point $x^*$ still quite far from $x_\text{gt}$ (purple curve).
Adding a conventional prior $R(x)$ tweaks the energy so that the optimizer $x^*$ is closer to the ground truth (green curve).
The deep image prior has a similar effect, but achieves it by tweaking the optimization trajectory  via re-parametrization, often with better results than conventional priors.
    \textbf{Right} --- in the second case, exemplified by denoising, the ground truth $x_{\text{gt}}$  has non-zero cost $E(x_{\text{gt}},x_0)>0$. Here, if run for long enough, fitting with deep image prior will obtain a solution with near zero cost quite far from $x_{\text{gt}}$. However, often the optimization path will pass close to $x_{\text{gt}}$, and an early stopping (here at time $t_3$) will recover good solution. Below, we show that deep image prior often helps for problems of both types.
    }\label{fig:priors_illustration}
\end{figure*}

A deep generator network is a parametric function $x=f_\theta(z)$ that maps a code vector $z$ to an image $x$.
Generators are often used to model a complex distribution $p(x)$ over images as the transformation of simple distribution $p(z)$ over the codes, such as a Gaussian distribution~\cite{goodfellow2014generative}.

One might think that knowledge about the distribution $p(x)$ is encoded in the parameters $\theta$ of the network, and is therefore learned from data by training the model.
Instead, we show here that a significant amount of information about the image distribution is contained in the \emph{structure} of the network even without performing any training of the model parameters.

We do so by interpreting the neural network as a \emph{parametrization} $x = f_\theta(z)$ of the image $x \in \mathbb{R}^{3\times H\times W}$.
In this view, the code is a fixed random tensor $z \in \mathbb{R}^{C'\times H'\times W'}$ and the network maps the parameters $\theta$, comprising the weights and bias of the filters in the network, to the image $x$.
The network itself has a standard structure and alternates filtering operations such as linear convolution, upsampling and non-linear activation functions.

\begin{figure*}
\centering
\begin{subfigure}[b]{0.49\linewidth}
    \centering
    \includegraphics[width=0.9\linewidth]{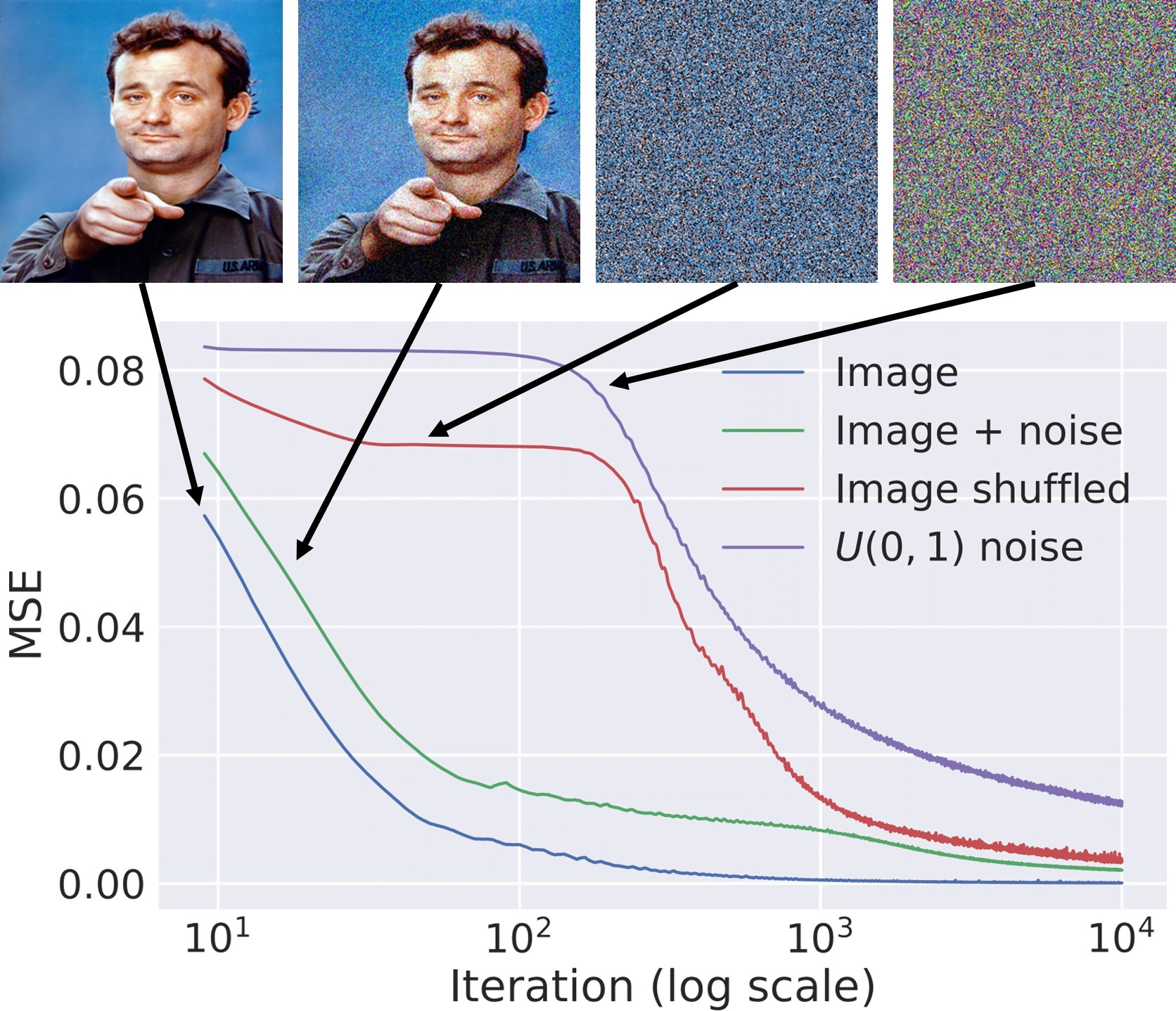}
\end{subfigure}
\begin{subfigure}[b]{0.49\linewidth}
    \centering
    \includegraphics[width=0.9\linewidth]{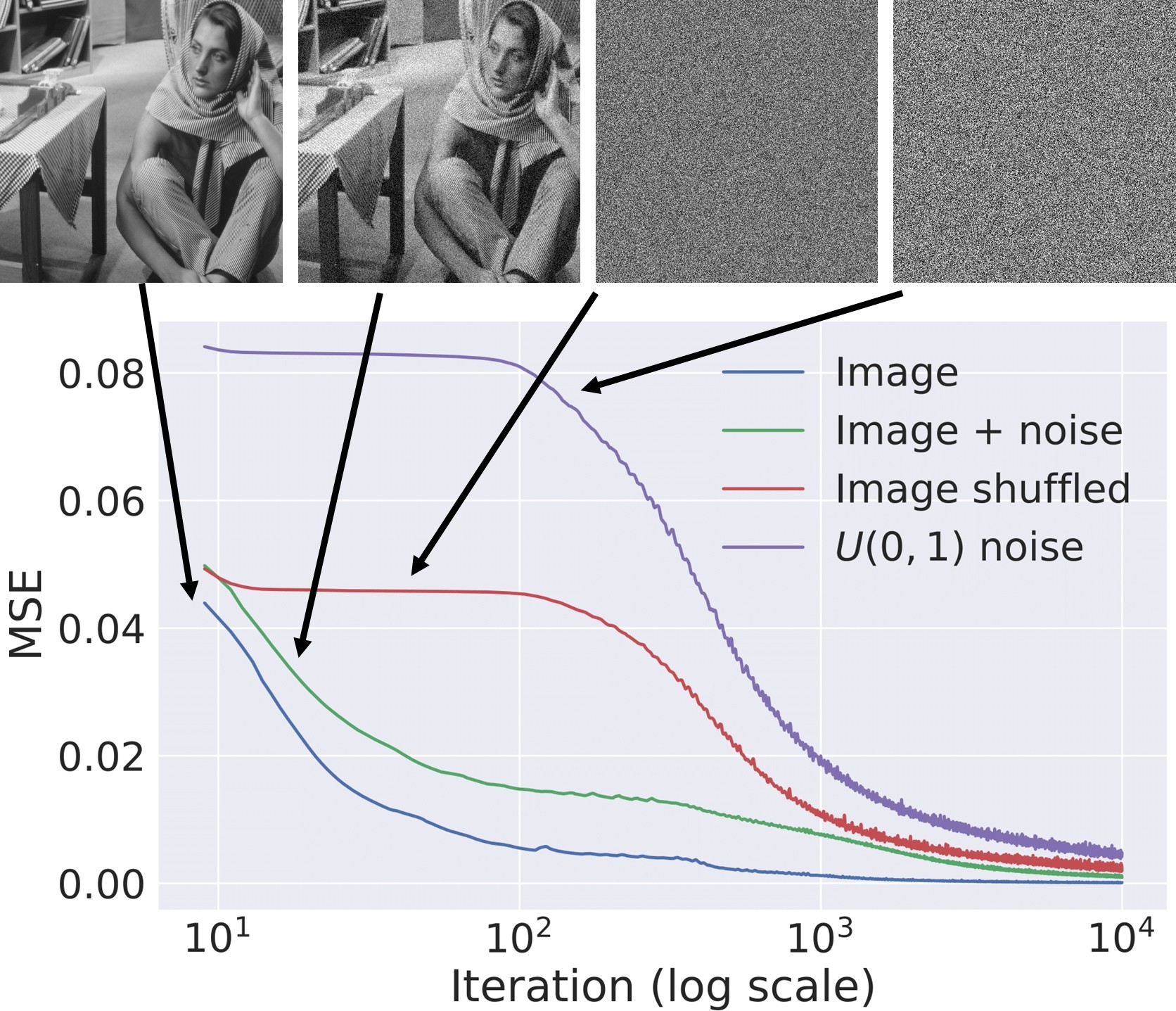}
\end{subfigure}

\caption{Learning curves for the reconstruction task using: a natural image, the same plus i.i.d.\ noise, the same randomly scrambled, and white noise. Naturally-looking images result in much faster convergence, whereas noise is rejected.}\label{fig:recon}
\end{figure*}

Without training on a dataset, we cannot expect the a network $f_\theta$ to know about specific concepts such as the appearance of certain objects classes.
However, we demonstrate that the untrained network does capture some of the \emph{low-level statistics} of natural images --- in particular, the local and translation invariant nature of convolutions and the usage of a sequence of such operators captures the relationship of pixel neighborhood at multiple scales.
This is sufficient for it to model \emph{conditional} image distributions $p(x|x_0)$ of the type that arise in image restoration problems, where $x$ has to be determined given a corrupted version $x_0$ of itself.
The latter can be used to solve inverse problems such as denoising~\cite{burger2012image}, super-resolution~\cite{dong2014learning} and inpainting.

Rather than working with distributions explicitly, we formulate such tasks as energy minimization problems of the type
\begin{equation}\label{eq:direct}
    x^* = \operatornamewithlimits{argmin}_x E(x;x_0) + R(x),
\end{equation}
where $E(x;x_0)$ is a task-dependent data term, $x_0$ is the noisy/low-resolution/occluded image, and $R(x)$ is a regularizer.

The choice of data term $E(x; x_0)$ is often directly dictated by the application and is thus not difficult.
The regularizer $R(x)$, on the other hand, is often not tied to a specific application because it captures the generic regularity of natural images.  A simple example is Total Variation (TV), which encourages images to contain uniform regions, but much research has gone into designing and learning good regularizers.

In this work, we drop the explicit regularizer $R(x)$ and use instead the implicit prior captured by the neural network parametrization, as follows:
\begin{equation}\label{eq:reparametrization}
   \theta^* = \operatornamewithlimits{argmin}_\theta E(f_\theta(z);x_0),\qquad x^* =f_{\theta^*}(z)\,.
\end{equation}
The (local) minimizer $\theta^*$ is obtained using an optimizer such as gradient descent, starting from a \emph{random initialization} of the parameters $\theta$ (see~\cref{fig:pipeline}).
Hence, the only empirical information available to the restoration process is the noisy image $x_0$.
Given the resulting (local) minimizer $\theta^*$, the result of the restoration process is obtained as $x^* = f_{\theta^*}(z)$.\footnote{\Cref{eq:reparametrization} can also be thought of as a regularizer $R(x)$ in the style of~\cref{eq:direct}, where $R(x)=0$ for all images that can be generated by a deep ConvNet of a certain architecture with the weights being not too far from random initialization, and $R(x)=+\infty$ for all other signals.} This approach is schematically depicted in~\cref{fig:priors_illustration} (left).

Since no aspect of the network $f_\theta$ is learned from data beforehand, such \textit{deep image prior} is effectively handcrafted, just like the TV norm. The contribution of the paper is to show that this hand-crafted prior works very well for various image restoration tasks, well beyond standard handcrafted priors, and approaching learning-based approaches in many cases.

As we show in the experiments, the choice of architecture does have an impact on the results.
In particular, most of our experiments are performed using a U-Net-like ``hourglass'' architecture with skip connections, where $z$ and $x$ have the same spatial dimensions and the network has several millions of parameters.
Furthermore, while it is also possible to optimize over the code $z$, in our experiments we do not do so. Thus, unless noted otherwise, $z$ is a fixed randomly-initialized $3D$ tensor.


\subsection{A parametrization with high noise impedance}\label{s:noise_impedance}

One may wonder why a high-capacity network $f_\theta$ can be used as a prior at all. In fact, one may expect to be able to find parameters $\theta$ recovering any possible image $x$, including random noise, so that the network should not impose any restriction on the generated image. We now show that, while indeed almost any image can be fitted by the model, the choice of network architecture has a major effect on how the solution space is searched by methods such as gradient descent. In particular, we show that the network  resists ``bad'' solutions and descends much more quickly towards naturally-looking images. The result is that minimizing~\eqref{eq:reparametrization} either results in a good-looking local optimum (\cref{fig:priors_illustration} --- left), or, at least, that the optimization trajectory passes near one (\cref{fig:priors_illustration} --- right).

In order to study this effect quantitatively, we consider the most basic reconstruction problem: given a target image $x_0$, we want to find the value of the parameters $\theta^*$ that reproduce that image. This can be setup as the optimization of~\eqref{eq:reparametrization} using a data term such as the $L^2$ distance that compares the generated image to $x_0$:
\begin{equation} \label{eq:denoise}
  E(x; x_0) = \| x -x_0 \|^2 \,.
\end{equation}

Plugging~\cref{eq:denoise} in~\cref{eq:reparametrization} leads us to the optimization problem
\begin{equation} \label{eq:denoise2}
  \min_\theta \| f_\theta(z) - x_0 \|^2 \,.
\end{equation}

\Cref{fig:recon} shows the value of the energy $E(x;x_0)$ as a function of the gradient descent iterations for four different choices for the image $x_0$: 1) a natural image, 2) the same image plus additive noise, 3) the same image after randomly permuting the pixels, and 4) white noise. It is apparent from the figure that the optimization is much faster for cases 1) and 2), whereas the parametrization presents significant ``inertia'' for cases 3) and 4).
Thus, although in the limit the parametrization \emph{can} fit noise as well, it does so very reluctantly. In other words, the parametrization offers high impedance to noise and low impedance to signal.

To use this fact in some of our applications, we restrict the number of iterations in the optimization process~\eqref{eq:reparametrization}. The resulting prior then corresponds to projection onto a reduced set of images that can be produced from $z$ by ConvNets with parameters $\theta$ that are not too far from the random initialization $\theta_0$. The use of deep image prior with the restriction on the number of iterations in the optimization process is schematically depicted in~\cref{fig:priors_illustration} (right).

\begin{figure*}
    \centering
    \begin{tabular}{ccccc}
    \includegraphics[width=3cm]{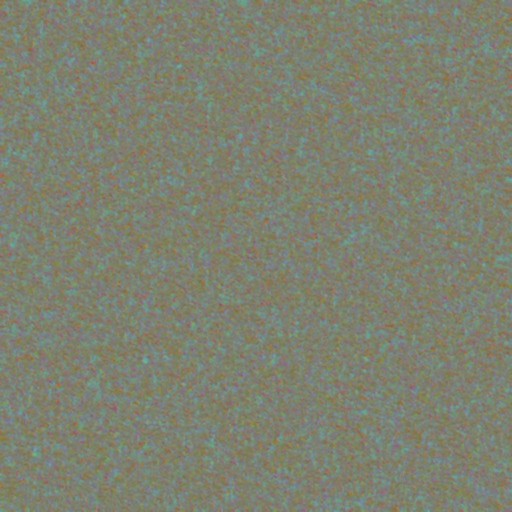}&
    \includegraphics[width=3cm]{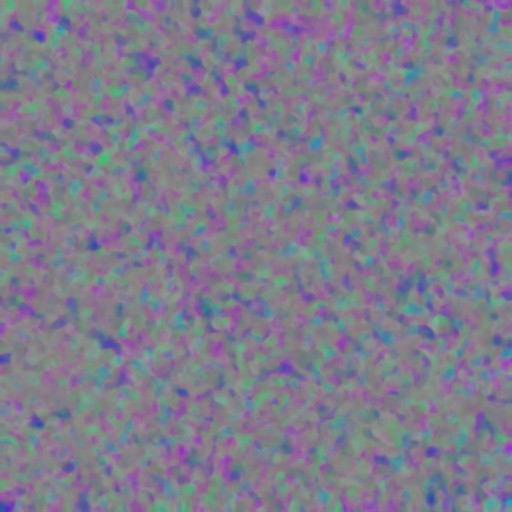}&
    \includegraphics[width=3cm]{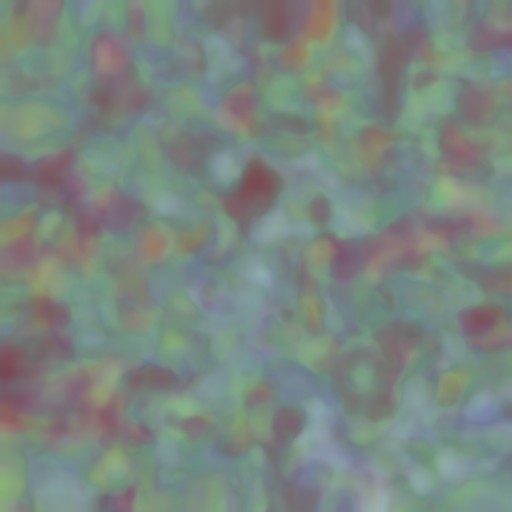}&
    \includegraphics[width=3cm]{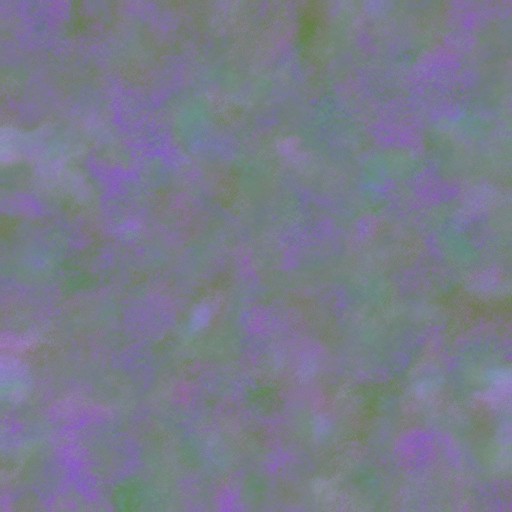}&
    \includegraphics[width=3cm]{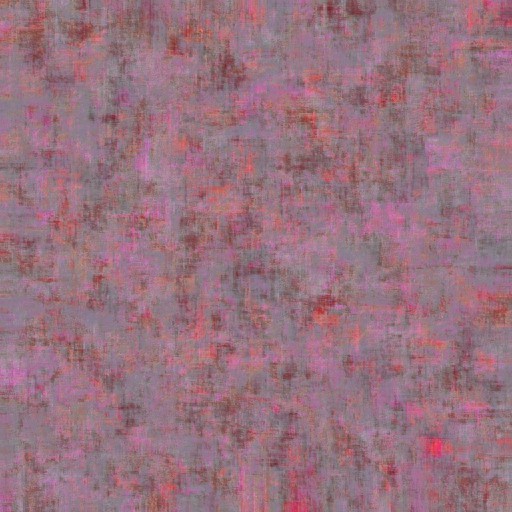}\\
    \includegraphics[width=3cm]{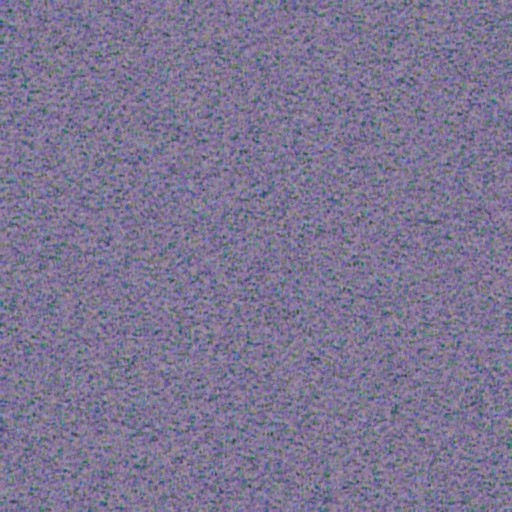}&
    \includegraphics[width=3cm]{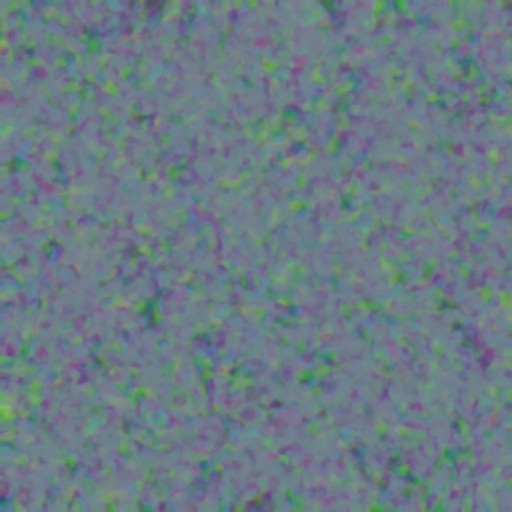}&
    \includegraphics[width=3cm]{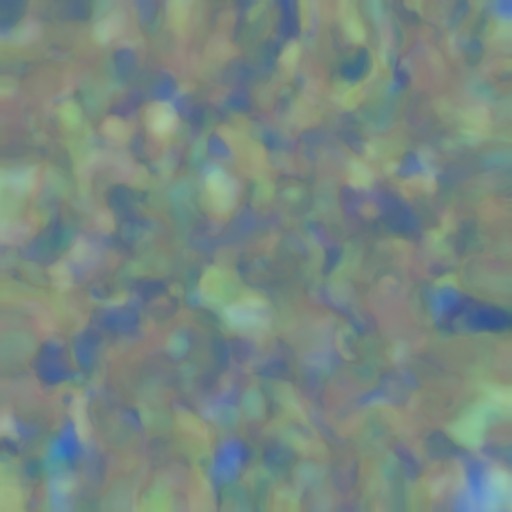}&
    \includegraphics[width=3cm]{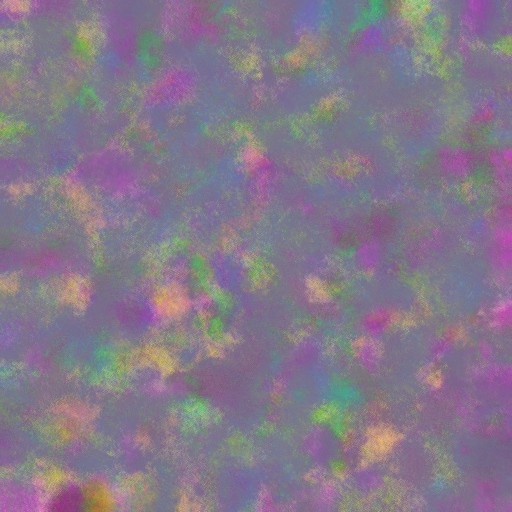}&
    \includegraphics[width=3cm]{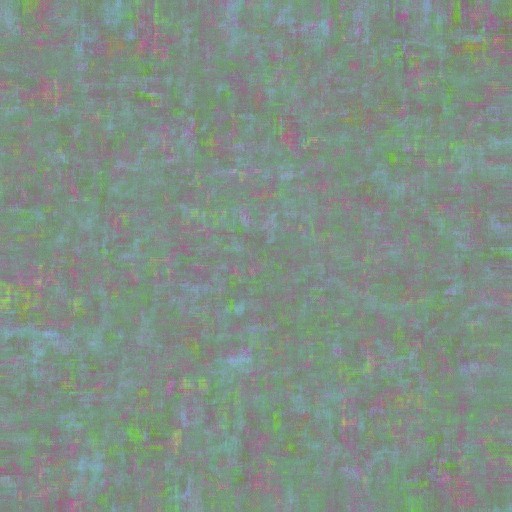}\\
    a) Hourglass-1 & b) Hourglass-3 & c) Hourglass-5 & d) Skip-5 & e) Skip-5-nearest
    \end{tabular}
    \caption{\textbf{``Samples'' from the deep image prior.} We show images that are produced by ConvNets with random weights from independent random uniform noise. Each column shows two images $f_\theta(z)$ for the same architecture, same input noise $z$, and two different random $\theta$. The following architectures are visualized: a) an hourglass architecture with one downsampling and one bilinear upsampling, b) a deeper hourglass architecture with three downsampling and three bilinear upsampling layers, c) an even deeper hourglass architecture with five downsampling and five bilinear upsampling layers, d) same as (c), but with skip connections (each skip connection has a convolution layer), e) same as (d), but with nearest upsampling.  Note how the resulting images are far from independent noise and correspond to stochastic processes producing spatial structures with clear self-similarity (e.g.\ each image has a distinctive palette). The scale of structures naturally changes with the depth of the network. ``Samples'' for hourglass networks with skip connections (U-Net type) combine structures of different scales, as is typical for natural images.}\label{fig:samples}
\end{figure*}

\subsection{``Sampling'' from the deep image prior}

The prior defined by~\cref{eq:reparametrization} is implicit and does not define a proper probability distribution in the image space.
Nevertheless, it is possible to draw ``samples'' (in the loose sense) from this prior by taking random values of the parameters $\theta$ and looking at the generated image $f_\theta(z)$.
In other words, we can visualize the starting points of the optimization process~\cref{eq:reparametrization} before fitting the parameters to the noisy image.
\Cref{fig:samples} shows such ``samples'' from the deep priors captured by different hourglass-type architectures. The samples exhibit spatial structures and self-similarities, whereas the scale of these structures depends on the depth of the network.
Adding skip connections results in images that contain structures of different characteristic scales, as is desirable for modeling natural images.
It is therefore natural that such architectures are the most popular choice for generative ConvNets.
They have also performed best in our image restoration experiments described next.

 \section{Applications}\label{s:applications}

We now show experimentally how the proposed prior works for diverse image reconstruction problems. More examples and interactive viewer can be found on the project webpage~\url{https://dmitryulyanov.github.io/deep_image_prior}. 

\addtolength{\tabcolsep}{-2mm}
\begin{figure*}
    \centering
    \deflen{twolenn}{3cm}
    \renewcommand\makespy[1]{%

        \begin{tikzpicture}[spy using outlines={rectangle,magnification=2.5, width=0.5\twolenn, height=0.25\twolenn, every spy on node/.append style={ultra thick}}]
                \node (nd1){\includegraphics[width=\twolenn]{#1}};

                \RelativeSpy{nd1-spy1}{nd1}{(0.37,0.355)}{(0.253,-0.12)}{green}
                \RelativeSpy{nd1-spy4}{nd1}{(0.87,0.694)}{(0.736,-0.12)}{blue}
        \end{tikzpicture}
    }

    \begin{tabular}{ccccc}
    \makespy{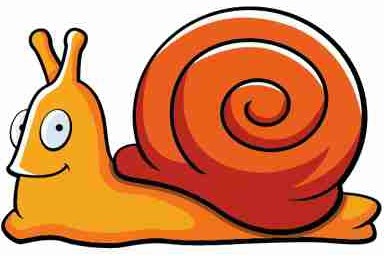}&
    \makespy{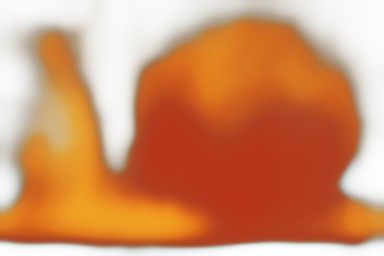}&
    \makespy{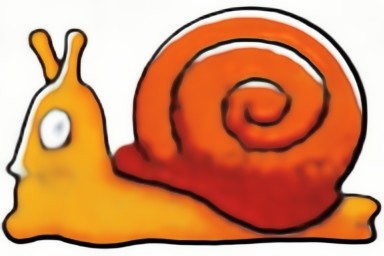}&
    \makespy{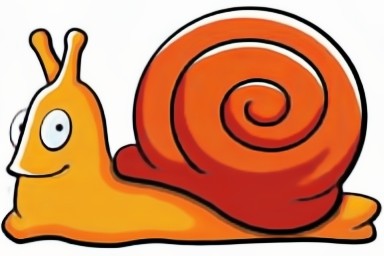}&
    \makespy{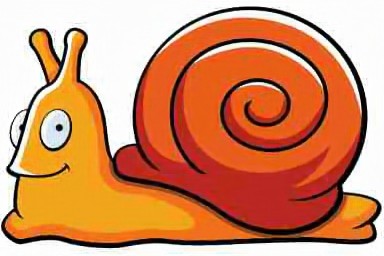}\\
    Corrupted&100 iterations&600 iterations&2400 iterations&50K iterations
    \end{tabular}

    \caption{\textbf{Blind restoration of a JPEG-compressed image.} (\textit{electronic zoom-in recommended}) Our approach can restore an image with a complex degradation (JPEG compression in this case). As the optimization process progresses, the deep image prior allows to recover most of the signal while getting rid of halos and blockiness (after 2400 iterations) before eventually overfitting to the input (at 50K iterations).}\label{fig:jpeg}
\end{figure*}
\addtolength{\tabcolsep}{2mm} 
\deflen{lendenoising}{0.195\textwidth}
\renewcommand\makespy[1]{%

    \begin{tikzpicture}[spy using outlines={rectangle,magnification=2.7, height=10cm, width=15cm, every spy on node/.append style={line width=2.5mm}}]
            \node (nd1){\includegraphics{#1}};

            \RelativeSpy{nd1-spy1}{nd1}{(0.6,0.49)}{(0.5,0.05)}{green}
    \end{tikzpicture}
}

\newcommand\makespyhill[1]{%

    \begin{tikzpicture}[spy using outlines={rectangle,magnification=2.7, height=5cm, width=7.5cm, every spy on node/.append style={line width=1.25mm}}]
            \node (nd1){\includegraphics{#1}};

            \RelativeSpy{nd1-spy1}{nd1}{(0.6,0.49)}{(0.5,0.05)}{green}
    \end{tikzpicture}
}

\newcommand\makespykodim[1]{%

    \begin{tikzpicture}[spy using outlines={rectangle,magnification=2.7, height=14cm, width=24cm, every spy on node/.append style={line width=2.5mm}}]
            \node (nd1){\includegraphics{#1}};

            \RelativeSpy{nd1-spy1}{nd1}{(0.6,0.49)}{(0.5,-0.15)}{green}
    \end{tikzpicture}
}

\begin{figure*}
    \centering


    \begin{subfigure}[b]{\lendenoising}
        \resizebox{1.02\textwidth}{!}{
            \makespy{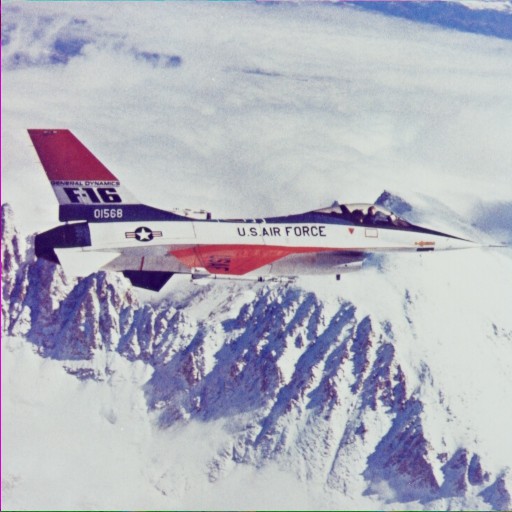}
        }
    \end{subfigure}
    \begin{subfigure}[b]{\lendenoising}
        \resizebox{1.02\textwidth}{!}{
            \makespy{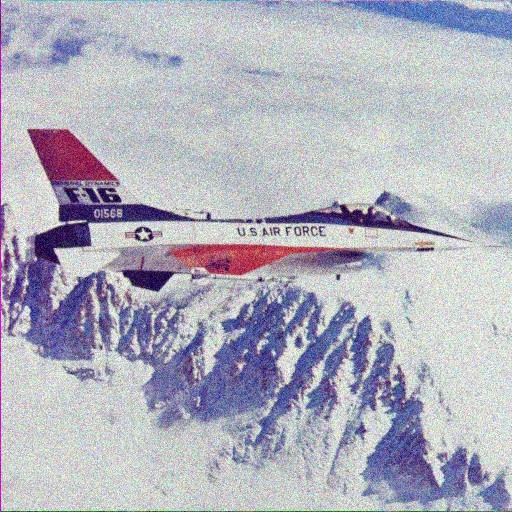}
        }
    \end{subfigure}
    \begin{subfigure}[b]{\lendenoising}
        \resizebox{1.02\textwidth}{!}{
            \makespy{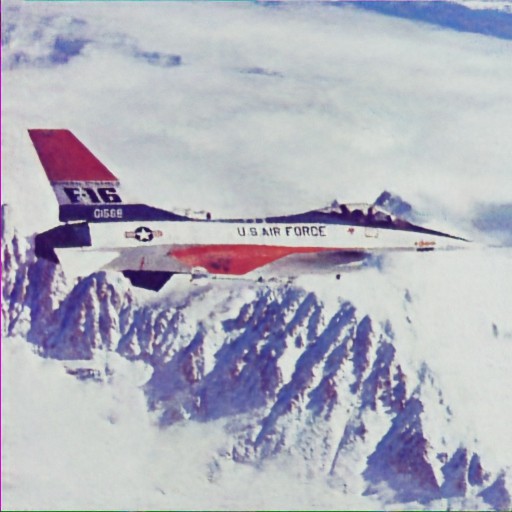}
        }
    \end{subfigure}
    \begin{subfigure}[b]{\lendenoising}
        \resizebox{1.02\textwidth}{!}{
            \makespy{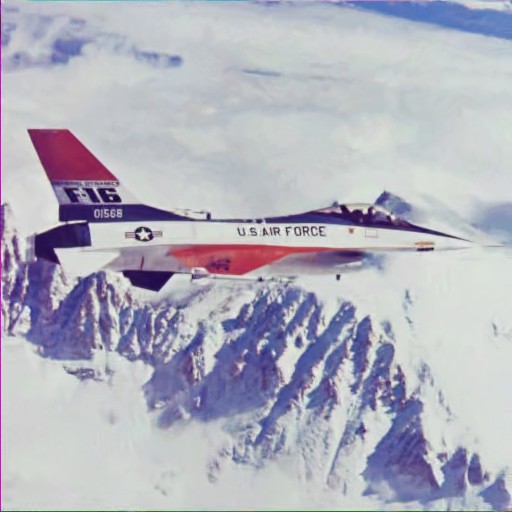}
        }
    \end{subfigure}
    \begin{subfigure}[b]{\lendenoising}
        \resizebox{1.02\textwidth}{!}{
            \makespy{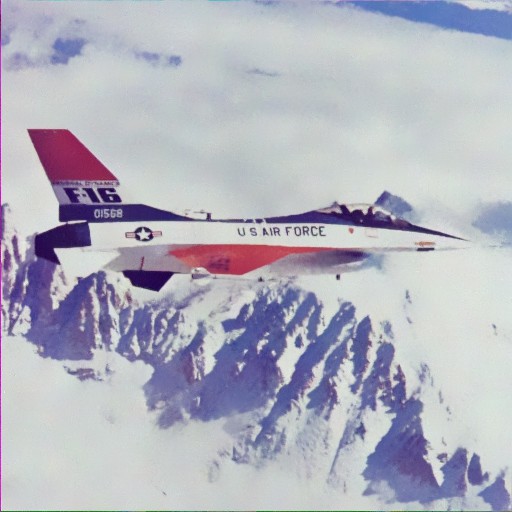}
        }
    \end{subfigure}\\%
    \begin{subfigure}[b]{\lendenoising}
        \vspace*{2mm}\resizebox{1.02\textwidth}{!}{
            \makespyhill{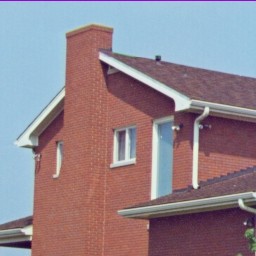}
        }
    \end{subfigure}
    \begin{subfigure}[b]{\lendenoising}
        \vspace*{2mm}\resizebox{1.02\textwidth}{!}{
            \makespyhill{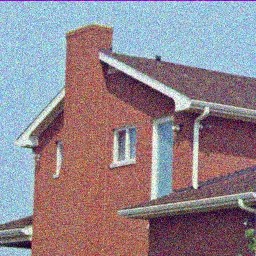}
        }
    \end{subfigure}
    \begin{subfigure}[b]{\lendenoising}
        \vspace*{2mm}\resizebox{1.02\textwidth}{!}{
            \makespyhill{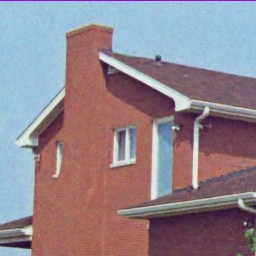}
        }
    \end{subfigure}
    \begin{subfigure}[b]{\lendenoising}
        \vspace*{2mm}\resizebox{1.02\textwidth}{!}{
            \makespyhill{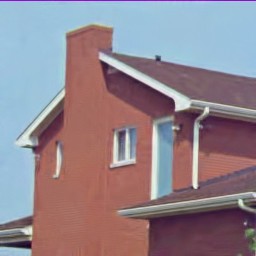}
        }
    \end{subfigure}
    \begin{subfigure}[b]{\lendenoising}
        \vspace*{2mm}\resizebox{1.02\textwidth}{!}{
            \makespyhill{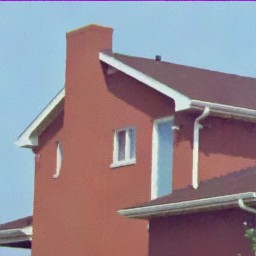}
        }
    \end{subfigure}
    \begin{subfigure}[b]{\lendenoising}
        \vspace*{2mm}\resizebox{1.02\textwidth}{!}{
            \makespykodim{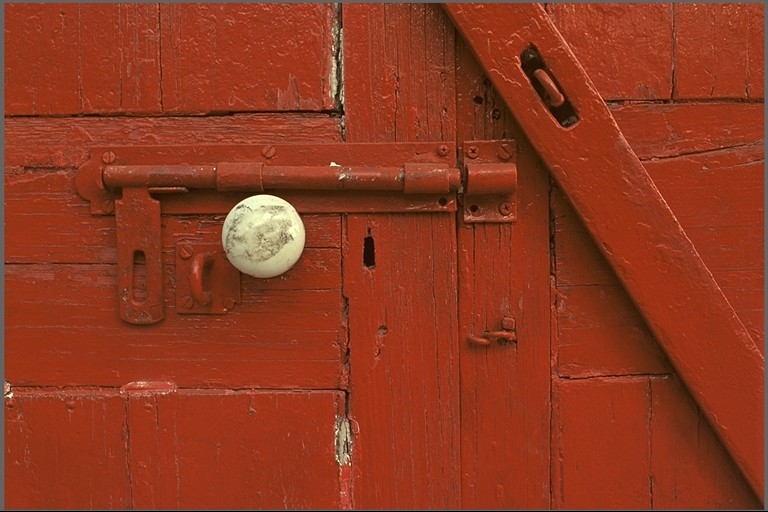}
        }
    \end{subfigure}
    \begin{subfigure}[b]{\lendenoising}
        \vspace*{2mm}\resizebox{1.02\textwidth}{!}{
            \makespykodim{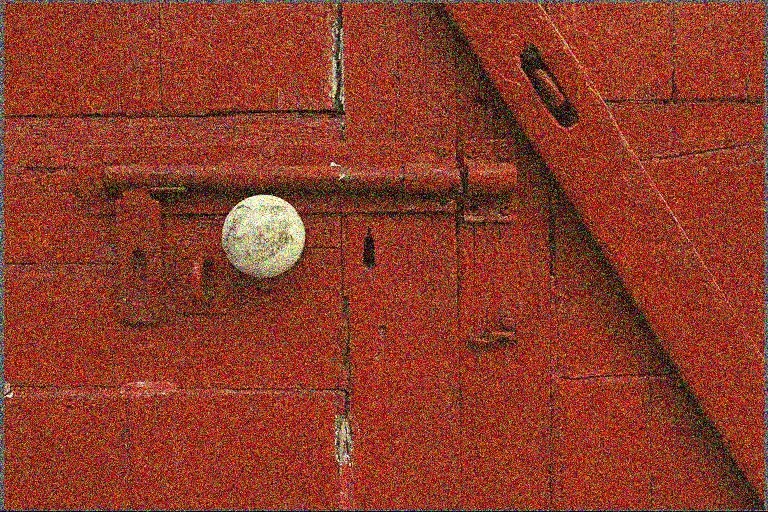}
        }
    \end{subfigure}
    \begin{subfigure}[b]{\lendenoising}
        \vspace*{2mm}\resizebox{1.02\textwidth}{!}{
            \makespykodim{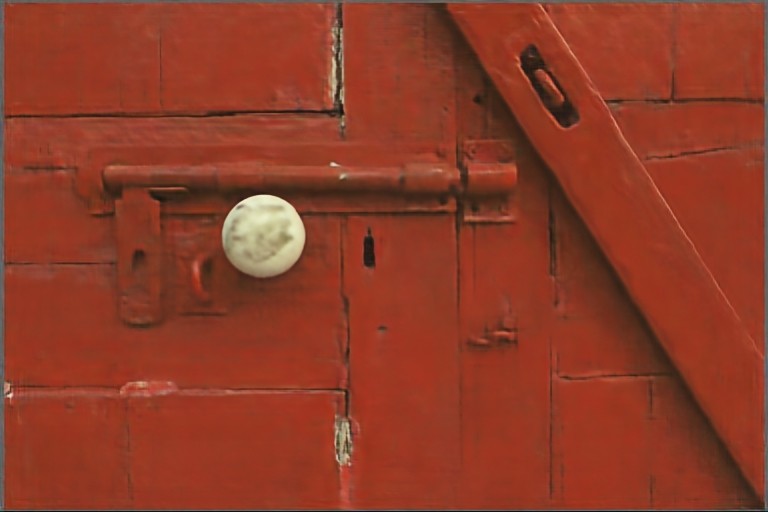}
        }
    \end{subfigure}
    \begin{subfigure}[b]{\lendenoising}
        \vspace*{2mm}\resizebox{1.02\textwidth}{!}{
            \makespykodim{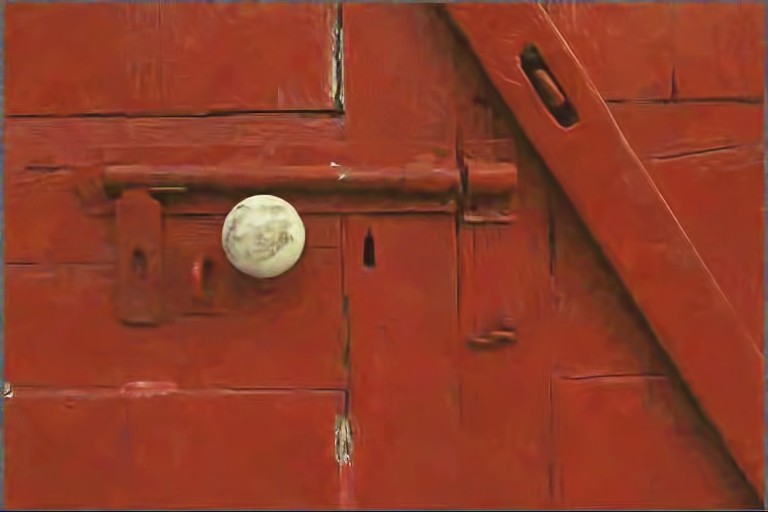}
        }
    \end{subfigure}
    \begin{subfigure}[b]{\lendenoising}
        \vspace*{2mm}\resizebox{1.02\textwidth}{!}{
            \makespykodim{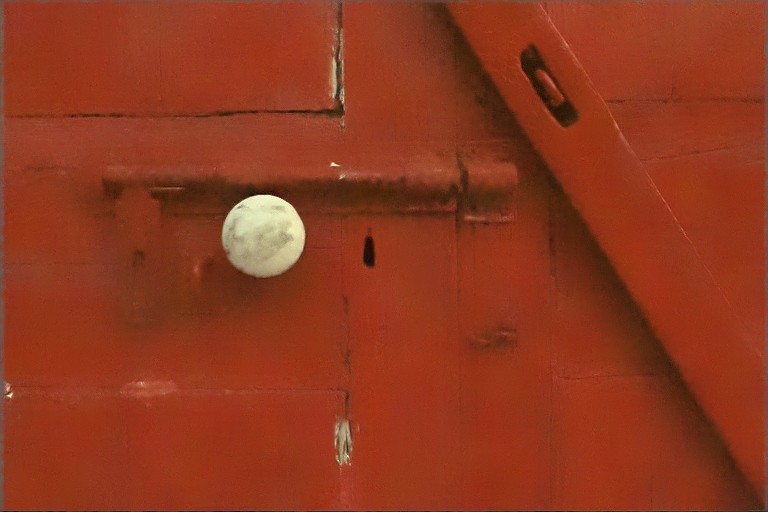}
        }
    \end{subfigure}
         \begin{subfigure}[b]{\lendenoising}
        \vspace*{2mm}\resizebox{1.02\textwidth}{!}{
            \makespy{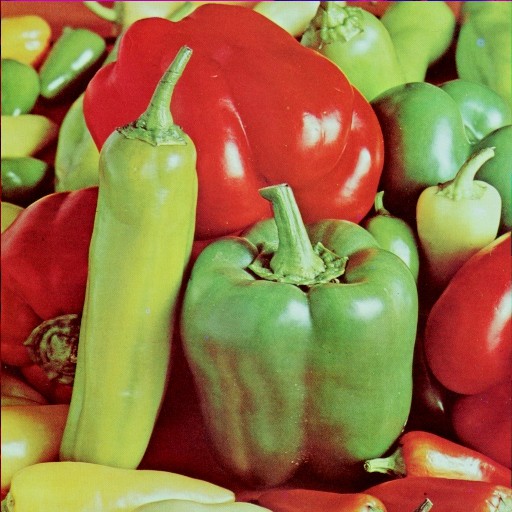}
        }
        \caption{GT}
    \end{subfigure}
    \begin{subfigure}[b]{\lendenoising}
        \vspace*{2mm}\resizebox{1.02\textwidth}{!}{
            \makespy{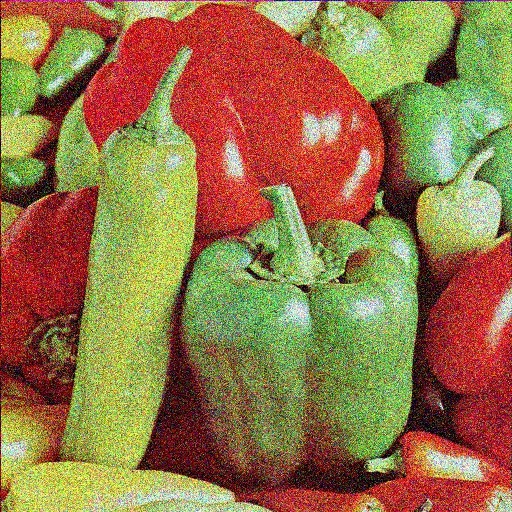}
        }
        \caption{Input}
    \end{subfigure}
    \begin{subfigure}[b]{\lendenoising}
        \vspace*{2mm}\resizebox{1.02\textwidth}{!}{
            \makespy{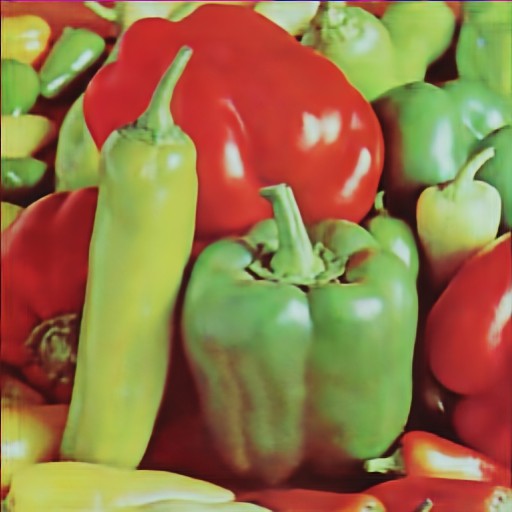}
        }
        \caption{Ours}
    \end{subfigure}
    \begin{subfigure}[b]{\lendenoising}
        \vspace*{2mm}\resizebox{1.02\textwidth}{!}{
            \makespy{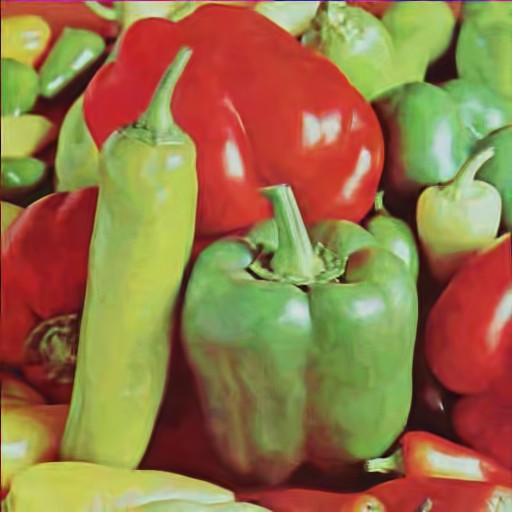}
        }
        \caption{CBM3D}
    \end{subfigure}
    \begin{subfigure}[b]{\lendenoising}
        \vspace*{2mm}\resizebox{1.02\textwidth}{!}{
            \makespy{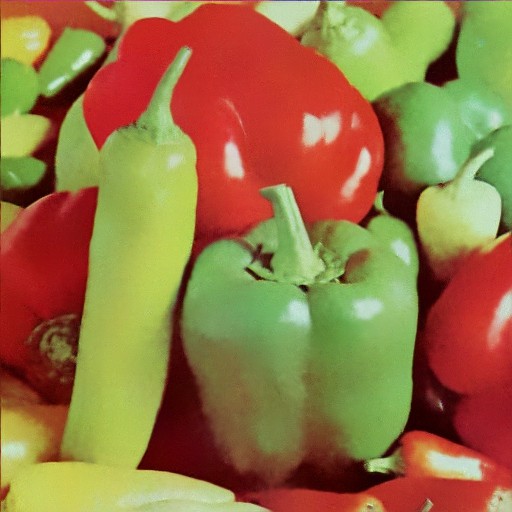}
        }
        \caption{NLM}
    \end{subfigure}

    \caption{\textbf{Blind image denoising.} The deep image prior is successful at recovering both man-made and natural patterns. For reference, the result of a state-of-the-art non-learned denoising approach~\cite{dabov2007image,buades2005non} is shown.}\label{fig:denoising}
\end{figure*}

\subsection{Denoising and generic reconstruction} As our parametrization presents high impedance to image noise, it can be naturally used to filter out noise from an image. The aim of denoising is to recover a clean image $x$ from a noisy observation $x_0$. Sometimes the degradation model is known: $x_0 = x + \epsilon$ where $\epsilon$ follows a particular distribution. However, more often in \emph{blind denoising} the noise model is unknown.

Here we work under the blindness assumption, but the method can be easily modified to incorporate information about noise model. We use the same exact formulation as~\cref{eq:denoise,eq:denoise2} and, given a noisy image $x_0$, recover a clean image $x^* = f_{\theta^*}(z)$ after substituting the minimizer $\theta^*$ of~\cref{eq:denoise2}.

Our approach does not require a model for the image degradation process that it needs to revert. This allows it to be applied in a ``plug-and-play'' fashion to image restoration tasks, where the degradation process is complex and/or unknown and where obtaining realistic data for supervised training is difficult. We demonstrate this capability by several qualitative examples in~\cref{fig:denoising}, where our approach uses the quadratic energy~\eqref{eq:denoise} leading to formulation~\eqref{eq:denoise2} to restore images degraded by complex and unknown compression artifacts. \Cref{fig:jpeg} (top row) also demonstrates the applicability of the method beyond natural images (a cartoon in this case).


We evaluate our denoising approach on the standard dataset\footnote{\url{http://www.cs.tut.fi/~foi/GCF-BM3D/index.html\#ref_results}}, consisting of 9 colored images with noise strength of $\sigma = 25$. We achieve a PSNR of $29.22$ after 1800 optimization steps. The score is improved up to $30.43$ if we additionally average the restored images obtained in the last iterations (using exponential sliding window). If averaged over two optimization runs our method further improves up to $31.00$ PSNR\@. For reference, the scores for the two popular approaches CMB3D~\cite{dabov2007image} and Non-local means~\cite{buades2005non}, that do not require pretraining, are $31.42$ and $30.26$ respectively.

To validate if the deep image prior is suitable for denoising images corrupted with real-world non-Gaussian noise we use the benchmark of~\cite{plotz2017benchmarking}. Using the same architecture and hyper-parameters as for~\cref{fig:jpeg} we get $41.95$ PSNR, while CBM3D's score is only $30.13$. We also use the deep image prior with different network architectures and get $35.05$ PSNR for UNet and $31.95$ for ResNet.  The details of each architecture are described in~\cref{s:tech_details}. Our hour-glass architecture resembles UNet, yet has less number of skip connections and additional BatchNorms before concatenation operators. We speculate that the overly wide skip-connections within UNet lead to a prior that are somewhat too weak and the fitting happens too fast; while the lack of skip-connections in ResNet leads to slow fitting and a prior that is too strong. Overall, this stark difference in the performance of different architectures emphasizes that different architectures impose rather different priors leading to very different results.

\deflen{fourlennsr}{0.194\textwidth}
\begin{figure*}
    \centering
    \vspace*{4mm} \textbf{4$\times$ super-resolution} \\ \vspace*{4mm}
    \renewcommand\makespy[1]{%
    \begin{adjustbox}{width=1.02\textwidth}
        \begin{tikzpicture}[spy using outlines={rectangle,magnification=3, height=4.5cm, width=8.95cm, every spy on node/.append style={line width=2mm}}]
                \node (nd1){\includegraphics{#1}};
                \RelativeSpy{nd1-spy1}{nd1}{(0.14,0.195)}{(0.251,-0.13)}{red}
                \RelativeSpy{nd1-spy4}{nd1}{(0.094,0.945)}{(0.748,-0.13)}{blue}
        \end{tikzpicture}%
    \end{adjustbox}
    }%
    \newcommand\makespyy[1]{%
    \begin{adjustbox}{width=1.033\textwidth}
        \begin{tikzpicture}[spy using outlines={rectangle,magnification=2, height=2.25cm, width=4.4cm, every spy on node/.append style={line width=1mm}}]
                \node (nd1){\includegraphics{#1}};
                \RelativeSpy{nd1-spy1}{nd1}{(0.63,0.485)}{(0.253,-0.13)}{red}
                \RelativeSpy{nd1-spy4}{nd1}{(0.19,0.745)}{(0.750,-0.13)}{blue}
        \end{tikzpicture}%
    \end{adjustbox}
    }%
    \newcommand\makespyyy[1]{%
    \begin{adjustbox}{width=1.033\textwidth}
        \begin{tikzpicture}[spy using outlines={rectangle,magnification=2, height=2.25cm, width=4.4cm, every spy on node/.append style={line width=1mm}}]
                \node (nd1){\includegraphics{#1}};
                \RelativeSpy{nd1-spy1}{nd1}{(0.63,0.685)}{(0.253,-0.13)}{red}
                \RelativeSpy{nd1-spy4}{nd1}{(0.19,0.345)}{(0.750,-0.13)}{blue}
        \end{tikzpicture}%
    \end{adjustbox}
    }%
    \newcommand\makespyyyy[1]{%
    \begin{adjustbox}{width=1.033\textwidth}
        \begin{tikzpicture}[spy using outlines={rectangle,magnification=2, height=4.5cm, width=8.95cm, every spy on node/.append style={line width=1mm}}]
                \node (nd1){\includegraphics{#1}};
                \RelativeSpy{nd1-spy1}{nd1}{(0.18,0.195)}{(0.253,-0.13)}{red}
                \RelativeSpy{nd1-spy4}{nd1}{(0.59,0.845)}{(0.750,-0.13)}{blue}
        \end{tikzpicture}%
    \end{adjustbox}
    }%
    \hspace*{-1.3mm}%
    \begin{subfigure}[h]{\fourlennsr}
        \makespyy{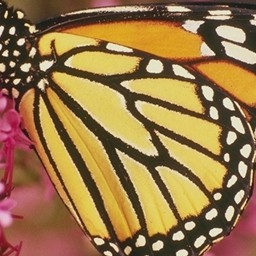}
    \end{subfigure}
    \begin{subfigure}[h]{\fourlennsr}
        \makespyy{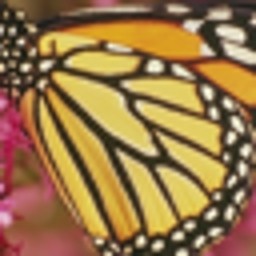}
    \end{subfigure}
    \begin{subfigure}[h]{\fourlennsr}
        \makespyy{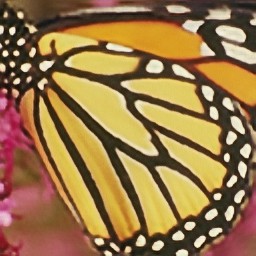}
    \end{subfigure}
    \begin{subfigure}[h]{\fourlennsr}
        \makespyy{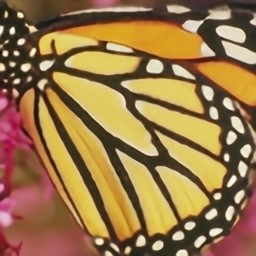}
    \end{subfigure}
    \begin{subfigure}[h]{\fourlennsr}
        \makespyy{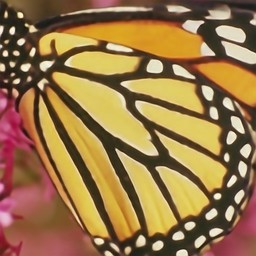}
    \end{subfigure}
    \\ 
    \begin{subfigure}[h]{\fourlennsr}
        \makespy{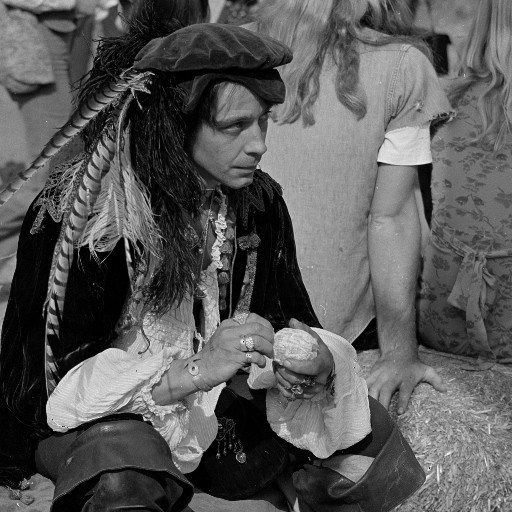}
        \vspace*{-4mm}\caption{Original image \\ \quad}
    \end{subfigure}
    \begin{subfigure}[h]{\fourlennsr}
    \makespy{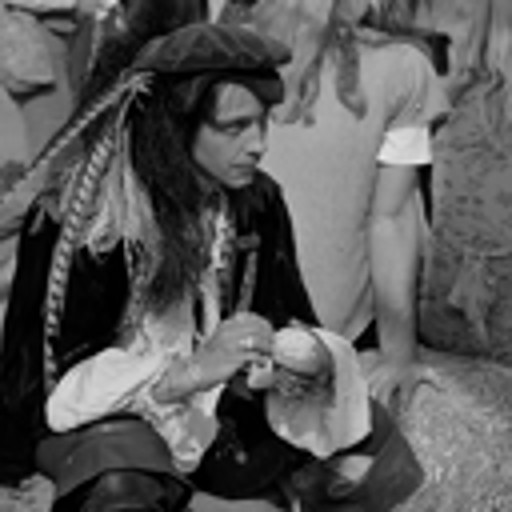}
        \vspace*{-4mm}\caption{Bicubic, \\ \textbf{Not trained}}
    \end{subfigure}
    \begin{subfigure}[h]{\fourlennsr}
    \makespy{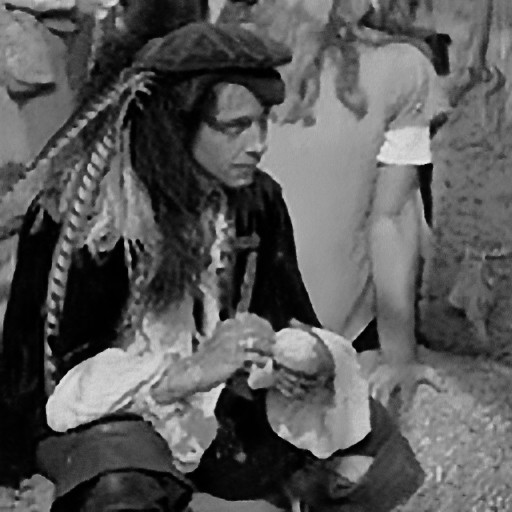}
        \vspace*{-4mm}\caption{Ours, \\ \textbf{Not trained}}
    \end{subfigure}
    \begin{subfigure}[h]{\fourlennsr}
           \makespy{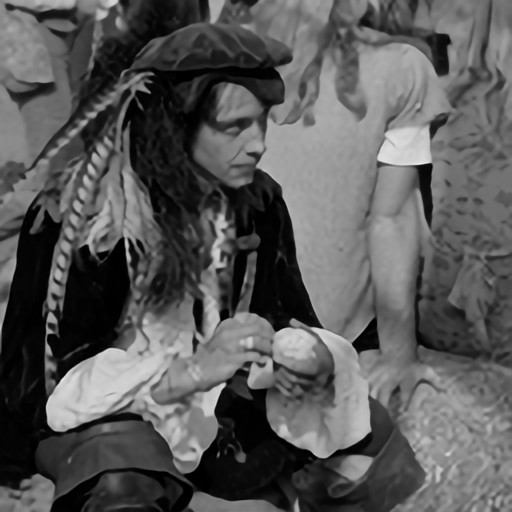}
        \vspace*{-4mm}\caption{LapSRN, \\ \textbf{Trained}}
    \end{subfigure}
    \begin{subfigure}[h]{\fourlennsr}
            \makespy{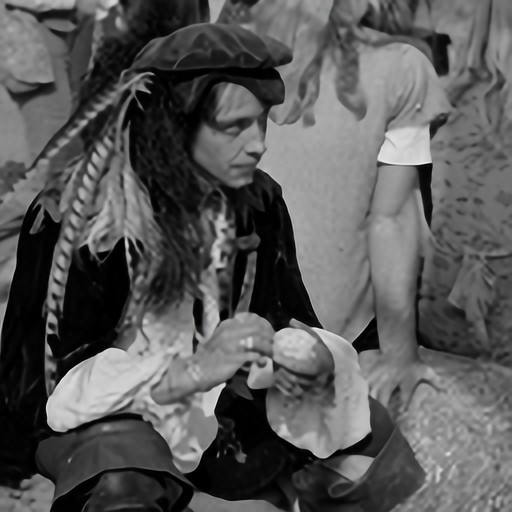}
        \vspace*{-4mm}\caption{SRResNet, \\ \textbf{Trained}}
    \end{subfigure} \\ \vspace*{4mm}
    \vspace*{4mm} \textbf{8$\times$ super-resolution} \\ \vspace*{4mm}
    \begin{subfigure}[h]{\fourlennsr}
        \makespyyy{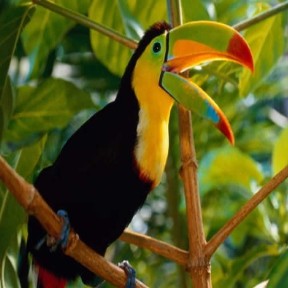}
    \end{subfigure}
    \begin{subfigure}[h]{\fourlennsr}
        \makespyyy{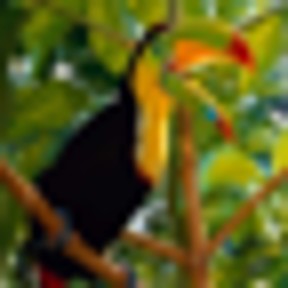}
    \end{subfigure}
    \begin{subfigure}[h]{\fourlennsr}
        \makespyyy{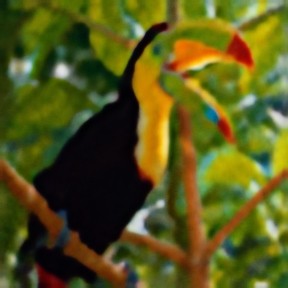}
    \end{subfigure}
    \begin{subfigure}[h]{\fourlennsr}
        \makespyyy{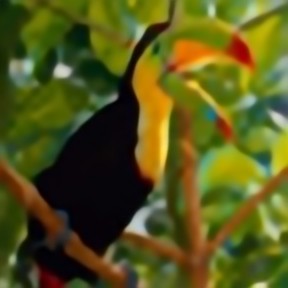}
    \end{subfigure}
    \begin{subfigure}[h]{\fourlennsr}
        \makespyyy{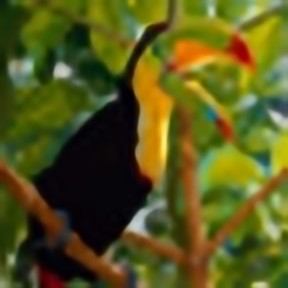}
    \end{subfigure}
    \\ 
    \begin{subfigure}[h]{\fourlennsr}
        \makespyyyy{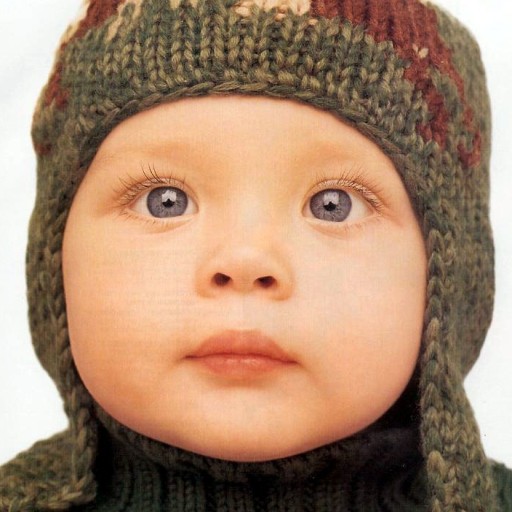}
        \vspace*{-4mm}\caption{Original image \\ \quad}
    \end{subfigure}
    \begin{subfigure}[h]{\fourlennsr}
        \makespyyyy{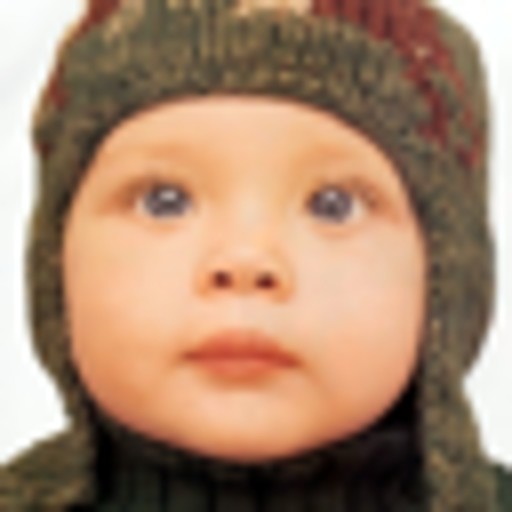}
        \vspace*{-4mm}\caption{Bicubic, \\ \textbf{Not trained}}
    \end{subfigure}
    \begin{subfigure}[h]{\fourlennsr}
    \makespyyyy{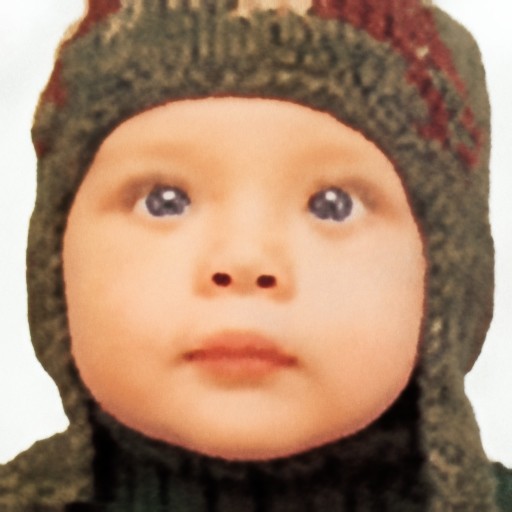}
        \vspace*{-4mm}\caption{Ours, \\ \textbf{Not trained}}
    \end{subfigure}
    \begin{subfigure}[h]{\fourlennsr}
           \makespyyyy{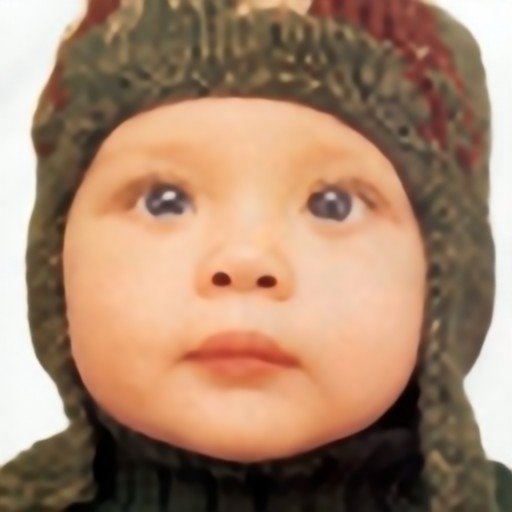}
        \vspace*{-4mm}\caption{LapSRN, \\ \textbf{Trained}}
    \end{subfigure}
    \begin{subfigure}[h]{\fourlennsr}
            \makespyyyy{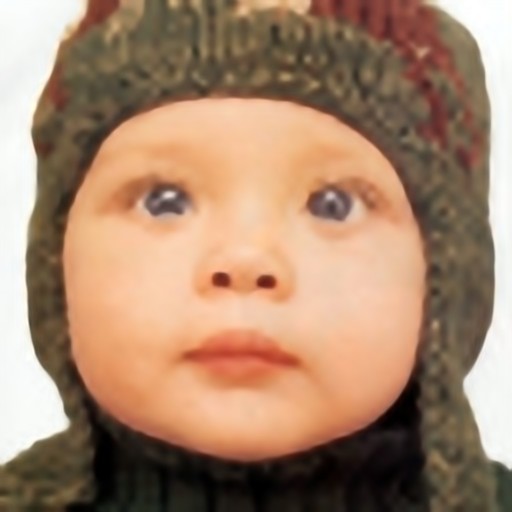}
        \vspace*{-4mm}\caption{VDSR, \\ \textbf{Trained}}
    \end{subfigure} \\ \vspace*{4mm}
    \caption{\textbf{4$\times$ and 8$\times$ Image super-resolution.} Similarly to e.g.\ bicubic upsampling, our method never has access to any data other than a single low-resolution image, and yet it produces much cleaner results with sharp edges close to state-of-the-art super-resolution methods (LapSRN~\cite{Lai17sr}, SRResNet~\cite{Ledig17sr}, VDSR~\cite{Kim16sr}) which utilize networks trained from large datasets.}\label{fig:sr}
\end{figure*}

\begin{figure*}
\centering
\deflen{fivelen}{0.195\linewidth}
\begin{subfigure}[b]{\fivelen}
    \includegraphics[width=\linewidth]{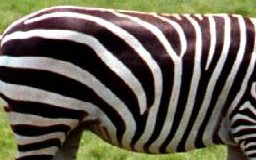}
    \caption{HR image}
\end{subfigure}
\begin{subfigure}[b]{\fivelen}
    \includegraphics[width=\linewidth]{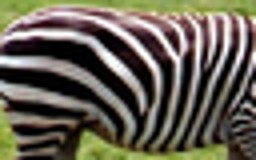}
    \caption{Bicubic upsampling}
\end{subfigure}
\begin{subfigure}[b]{\fivelen}
    \includegraphics[width=\linewidth]{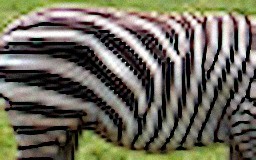}
    \caption{No prior}
\end{subfigure}
\begin{subfigure}[b]{\fivelen}
    \includegraphics[width=\linewidth]{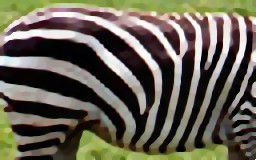}
    \caption{TV prior}
\end{subfigure}
\begin{subfigure}[b]{\fivelen}
    \includegraphics[width=\linewidth]{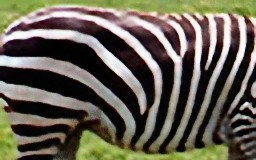}
    \caption{Deep image prior}
\end{subfigure}
\vspace*{-1mm}\caption{\textbf{Prior effect in super-resolution.} Direct optimization of data term $E(x; x_0)$ with respect to the pixels (c) leads to ringing artifacts. TV prior removes ringing artifacts (d) but introduces cartoon effect. Deep prior (e) leads to the result that is both clean and sharp.}\label{fig:prior_effect}
\end{figure*} \begin{table*}
\centering
\textbf{4$\times$ super-resolution} \\ \vspace*{2mm}
\resizebox{\textwidth}{!}{\begin{tabular}{@{}lccccccccccccccc@{}}\toprule
    & \small{Baboon} & \small{Barbara} & \small{Bridge} & \small{Coastguard} & \small{Comic} & \small{Face} & \small{Flowers} & \small{Foreman} & \small{Lenna} & \small{Man} & \small{Monarch} & \small{Pepper} & \small{Ppt3} & \small{Zebra} & \small{\textbf{Avg.}} \\ \midrule
\small{No prior} & $22.24$ & $24.89$ & $23.94$ & $24.62$ & $21.06$ & $29.99$ & $23.75$ & $29.01$ & $28.23$ & $24.84$ & $25.76$ & $28.74$ & $20.26$ & $21.69$ & $24.93$ \\
\small{Bicubic} &  $\bm{22.44}$ & $25.15$ & $24.47$ & $25.53$ & $21.59$ & $\bm{31.34}$ & $25.33$ & $29.45$ & $29.84$ & $25.7$ & $27.45$ & $30.63$ & $21.78$ & $24.01$  & $26.05$  \\
\small{TV prior} & $22.34$ & $24.78$ & $24.46$ & $25.78$ & $21.95$ & $\bm{31.34}$ & $25.91$ & $30.63$ & $29.76$ & $25.94$ & $28.46$ & $31.32$ & $22.75$ & $24.52$  & $26.42$  \\
\small{Glasner et al.} & $\bm{22.44}$ & $25.38$ & $\bm{24.73}$ & $25.38$ & $21.98$ & $31.09$ & $25.54$ & $30.40$ & $30.48$ & $\bm{26.33}$ & $28.22$ & $32.02$ & $22.16$ & $24.34$ & $26.46$  \\
\small{Ours} & $22.29$ & $\bm{25.53}$ & $24.38$ & $\bm{25.81}$ & $\bm{22.18}$ & $31.02$ & $\bm{26.14}$ & $\bm{31.66}$ & $\bm{30.83}$ & $26.09$ & $\bm{29.98}$ & $\bm{32.08}$ & $\bm{24.38}$ & $\bm{25.71}$  & $\bm{27.00}$  \\ \midrule
\small{SRResNet-MSE} & $23.0$ & $26.08$ & $25.52$ & $26.31$ & $23.44$ & $32.71$ & $28.13$ & $33.8$ & $32.42$ & $27.43$ & $32.85$ & $34.28$ & $26.56$ & $26.95$  & $28.53$ \\
\small{LapSRN} & $22.83$ & $25.69$ & $25.36$ & $26.21$ & $22.9$ & $32.62$ & $27.54$ & $33.59$ & $31.98$ & $27.27$ & $31.62$ & $33.88$ & $25.36$ & $26.98$ & $28.13$ \\ \bottomrule
\end{tabular}} \\ \vspace*{4mm} \textbf{8$\times$ super-resolution} \\ \vspace*{2mm}
\resizebox{\textwidth}{!}{\begin{tabular}{@{}lccccccccccccccc@{}}\toprule
        & \small{Baboon} & \small{Barbara} & \small{Bridge} & \small{Coastguard} & \small{Comic} & \small{Face} & \small{Flowers} & \small{Foreman} & \small{Lenna} & \small{Man} & \small{Monarch} & \small{Pepper} & \small{Ppt3} & \small{Zebra} & \small{\textbf{Avg.}} \\ \midrule
    \small{No prior} & $21.09$ & $23.04$ & $21.78$ & $23.63$ & $18.65$ & $27.84$ & $21.05$ & $25.62$ & $25.42$ & $22.54$ & $22.91$ & $25.34$ & $18.15$ & $18.85$ & $22.56$ \\
    \small{Bicubic}  & $21.28$ & $23.44$ & $22.24$ & $23.65$ & $19.25$ & $28.79$ & $22.06$ & $25.37$ & $26.27$ & $23.06$ & $23.18$ & $26.55$ & $18.62$ & $19.59$ & $23.09$ \\
    \small{TV prior} & $21.30$ & $23.72$ & $\bm{22.30}$ & $23.82$ & $19.50$ & $28.84$ & $22.50$ & $26.07$ & $26.74$ & $23.53$ & $23.71$ & $27.56$ & $19.34$ & $19.89$ & $23.48$ \\
    \small{SelfExSR} & $21.37$ & $23.90$ & $22.28$ & $24.17$ & $19.79$ & $29.48$ & $\bm{22.93}$ & $27.01$ & $27.72$ & $\bm{23.83}$ & $24.02$ & $28.63$ & $20.09$ & $20.25$ & $23.96$ \\
    \small{Ours}     & $\bm{21.38}$ & $\bm{23.94}$ & $22.20$ & $\bm{24.21}$ & $\bm{19.86}$ & $\bm{29.52}$ & $22.86$ & $\bm{27.87}$ & $\bm{27.93}$ & $23.57$ & $\bm{24.86}$ & $\bm{29.18}$ & $\bm{20.12}$ & $\bm{20.62}$ & $\bm{24.15}$ \\ \midrule
    \small{LapSRN}   & $21.51$ & $24.21$ & $22.77$ & $24.10$ & $20.06$ & $29.85$ & $23.31$ & $28.13$ & $28.22$ & $24.20$ & $24.97$ & $29.22$ & $20.13$ & $20.28$ & $24.35$ \\ \bottomrule
\end{tabular}}
\caption{Detailed super-resolution PSNR comparison on the Set14 dataset with different scaling factors.}\label{tab:sr_s14}
\end{table*}

\begin{table}
    \centering
    \textbf{4$\times$ super-resolution} \\ \vspace*{2mm}
    \resizebox{\linewidth}{!}{\begin{tabular}{@{}lcccccc@{}}\toprule
	    & \small{Baby} & \small{Bird} & \small{Butterfly} & \small{Head} & \small{Woman} & \small{\textbf{Avg.}} \\ \midrule
	\small{No prior} & $30.16$ & $27.67$ & $19.82$ & $29.98$ & $25.18$ & $26.56$ \\
	\small{Bicubic} & $31.78$ & $30.2$ & $22.13$ & $31.34$ & $26.75$ & $28.44$ \\
	\small{TV prior} & $31.21$ & $30.43$ & $24.38$ & $31.34$ & $26.93$ & $28.85$ \\
	\small{Glasner et al.} & $\bm{32.24}$ & $31.10$ & $22.36$ & $\bm{31.69}$ & $26.85$ & $28.84$ \\
	\small{Ours} & $31.49$ & $\bm{31.80}$ & $\bm{26.23}$ & $31.04$ & $\bm{28.93}$ & $\bm{29.89}$ \\ \midrule
	\small{LapSRN} & $33.55$ & $33.76$ & $27.28$ & $32.62$ & $30.72$ & $31.58$\\
	\small{SRResNet-MSE} & $33.66$ & $35.10$ & $28.41$ & $32.73$ & $30.6$ & $32.10$ \\ \bottomrule
    \end{tabular}} \\ \vspace*{4mm} \textbf{8$\times$ super-resolution} \\ \vspace*{2mm}
    \resizebox{\linewidth}{!}{\begin{tabular}{@{}lcccccc@{}}\toprule
	    & \small{Baby} & \small{Bird} & \small{Butterfly} & \small{Head} & \small{Woman} & \small{\textbf{Avg.}} \\ \midrule
	\small{No prior} & $26.28$ & $24.03$ & $17.64$ & $27.94$ & $21.37$ & $23.45$ \\
	\small{Bicubic} & $27.28$ & $25.28$ & $17.74$ & $28.82$ & $22.74$ & $24.37$  \\
	\small{TV prior} & $27.93$ & $25.82$ & $18.40$ & $28.87$ & $23.36$ & $24.87$  \\
	\small{SelfExSR} & $\bm{28.45}$ & $26.48$ & $18.80$ & $29.36$ & $24.05$ & $25.42$  \\
	\small{Ours} & $28.28$ & $\bm{27.09}$ & $\bm{20.02}$ & $\bm{29.55}$ & $\bm{24.50}$ & $\bm{25.88}$ \\ \midrule
	\small{LapSRN} & $28.88$ & $27.10$ & $19.97$ & $29.76$ & $24.79$ & $26.10$ \\ \bottomrule
	\end{tabular}}
    \caption{Detailed super-resolution PSNR comparison on the Set5 dataset with different scaling factors.}\label{tab:sr_s5}
\end{table}

\subsection{Super-resolution} The goal of super-resolution is to take a low resolution (LR) image $x_0\in\mathbb{R}^{3\times H\times W}$ and upsampling factor $t$, and generate a corresponding high resolution (HR) version $x \in \mathbb{R}^{3 \times tH\times tW}$. To solve this inverse problem, the data term in~\eqref{eq:reparametrization} is set to:
\begin{equation}\label{eq:sr_direct}
  E(x; x_0) = \| d(x) -x_0 \|^2\,,
\end{equation}
where $d(\cdot): \mathbb{R}^{3\times tH\times tW} \rightarrow \mathbb{R}^{3 \times H\times W}$ is a \emph{downsampling operator} that resizes an image by a factor $t$. Hence, the problem is to find the HR image $x$ that, when downsampled, is the same as the LR image $x_0$.
Super-resolution is an ill-posed problem because there are infinitely many HR images $x$ that reduce to the same LR image $x_0$ (i.e.\ the operator $d$ is far from injective). Regularization is required in order to select, among the infinite minimizers of~\eqref{eq:sr_direct}, the most plausible ones.

 Following~\cref{eq:reparametrization}, we regularize the problem by considering the re-parametrization $x=f_\theta(z)$ and optimizing the resulting energy w.r.t.\ $\theta$. Optimization still uses gradient descent, exploiting the fact that both the neural network and the most common downsampling operators, such as Lanczos, are differentiable.

We evaluate super-resolution ability of our approach using {Set5}~\cite{set5} and {Set14}~\cite{set14} datasets. We use a scaling factor of $4$ and $8$ to compare to other works in~\cref{fig:sr}.

Qualitative comparison with bicubic upsampling and state-of-the art learning-based methods SRResNet~\cite{Ledig17sr}, LapSRN~\cite{Tai17sr} is presented in~\cref{fig:sr}. Our method can be fairly compared to bicubic, as both methods never use other data than a given low-resolution image. Visually, we approach the quality of learning-based methods that use the MSE loss. GAN-based~\cite{goodfellow2014generative} methods SRGAN~\cite{Ledig17sr} and EnhanceNet~\cite{Sajjadi17sr} (not shown in the comparison) intelligently hallucinate fine details of the image, which is impossible with our method that uses absolutely no information about the world of HR images.

We compute PSNRs using center crops of the generated images (\cref{tab:sr_s5,tab:sr_s14}). While our method is still outperformed by learning-based approaches, it does considerably better than the non-trained ones (bicubic,~\cite{glasner2009super},~\cite{Huang15sr}).  Visually, it seems to close most of the gap between non-trained methods and state-of-the-art trained ConvNets (c.f.~\cref{fig:splash,fig:sr}).

In~\cref{fig:prior_effect} we compare our deep prior to non-regularized solution and a vanilla TV prior. Our result do not have both ringing artifacts and cartoonish effect.

\begin{figure*}
    \centering
    \deflen{fourlennnnn}{0.244\linewidth}
    \renewcommand\makespy[1]{%
        \includegraphics[width=\textwidth]{#1}
    }
    \begin{subfigure}[b]{\fourlennnnn}
        \resizebox{\textwidth}{!}{
            \makespy{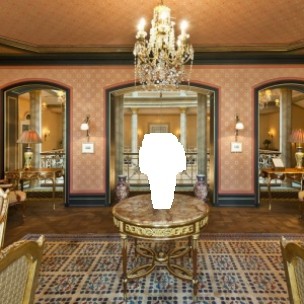}
        }
        \caption{Corrupted image}
    \end{subfigure}
    \begin{subfigure}[b]{\fourlennnnn}
        \resizebox{\linewidth}{!}{
            \makespy{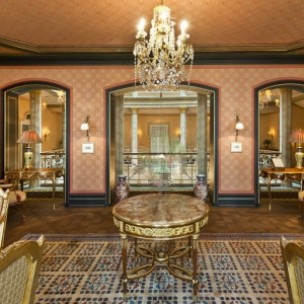}
        }
        \caption{Global-Local GAN~\cite{IizukaSIGGRAPH2017}}
    \end{subfigure}
    \begin{subfigure}[b]{\fourlennnnn}
        \resizebox{\linewidth}{!}{
            \makespy{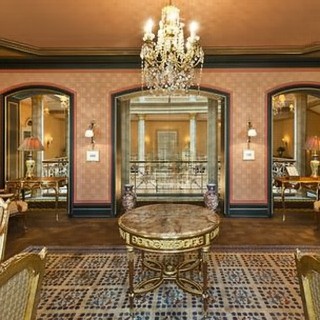}
        }
        \caption{Ours, $\text{LR}=0.01$}
    \end{subfigure}
    \begin{subfigure}[b]{\fourlennnnn}
        \resizebox{\linewidth}{!}{
            \makespy{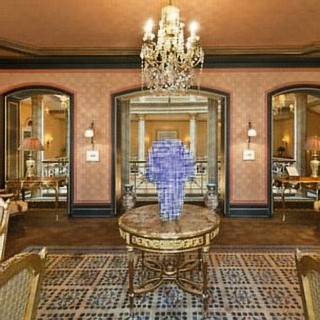}
        }
        \caption{Ours, $\text{LR}=10^{-4}$}
    \end{subfigure}
    \caption{\textbf{Region inpainting.} In many cases, deep image prior is sufficient to successfully inpaint large regions. Despite using no learning, the results may be comparable to~\cite{IizukaSIGGRAPH2017}  which does. The choice of hyper-parameters is important (for example (d) demonstrates sensitivity to the learning rate), but a good setting works well for most images we tried.}\label{fig:inpainting_region}
\end{figure*} \begin{figure*}
    \centering

    \deflen{fourlen}{0.245\linewidth}

    \renewcommand\makespy[1]{%
        \begin{tikzpicture}[spy using outlines={rectangle,magnification=2.7, height=3.0cm, width=5.9cm, every spy on node/.append style={line width=1.5mm}}]
                \node (nd1){\includegraphics{#1}};
                \RelativeSpy{nd1-spy1}{nd1}{(0.17,0.827)}{(0.167,-0.085)}{yellow}
                \RelativeSpy{nd1-spy2}{nd1}{(0.34,0.373)}{(0.5,-0.085)}{blue}
                \RelativeSpy{nd1-spy4}{nd1}{(0.57,0.592)}{(0.833,-0.085)}{green}
        \end{tikzpicture}
    }

    \newcommand\makespyww[1]{%
        \begin{tikzpicture}[spy using outlines={rectangle,magnification=2.7, height=3.0cm, width=5.9cm, every spy on node/.append style={line width=1.5mm}}]
                \node (nd1){\includegraphics{#1}};
                \RelativeSpy{nd1-spy1}{nd1}{(0.57,0.932)}{(0.167,-0.085)}{yellow}
                \RelativeSpy{nd1-spy2}{nd1}{(0.44,0.064)}{(0.5,-0.085)}{blue}
                \RelativeSpy{nd1-spy4}{nd1}{(0.82,0.169)}{(0.833,-0.085)}{green}
        \end{tikzpicture}
    }

    \begin{subfigure}[b]{\fourlen}
        \resizebox{\textwidth}{!}{
            \makespy{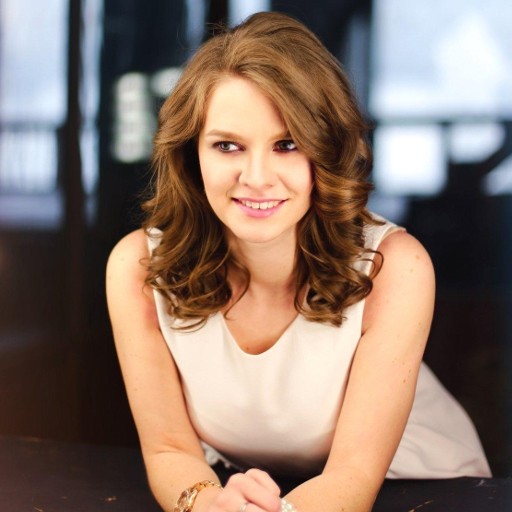}
        }
    \end{subfigure}
    \begin{subfigure}[b]{\fourlen}
        \resizebox{\linewidth}{!}{
            \makespy{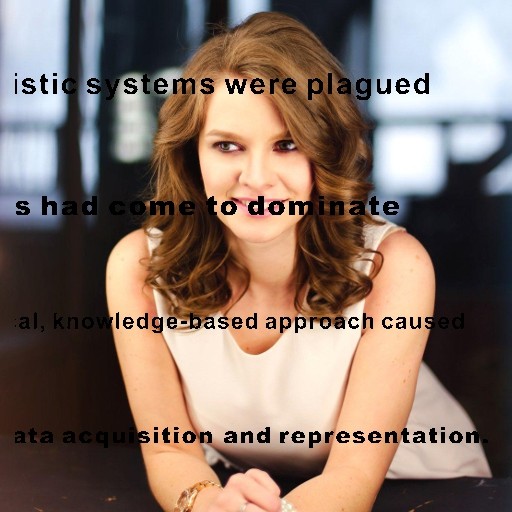}
        }
    \end{subfigure}
    \begin{subfigure}[b]{\fourlen}
        \resizebox{\linewidth}{!}{
            \makespy{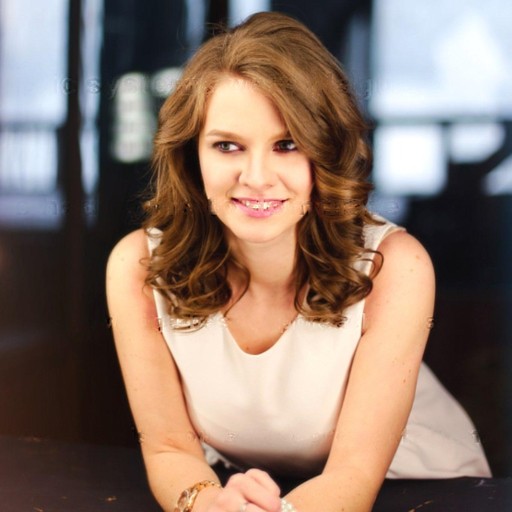}
        }
    \end{subfigure}
    \begin{subfigure}[b]{\fourlen}
        \resizebox{\linewidth}{!}{
            \makespy{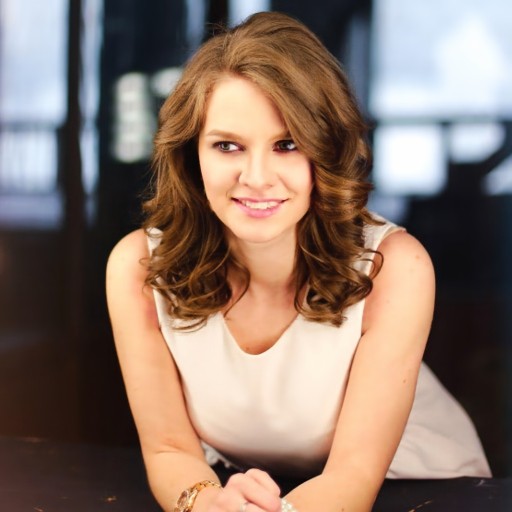}
        }
    \end{subfigure}\\
    \begin{subfigure}[b]{\fourlen}
        \vspace*{1mm}\resizebox{\textwidth}{!}{
            \makespyww{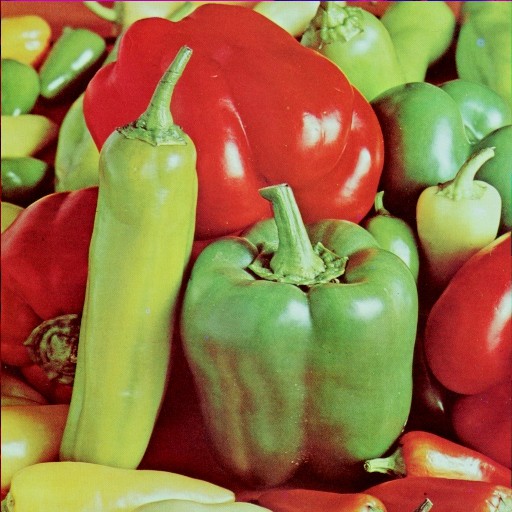}
        }
        \vspace*{-4mm}\caption{Original image}
    \end{subfigure}
    \begin{subfigure}[b]{\fourlen}
        \vspace*{1mm}\resizebox{\linewidth}{!}{
            \makespyww{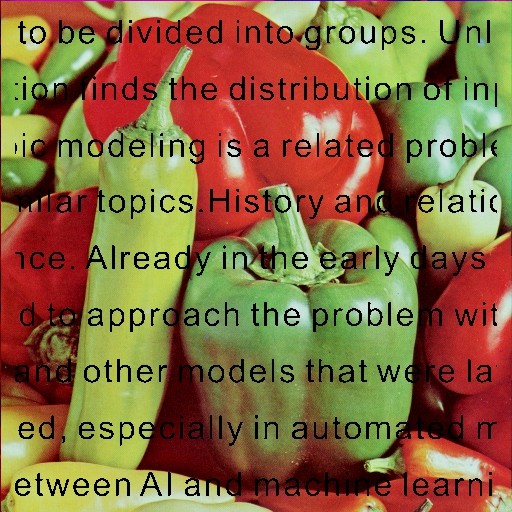}
        }
        \vspace*{-4mm}\caption{Corrupted image}
    \end{subfigure}
    \begin{subfigure}[b]{\fourlen}
        \vspace*{1mm}\resizebox{\linewidth}{!}{
            \makespyww{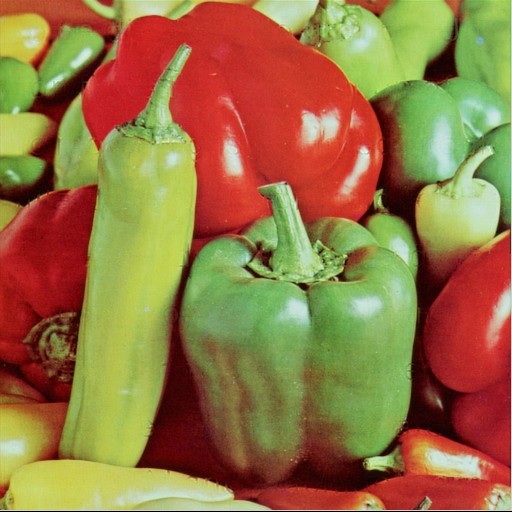}
        }
        \vspace*{-4mm}\caption{Shepard networks~\cite{RenXYS15}}
    \end{subfigure}
    \begin{subfigure}[b]{\fourlen}
        \vspace*{1mm}\resizebox{\linewidth}{!}{
            \makespyww{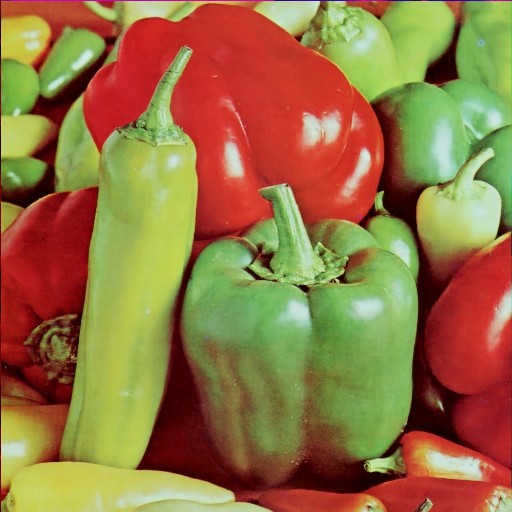}
        }
        \vspace*{-4mm}\caption{Deep Image Prior}
    \end{subfigure}
    \caption{Comparison with Shepard networks~\cite{RenXYS15} on text the inpainting task. Even though~\cite{RenXYS15} utilizes learning, the images recovered using our approach look more natural and do not have halo artifacts.}\label{fig:shepard}
\end{figure*}

\begin{figure*}
    \centering
    \begin{tabular}{@{}lccccccccccc@{}}\toprule
                              & \small{Barbara}  & \small{Boat}  & \small{House} & \small{Lena} & \small{Peppers} & \small{C.man} & \small{Couple} & \small{Finger}& \small{Hill}     & \small{Man}   & \small{Montage} \\ \midrule
        \small{Papyan et al.} & 28.14            & 31.44         & 34.58         & 35.04        & {31.11}         & 27.90          & 31.18         & 31.34                 & 32.35            & 31.92         & 28.05   \\
        \small{Ours}      & \textbf{32.22}   & \textbf{33.06}& \textbf{39.16}&\textbf{36.16}&\textbf{33.05}   &\textbf{29.8}  &\textbf{32.52} &\textbf{32.84}                      & \textbf{32.77}   & \textbf{32.20}& \textbf{34.54}  \\ \bottomrule
    \end{tabular}
    \caption{Comparison between our method and the algorithm in~\cite{PapyanRSE17}. See \cref{fig:papyan} for visual comparison.}\label{tab:papyan}\vspace*{4mm}
    \centering
    \deflen{fourlenneew}{0.245\linewidth}
    \renewcommand\makespy[1]{%

        \begin{tikzpicture}[spy using outlines={rectangle,magnification=2.7, height=3.0cm, width=5.9cm, every spy on node/.append style={line width=1.5mm}}]
                \node (nd1){\includegraphics{#1}};

                \RelativeSpy{nd1-spy1}{nd1}{(0.37,0.355)}{(0.167,-0.08)}{yellow}
                \RelativeSpy{nd1-spy2}{nd1}{(0.68,0.575)}{(0.5,-0.08)}{blue}
                \RelativeSpy{nd1-spy4}{nd1}{(0.90,0.694)}{(0.833,-0.08)}{green}
        \end{tikzpicture}
    }
    \begin{subfigure}[b]{\fourlenneew}
        \resizebox{\textwidth}{!}{
            \makespy{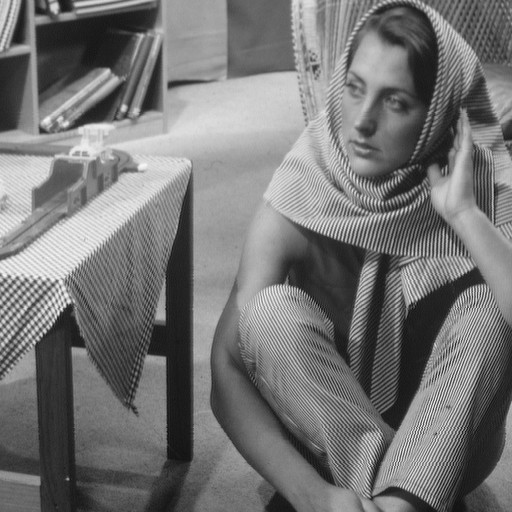}
        }
        \vspace*{-4mm}\caption{Original image}
    \end{subfigure}
    \begin{subfigure}[b]{\fourlenneew}
        \resizebox{\linewidth}{!}{
            \makespy{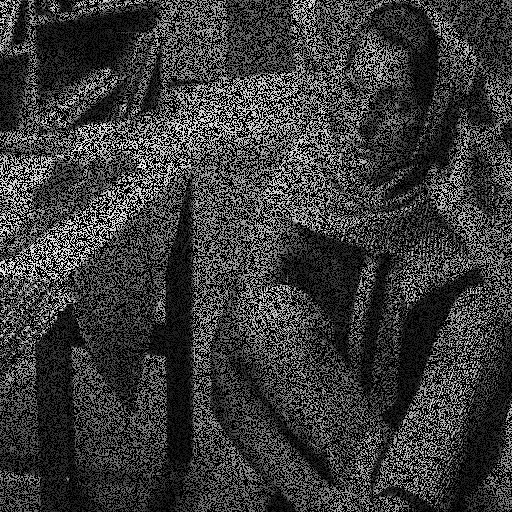}
        }
        \vspace*{-4mm}\caption{Corrupted image}
    \end{subfigure}
    \begin{subfigure}[b]{\fourlenneew}
        \resizebox{\linewidth}{!}{
            \makespy{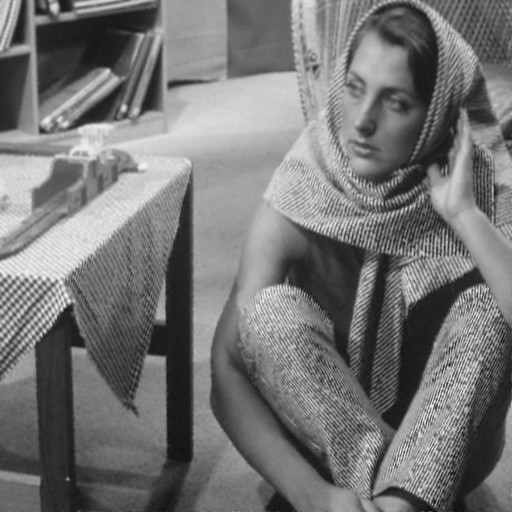}
        }
        \vspace*{-4mm}\caption{CSC~\cite{PapyanRSE17}}
    \end{subfigure}
    \begin{subfigure}[b]{\fourlenneew}
        \resizebox{\linewidth}{!}{
            \makespy{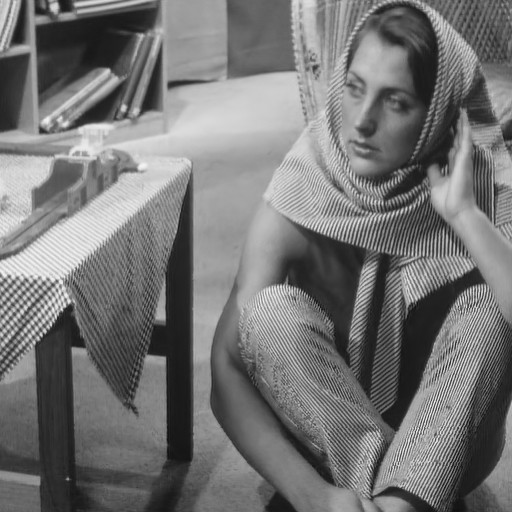}
        }
        \vspace*{-4mm}\caption{Deep image prior}
    \end{subfigure}%
\caption{Comparison with convolutional sparse coding (CSC)~\cite{PapyanRSE17} on inpainting 50\% of missing pixels. Our approach recovers a natural image with more accurate fine details than convolutional sparse coding.}\label{fig:papyan}
\end{figure*}

 \begin{figure*}
    \centering
    \deflen{mylength}{0.490\linewidth}
    \begin{subfigure}[b]{\mylength}
        \includegraphics[width=\linewidth]{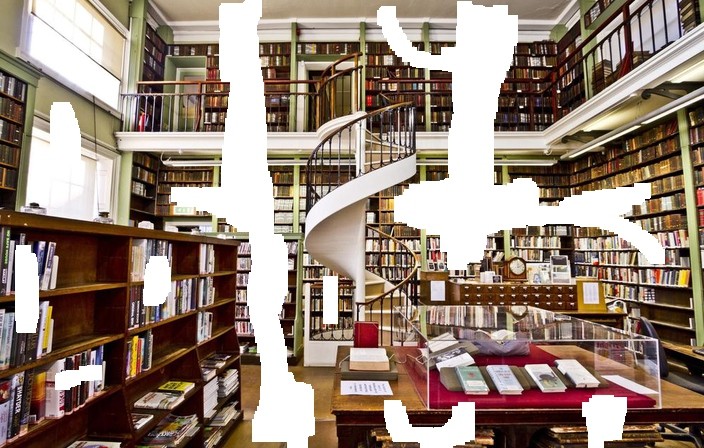}
        \caption{Input (white=masked)}
    \end{subfigure}
    \begin{subfigure}[b]{\mylength}
        \includegraphics[width=\linewidth]{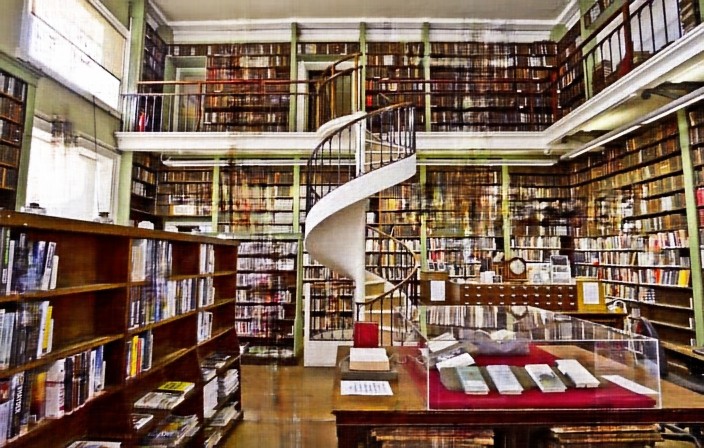}
        \caption{Encoder-decoder, depth=6}
    \end{subfigure}\\
    \begin{subfigure}[b]{\mylength}
        \includegraphics[width=\linewidth]{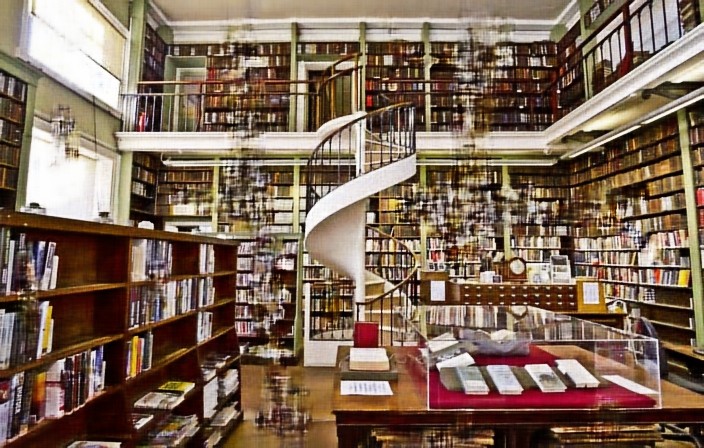}
        \caption{Encoder-decoder, depth=4}
    \end{subfigure}
    \begin{subfigure}[b]{\mylength}
        \includegraphics[width=\linewidth]{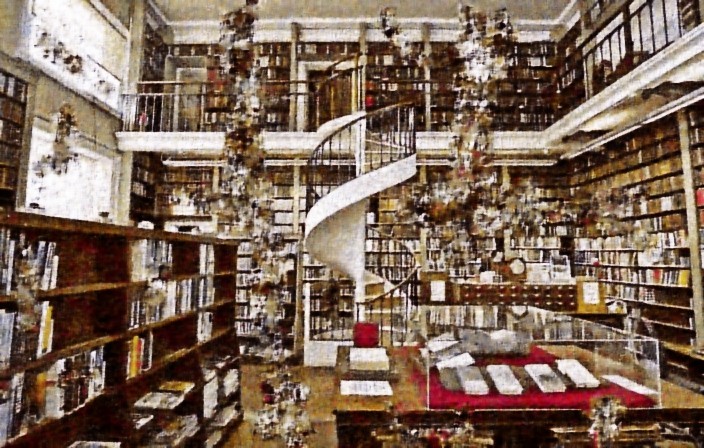}
        \caption{Encoder-decoder, depth=2}
    \end{subfigure}\\
    \begin{subfigure}[b]{\mylength}
        \includegraphics[width=\linewidth]{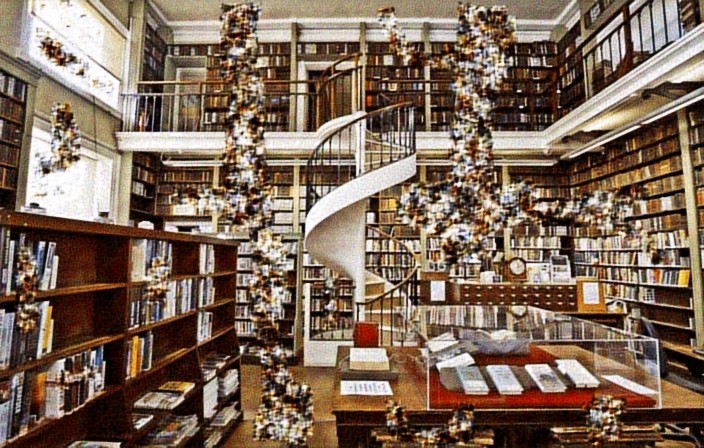}
        \caption{ResNet, depth=8}
    \end{subfigure}
    \begin{subfigure}[b]{\mylength}
        \includegraphics[width=\linewidth]{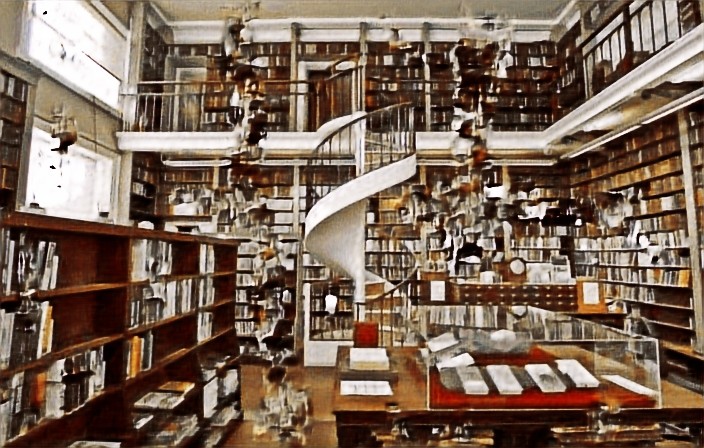}
        \caption{U-net, depth=5}
    \end{subfigure}%
\caption{\textbf{Inpainting using different depths and architectures.} The figure shows that much better inpainting results can be obtained by using deeper random networks. However, adding skip connections to ResNet in U-Net is highly detrimental for the deep image prior.}\label{fig:inpainting_comparison}
\end{figure*}

\subsection{Inpainting}

In image inpainting, one is given an image $x_0$ with missing pixels in correspondence of a binary mask $m \in \{0,1\}^{H\times W}$; the goal is to reconstruct the missing data. The corresponding data term is given by
\begin{equation}\label{eq:inpainting}
  E(x; x_0) = \| (x  - x_0) \odot m \|^2\,,
\end{equation}
where $\odot$ is Hadamard's product. The necessity of a data prior is obvious as this energy is independent of the values of the missing pixels, which would therefore never change after initialization if the objective was optimized directly over pixel values $x$. As before, the prior is introduced by optimizing the data term w.r.t.\ the re-parametrization~\eqref{eq:reparametrization}.

In the first example (\cref{fig:shepard}) inpainting is used to remove text overlaid on an image. Our approach is compared to the method of~\cite{RenXYS15} specifically designed for inpainting. Our approach leads to almost perfect results with virtually no artifacts, while for~\cite{RenXYS15} the text mask remains visible in some regions.

Next, \cref{fig:papyan} considers inpainting with masks randomly sampled according to a binary Bernoulli distribution. First, a mask is sampled to drop $50\%$ of pixels at random. We compare our approach to a method of~\cite{PapyanRSE17} based on convolutional sparse coding. To obtain results for~\cite{PapyanRSE17} we first decompose the corrupted image $x_0$ into low and high frequency components similarly to~\cite{GuZXMFZ15} and run their method on the high frequency part. For a fair comparison we use the version of their method, where a dictionary is built using the input image (shown to perform better in~\cite{PapyanRSE17}). The quantitative comparison on the standard data set~\cite{heide2015fast} for our method is given in~\cref{tab:papyan}, showing a strong quantitative advantage of the proposed approach compared to convolutional sparse coding. In~\cref{fig:papyan} we present a representative qualitative visual comparison with~\cite{PapyanRSE17}.

We also apply our method to inpainting of large holes. Being non-trainable, our method is not expected to work correctly for ``highly-semantical'' large-hole inpainting (e.g.\ face inpainting). Yet, it works surprisingly well for other situations. We compare to a learning-based method of~\cite{IizukaSIGGRAPH2017} in~\cref{fig:inpainting_region}. The deep image prior utilizes context of the image and interpolates the unknown region with textures from the known part. Such behavior highlights the relation between the deep image prior and traditional self-similarity priors.

In \cref{fig:inpainting_comparison}, we compare deep priors corresponding to several architectures. Our findings here (and in other similar comparisons) seem to suggest that having deeper architecture is beneficial, and that having skip-connections that work so well for recognition tasks (such as semantic segmentation) is highly detrimental for the deep image prior\@.

\newcommand{\setsize}[1]{\small{#1}}
\newcommand{\conv}[1]{\small{conv#1}}
\newcommand{\fc}[1]{\small{fc#1}}

\begin{figure*}
\begin{center}
\setlength{\tabcolsep}{0.05cm}
\deflen{mylinewidth}{=0.49\linewidth}
\deflen{myleninv}{=0.21\mylinewidth}
\renewcommand{\arraystretch}{0.0001}
\begin{tabular}{ccccccccc}
     \setsize{Image} & \conv1 & \conv2 & \conv3 & \conv4 & \conv5 & \fc6 & \fc7 & \fc8
        \\ 
  \raisebox{-.5\height}{\includegraphics[width=\myleninv]{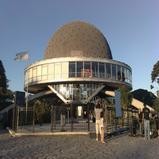}} &
  \raisebox{-.5\height}{\includegraphics[width=\myleninv]{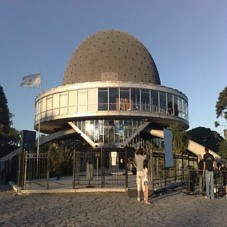}} &
  \raisebox{-.5\height}{\includegraphics[width=\myleninv]{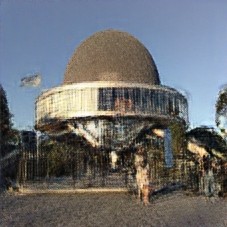}} &
  \raisebox{-.5\height}{\includegraphics[width=\myleninv]{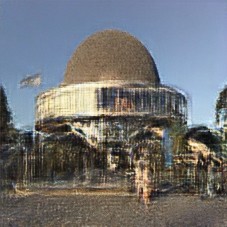}} &
  \raisebox{-.5\height}{\includegraphics[width=\myleninv]{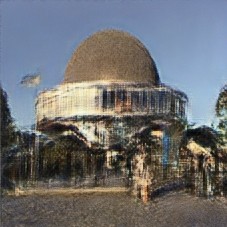}} &
  \raisebox{-.5\height}{\includegraphics[width=\myleninv]{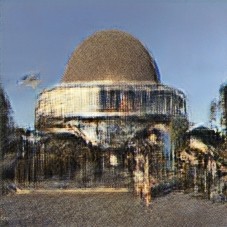}} &
  \raisebox{-.5\height}{\includegraphics[width=\myleninv]{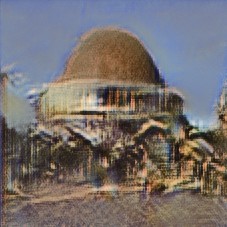}} &
  \raisebox{-.5\height}{\includegraphics[width=\myleninv]{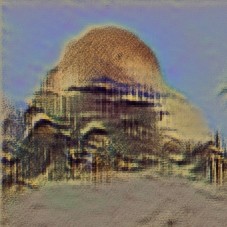}} &
  \raisebox{-.5\height}{\includegraphics[width=\myleninv]{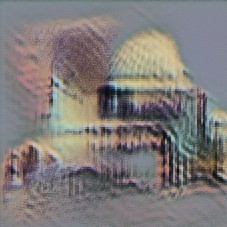}}\vspace*{0.5mm}
  \\
  \multicolumn{9}{c}{Inversion with deep image prior} \vspace*{1mm}\\
  \raisebox{-.5\height}{\includegraphics[width=\myleninv]{orig.jpg}} &
  \raisebox{-.5\height}{\includegraphics[width=\myleninv]{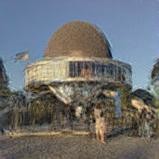}} &
  \raisebox{-.5\height}{\includegraphics[width=\myleninv]{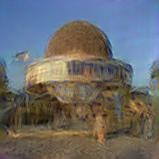}} &
  \raisebox{-.5\height}{\includegraphics[width=\myleninv]{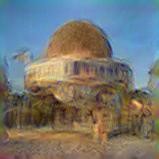}} &
  \raisebox{-.5\height}{\includegraphics[width=\myleninv]{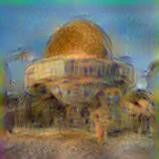}} &
  \raisebox{-.5\height}{\includegraphics[width=\myleninv]{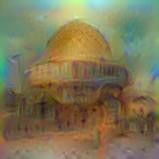}} &
  \raisebox{-.5\height}{\includegraphics[width=\myleninv]{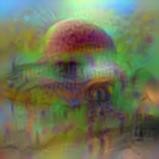}} &
  \raisebox{-.5\height}{\includegraphics[width=\myleninv]{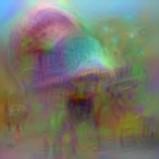}} &
  \raisebox{-.5\height}{\includegraphics[width=\myleninv]{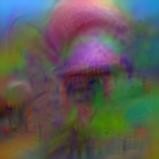}}\vspace*{0.5mm}
   \\
  \multicolumn{9}{c}{Inversion with TV prior~\cite{mahendran15understanding}}\vspace*{1mm}
  \\
  \raisebox{-.5\height}{\includegraphics[width=\myleninv]{orig.jpg}} &
  \raisebox{-.5\height}{\includegraphics[width=\myleninv]{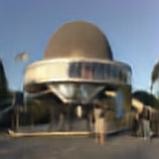}} &
  \raisebox{-.5\height}{\includegraphics[width=\myleninv]{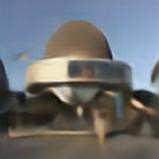}} &
  \raisebox{-.5\height}{\includegraphics[width=\myleninv]{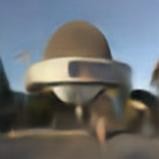}} &
  \raisebox{-.5\height}{\includegraphics[width=\myleninv]{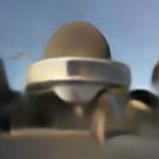}} &
  \raisebox{-.5\height}{\includegraphics[width=\myleninv]{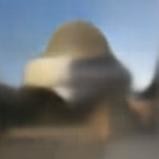}} &
  \raisebox{-.5\height}{\includegraphics[width=\myleninv]{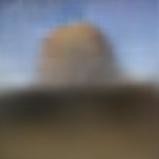}} &
  \raisebox{-.5\height}{\includegraphics[width=\myleninv]{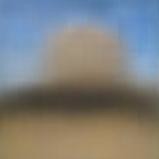}} &
  \raisebox{-.5\height}{\includegraphics[width=\myleninv]{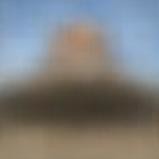}} \vspace*{0.5mm}
  \\
  \multicolumn{9}{c}{Pre-trained deep inverting network~\cite{dosovitskiy16inverting}}

\end{tabular}
\end{center}
   \caption{\textbf{AlexNet inversion.} Given the image on the left, we show the natural pre-image obtained by inverting different layers of AlexNet (trained for classification on ImageNet ILSVRC) using three different regularizers: the deep image prior, the TV norm prior of~\cite{mahendran15understanding}, and the network trained to invert representations on a hold-out set~\cite{dosovitskiy16inverting}. The reconstructions obtained with the deep image prior are in many ways at least as natural as~\cite{dosovitskiy16inverting}, yet they are not biased by the learning process.}\label{fig:inv}
\end{figure*}

\begin{figure*}
\begin{center}
\setlength{\tabcolsep}{0.05cm}
\let\mylinewidth\undefined
\let\myleninv\undefined
\deflen{mylinewidth}{=0.49\linewidth}
\deflen{myleninv}{=0.21\mylinewidth}
\renewcommand{\arraystretch}{0.0001}
 \resizebox{\textwidth}{!}{\begin{tabular}{cccccccccc}
     \setsize{Image} & \conv{1\_1} & \conv{1\_2} & \conv{2\_1} & \conv{2\_2} & \conv{3\_1} & \conv{3\_2} & \conv{3\_3} & \conv{3\_4} & \conv{4\_1}  \\ 
  \raisebox{-.5\height}{\includegraphics[width=\myleninv]{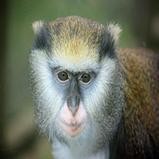}} &
  \raisebox{-.5\height}{\includegraphics[width=\myleninv]{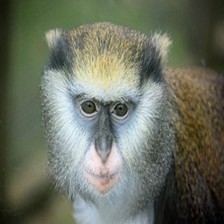}} &
  \raisebox{-.5\height}{\includegraphics[width=\myleninv]{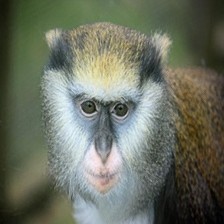}} &
  \raisebox{-.5\height}{\includegraphics[width=\myleninv]{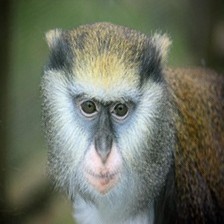}} &
  \raisebox{-.5\height}{\includegraphics[width=\myleninv]{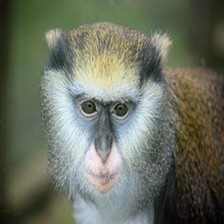}} &
  \raisebox{-.5\height}{\includegraphics[width=\myleninv]{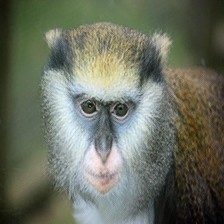}} &
  \raisebox{-.5\height}{\includegraphics[width=\myleninv]{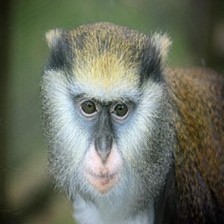}} &
  \raisebox{-.5\height}{\includegraphics[width=\myleninv]{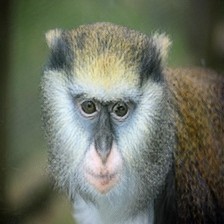}} &
  \raisebox{-.5\height}{\includegraphics[width=\myleninv]{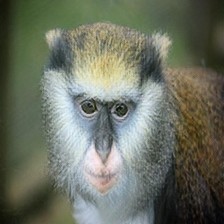}} &
  \raisebox{-.5\height}{\includegraphics[width=\myleninv]{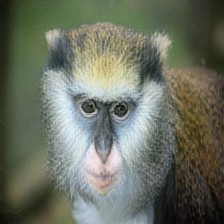}}\vspace*{0.5mm} \\
  \multicolumn{9}{c}{Inversion with deep image prior} \vspace*{1mm}\\
    \raisebox{-.5\height}{\includegraphics[width=\myleninv]{monkey.jpg}} &
  \raisebox{-.5\height}{\includegraphics[width=\myleninv]{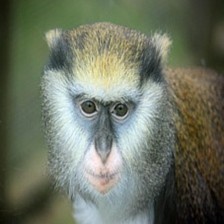}} &
  \raisebox{-.5\height}{\includegraphics[width=\myleninv]{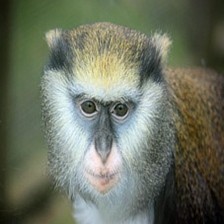}} &
  \raisebox{-.5\height}{\includegraphics[width=\myleninv]{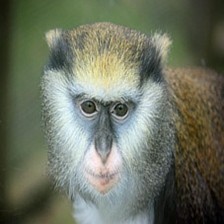}} &
  \raisebox{-.5\height}{\includegraphics[width=\myleninv]{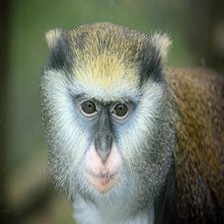}} &
  \raisebox{-.5\height}{\includegraphics[width=\myleninv]{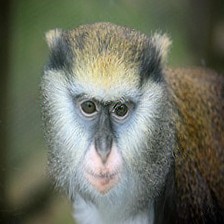}} &
  \raisebox{-.5\height}{\includegraphics[width=\myleninv]{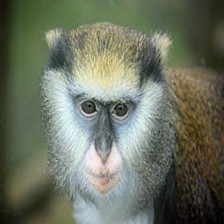}} &
  \raisebox{-.5\height}{\includegraphics[width=\myleninv]{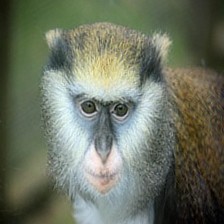}} &
  \raisebox{-.5\height}{\includegraphics[width=\myleninv]{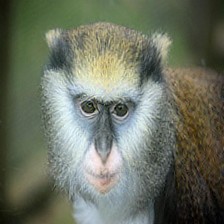}} &
  \raisebox{-.5\height}{\includegraphics[width=\myleninv]{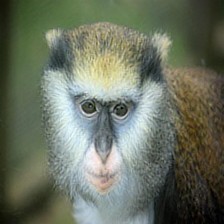}} \vspace*{0.5mm} \\
  \multicolumn{9}{c}{Inversion with TV prior~\cite{mahendran15understanding}}\vspace*{1mm}
\end{tabular}}\\ \vspace*{2.5mm}
\resizebox{\textwidth}{!}{\begin{tabular}{cccccccccc}
     \conv{4\_2} & \conv{4\_3} & \conv{4\_4} & \conv{5\_1} & \conv{5\_2} & \conv{5\_3} & \conv{5\_4} & \fc{6} & \fc{7} & \fc{8} \vspace*{0.5mm} \\
  \raisebox{-.5\height}{\includegraphics[width=\myleninv]{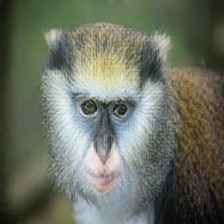}} &
  \raisebox{-.5\height}{\includegraphics[width=\myleninv]{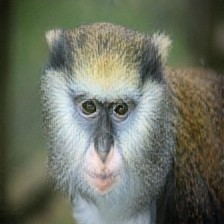}} &
  \raisebox{-.5\height}{\includegraphics[width=\myleninv]{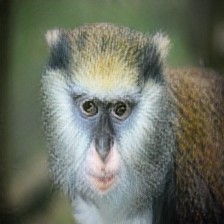}} &
  \raisebox{-.5\height}{\includegraphics[width=\myleninv]{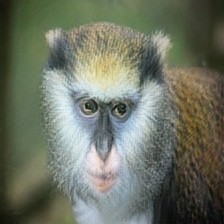}} &
  \raisebox{-.5\height}{\includegraphics[width=\myleninv]{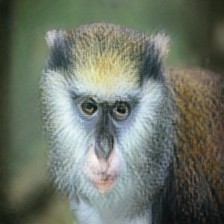}} &
  \raisebox{-.5\height}{\includegraphics[width=\myleninv]{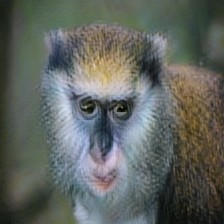}} &
  \raisebox{-.5\height}{\includegraphics[width=\myleninv]{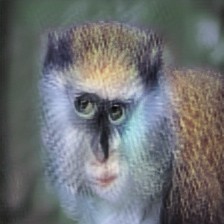}} &
  \raisebox{-.5\height}{\includegraphics[width=\myleninv]{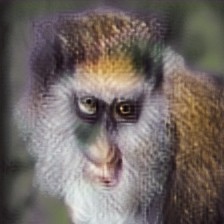}} &
  \raisebox{-.5\height}{\includegraphics[width=\myleninv]{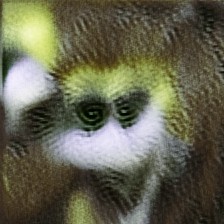}} &
  \raisebox{-.5\height}{\includegraphics[width=\myleninv]{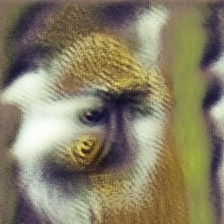}}\vspace*{0.5mm}\\
  \multicolumn{9}{c}{Inversion with deep image prior} \vspace*{1mm}\\
  \raisebox{-.5\height}{\includegraphics[width=\myleninv]{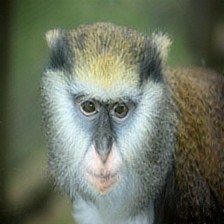}} &
  \raisebox{-.5\height}{\includegraphics[width=\myleninv]{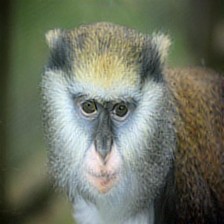}} &
  \raisebox{-.5\height}{\includegraphics[width=\myleninv]{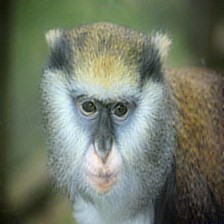}} &
  \raisebox{-.5\height}{\includegraphics[width=\myleninv]{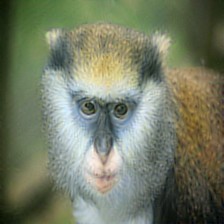}} &
  \raisebox{-.5\height}{\includegraphics[width=\myleninv]{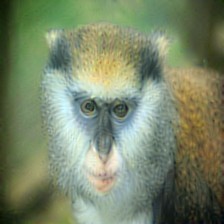}} &
  \raisebox{-.5\height}{\includegraphics[width=\myleninv]{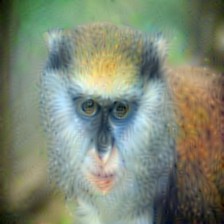}} &
  \raisebox{-.5\height}{\includegraphics[width=\myleninv]{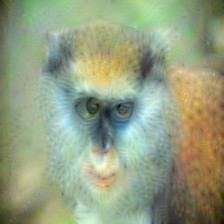}} &
  \raisebox{-.5\height}{\includegraphics[width=\myleninv]{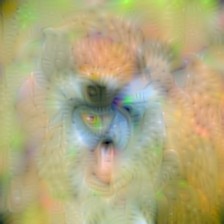}} &
  \raisebox{-.5\height}{\includegraphics[width=\myleninv]{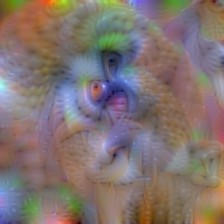}} &
  \raisebox{-.5\height}{\includegraphics[width=\myleninv]{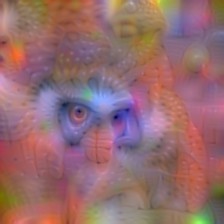}} \vspace*{0.5mm} \\
  \multicolumn{9}{c}{Inversion with TV prior~\cite{mahendran15understanding}}\vspace*{1mm}
\end{tabular}}\\
\end{center}
   \vspace*{-4mm}\caption{Inversion of VGG-19~\cite{Simonyan14c} network activations at different layers with different priors.}\label{fig:vgg_inv}
\end{figure*}
\subsection{Natural pre-image} The natural pre-image method of~\cite{mahendran15understanding} is a \emph{diagnostic} tool to study the invariances of a lossy function, such as a deep network, that operates on natural images. Let $\Phi$ be the first several layers of a neural network trained to perform, say, image classification. The pre-image is the set
\begin{equation}
\Phi^{-1}(\Phi(x_0)) = \{ x \in \mathcal{X} : \Phi(x) = \Phi(x_0)\}
\end{equation}
of images that result in the \emph{same representation} $\Phi(x_0)$. Looking at this set reveals which information is lost by the network, and which invariances are gained.

Finding pre-image points can be formulated as minimizing the data term
\begin{equation}
E(x;x_0) = \| \Phi(x) - \Phi(x_0) \|^2\,.
\end{equation}

However, optimizing this function directly may find ``artifacts'', i.e.\ non-natural images for which the behavior of the network $\Phi$ is in principle unspecified and that can thus drive it arbitrarily. More meaningful visualization can be obtained by restricting the pre-image to a set $\mathcal{X}$ of natural images, called a \emph{natural pre-image} in~\cite{mahendran15understanding}.

In practice, finding points in the natural pre-image can be done by regularizing the data term similarly to the other inverse problems seen above. The authors of~\cite{mahendran15understanding} prefer to use the TV norm, which is a weak natural image prior, but is relatively unbiased. On the contrary, papers such as~\cite{dosovitskiy16inverting} learn to invert a neural network from examples, resulting in better looking reconstructions, which however may be biased towards the learned data-driven inversion prior. Here, we propose to use the deep image prior~\eqref{eq:reparametrization} instead. As this is handcrafted like the TV-norm, it is not biased towards a particular training set. On the other hand, it results in inversions at least as interpretable as the ones of~\cite{dosovitskiy16inverting}.

For evaluation, our method is compared to the ones of~\cite{mahendran16visualizing} and~\cite{dosovitskiy16inverting}. \Cref{fig:inv} shows the results of inverting representations $\feat$ obtained by considering progressively deeper subsets of AlexNet~\cite{AlexNet}: \texttt{conv1}, \texttt{conv2}, \dots, \texttt{conv5}, \texttt{fc6}, \texttt{fc7}, and \texttt{fc8}. Pre-images are found either by optimizing~\eqref{eq:reparametrization} using a structured prior.

As seen in~\cref{fig:inv}, our method results in dramatically improved image clarity compared to the simple TV-norm. The difference is particularly remarkable for deeper layers such as \texttt{fc6} and \texttt{fc7}, where the TV norm still produces noisy images, whereas the structured regularizer produces images that are often still interpretable. Our approach also produces more informative inversions than a learned prior of~\cite{dosovitskiy16inverting}, which have a clear tendency to regress to the mean. Note that~\cite{dosovitskiy16inverting} has been followed-up by~\cite{DosovitskiyB16} where they used a learnable discriminator and a perceptual loss to train the model. While the usage of a more complex loss clearly improved their results, we do not compare to their method here as our goal is to demonstrated what can be achieved with a prior not obtained from a training set.

We perform similar experiment and invert layers of VGG-19~\cite{Simonyan14c} in~\cref{fig:vgg_inv} and also observe an improvement.

\deflen{fourlenn}{0.240\textwidth}
\renewcommand\makespy[1]{\includegraphics[width=\textwidth]{#1}}

\begin{figure*}
\begin{center}
\deflen{mylinewidthw}{\linewidth}
\deflen{myleninvw}{=0.23\mylinewidthw}
\begin{tabular}{cccc}
    Black Swan & Goose & Frog & Cheeseburger \\
      {\includegraphics[width=\myleninvw]{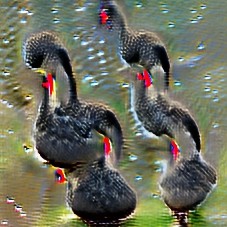}} &
      {\includegraphics[width=\myleninvw]{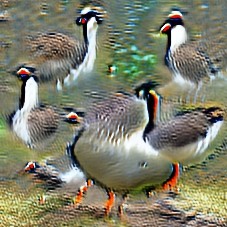}} &
      {\includegraphics[width=\myleninvw]{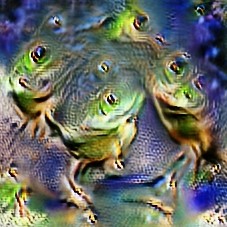}} &
      {\includegraphics[width=\myleninvw]{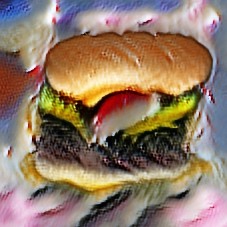}}\vspace*{-0.5mm}
      \\
      a & b & c & d \\
      \multicolumn{4}{c}{AlexNet activation maximization with Deep Image Prior}\vspace*{1mm}\\
      {\includegraphics[width=\myleninvw]{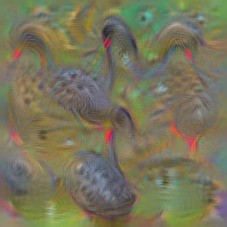}} &
      {\includegraphics[width=\myleninvw]{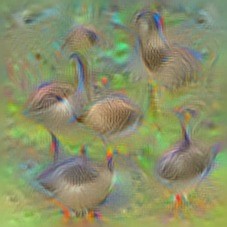}} &
      {\includegraphics[width=\myleninvw]{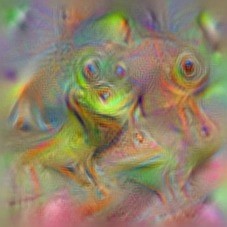}} &
      {\includegraphics[width=\myleninvw]{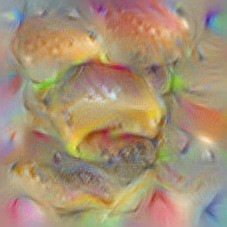}}\vspace*{-0.5mm}
       \\
      \multicolumn{4}{c}{AlexNet activation maximization with Total Variation prior~\cite{mahendran15understanding}}\vspace*{5mm}
      \\
      {\includegraphics[width=\myleninvw]{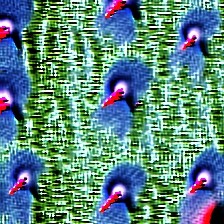}} &
      {\includegraphics[width=\myleninvw]{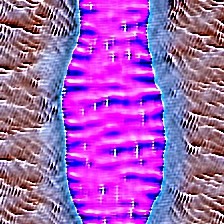}} &
      {\includegraphics[width=\myleninvw]{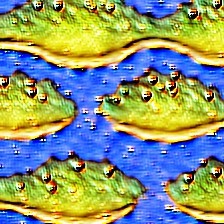}} &
      {\includegraphics[width=\myleninvw]{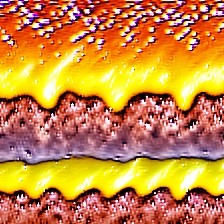}}\vspace*{-0.5mm}
      \\
      \multicolumn{4}{c}{VGG-16 activation maximization with Deep Image Prior} \vspace*{1mm}
      \\
      {\includegraphics[width=\myleninvw]{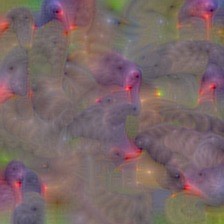}} &
      {\includegraphics[width=\myleninvw]{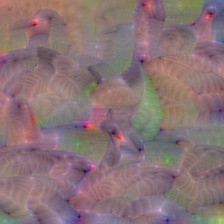}} &
      {\includegraphics[width=\myleninvw]{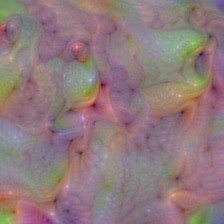}} &
      {\includegraphics[width=\myleninvw]{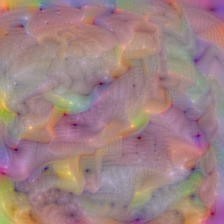}}\vspace*{-0.5mm}
       \\
      \multicolumn{4}{c}{VGG-16 activation maximization with Total Variation prior~\cite{mahendran15understanding}}\vspace*{3mm}
\end{tabular}%
\end{center}
\caption{\textbf{Class activation maximization.} For a given class label shown at the very top, we show images obtained by maximizing the corresponding class activation (before soft-max) of AlexNet (top) and VGG-16 (bottom) architectures using different regularizers: the deep image prior proposed here (rows 1 and 3), and the total variation prior of~\cite{rudin1992tv}. For both architectures (AlexNet) in particular, inversion with deep image prior leads to more interpretable results.}\label{fig:actmax}
\end{figure*}

\begin{figure*}
\begin{center}
\setlength{\tabcolsep}{2pt}
\deflen{myleninvww}{=0.17\linewidth}
\resizebox{\textwidth}{!}{\begin{tabular}{cccccc}
{\includegraphics[width=\myleninvww]{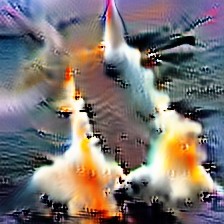}}  &
{\includegraphics[width=\myleninvww]{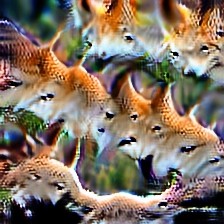}}  &
{\includegraphics[width=\myleninvww]{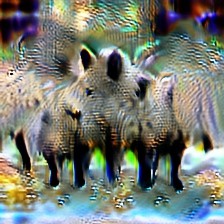}}  &
{\includegraphics[width=\myleninvww]{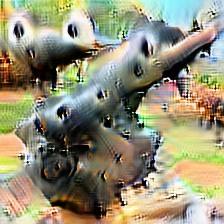}}  &
{\includegraphics[width=\myleninvww]{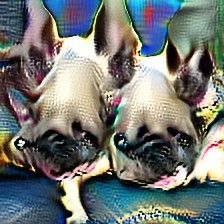}}  &
{\includegraphics[width=\myleninvww]{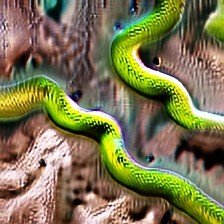}}  \\
 missile & dingo & wild boar & cannon & French bulldog & green snake  \\
{\includegraphics[width=\myleninvww]{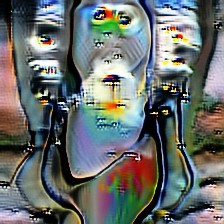}}  &
{\includegraphics[width=\myleninvww]{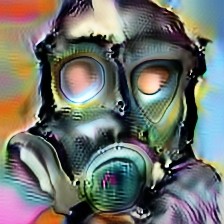}}  &
{\includegraphics[width=\myleninvww]{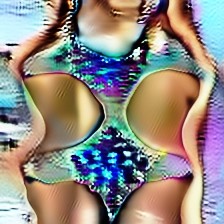}}  &
{\includegraphics[width=\myleninvww]{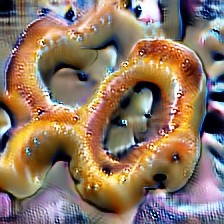}}  &
{\includegraphics[width=\myleninvww]{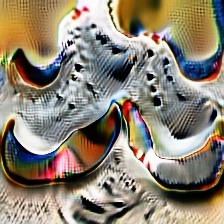}}  &
{\includegraphics[width=\myleninvww]{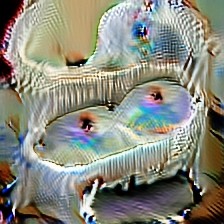}}  \\
 gas pump & mask & maillot & pretzel & running shoe & bassinet  \\
{\includegraphics[width=\myleninvww]{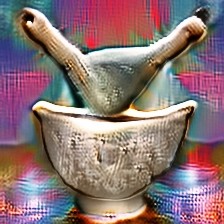}}  &
{\includegraphics[width=\myleninvww]{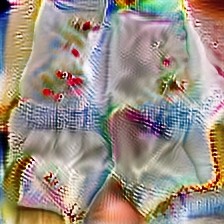}}  &
{\includegraphics[width=\myleninvww]{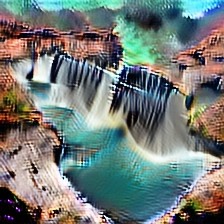}}  &
{\includegraphics[width=\myleninvww]{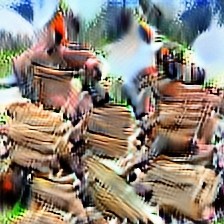}}  &
{\includegraphics[width=\myleninvww]{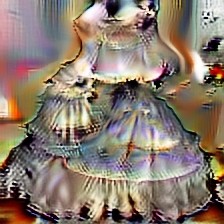}}  &
{\includegraphics[width=\myleninvww]{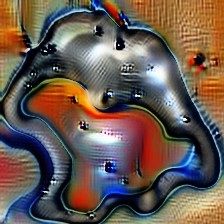}}  \\
 mortar & handkerchief & dam & lumbermill & hoopskirt & vacuum  \\
{\includegraphics[width=\myleninvww]{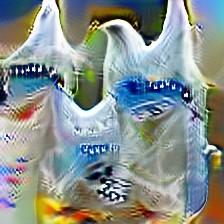}}  &
{\includegraphics[width=\myleninvww]{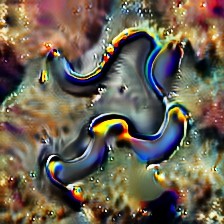}}  &
{\includegraphics[width=\myleninvww]{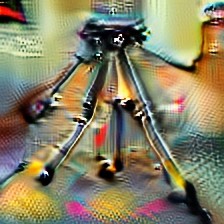}}  &
{\includegraphics[width=\myleninvww]{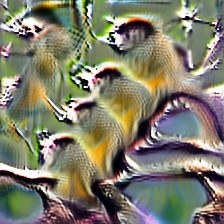}}  &
{\includegraphics[width=\myleninvww]{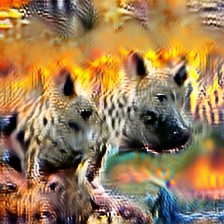}}  &
{\includegraphics[width=\myleninvww]{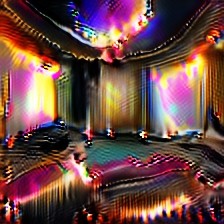}}  \\
 plastic bag & flatworm & tripod & spider monkey & hyena & cinema  \\
{\includegraphics[width=\myleninvww]{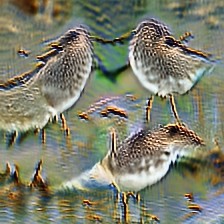}}  &
{\includegraphics[width=\myleninvww]{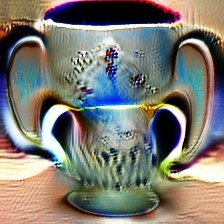}}  &
{\includegraphics[width=\myleninvww]{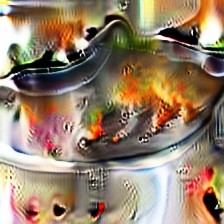}}  &
{\includegraphics[width=\myleninvww]{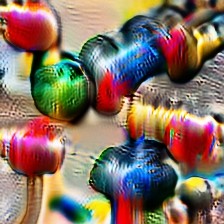}}  &
{\includegraphics[width=\myleninvww]{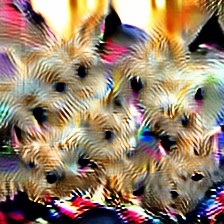}}  &
{\includegraphics[width=\myleninvww]{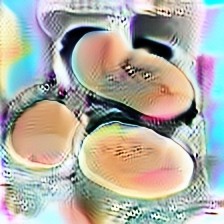}}  \\
 dowitcher & coffee mug & Crock Pot & abacus & Norwich terrier & face powder  \\
{\includegraphics[width=\myleninvww]{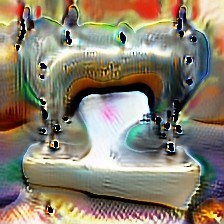}}  &
{\includegraphics[width=\myleninvww]{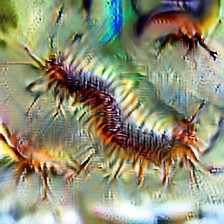}}  &
{\includegraphics[width=\myleninvww]{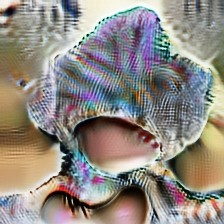}}  &
{\includegraphics[width=\myleninvww]{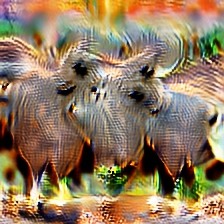}}  &
{\includegraphics[width=\myleninvww]{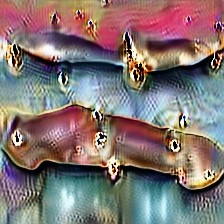}}  &
{\includegraphics[width=\myleninvww]{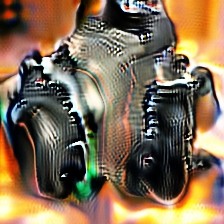}}  \\
 sewing machine & centipede & bonnet & warthog & scabbard & reflex camera  \\
{\includegraphics[width=\myleninvww]{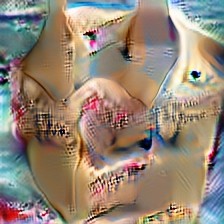}}  &
{\includegraphics[width=\myleninvww]{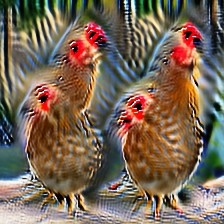}}  &
{\includegraphics[width=\myleninvww]{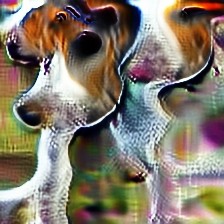}}  &
{\includegraphics[width=\myleninvww]{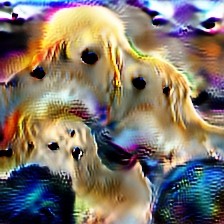}}  &
{\includegraphics[width=\myleninvww]{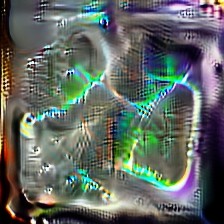}}  &
{\includegraphics[width=\myleninvww]{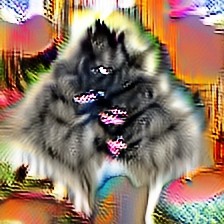}}  \\
 carton & hen & English foxhound & golden retriever & oscilloscope & keeshond
\end{tabular}}%
\end{center}
\caption{AlexNet activation maximization regularized with deep image prior for different randomly-selected ILSVRC class labels.}\label{fig:actmax_more}
\end{figure*}

\subsection{Activation maximization}

Along with the pre-image method, the \textit{activation maximization} method is used to visualize internals of a deep neural network. It aims to synthesize an image that highly activates a certain neuron by solving the following optimization problem:
\begin{equation}
x^* = \arg \max_x \Phi(x)_m\,,
\end{equation}
where $m$ is an index of a chosen neuron. $\Phi(x)_m$ corresponds to $m$-th output if $\Phi$ ends with fully-connected layer and central pixel of the $m$-th feature map if the $\Phi(x)$ has spatial dimensions.

We compare the proposed deep prior to TV prior from~\cite{mahendran15understanding} in~\cref{fig:actmax}, where we aim to maximize activations of the last \texttt{fc8} layer of AlexNet and VGG-16. For AlexNet deep image prior leads to more natural and interpretable images, while the effect is not as clear in the case of VGG-16. In~\cref{fig:actmax_more} we show more examples, where we maximize the activation for a certain class.

\subsection{Image enhancement}
\deflen{mylengthoop}{0.193\textwidth}
\begin{figure*}
    \centering
    \begin{subfigure}[h]{\mylengthoop}
        \includegraphics[width=\linewidth]{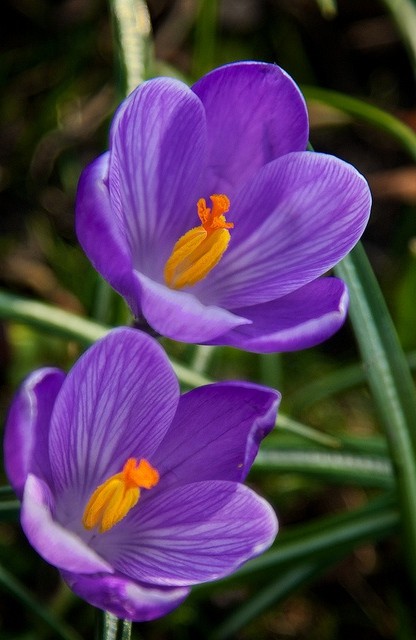}
    \end{subfigure}
    \begin{subfigure}[h]{\mylengthoop}
        \includegraphics[width=\linewidth]{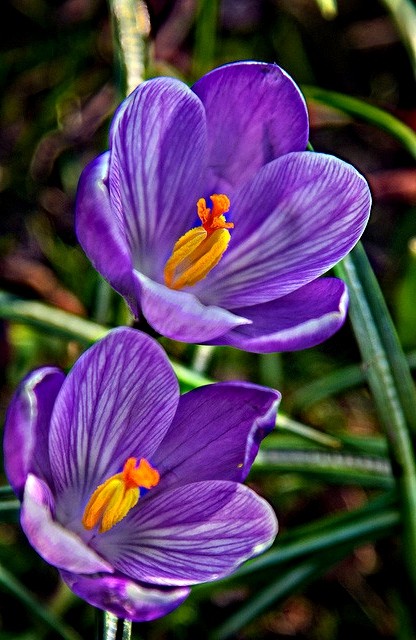}
    \end{subfigure}
    \begin{subfigure}[h]{\mylengthoop}
        \includegraphics[width=\linewidth]{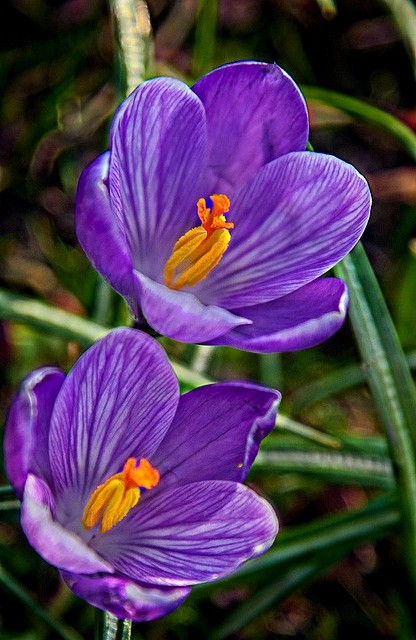}
    \end{subfigure}
    \begin{subfigure}[h]{\mylengthoop}
        \includegraphics[width=\linewidth]{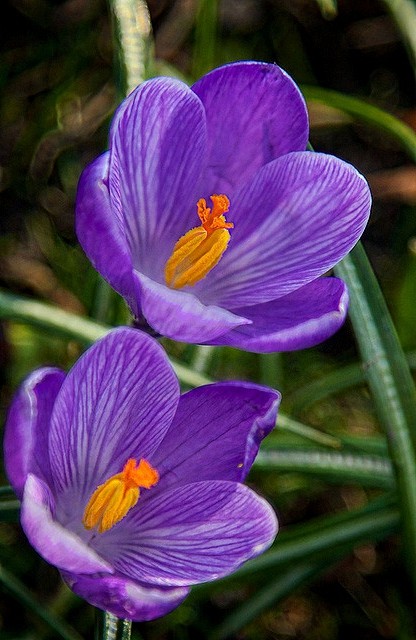}
    \end{subfigure}
    \begin{subfigure}[h]{\mylengthoop}
        \includegraphics[width=\linewidth]{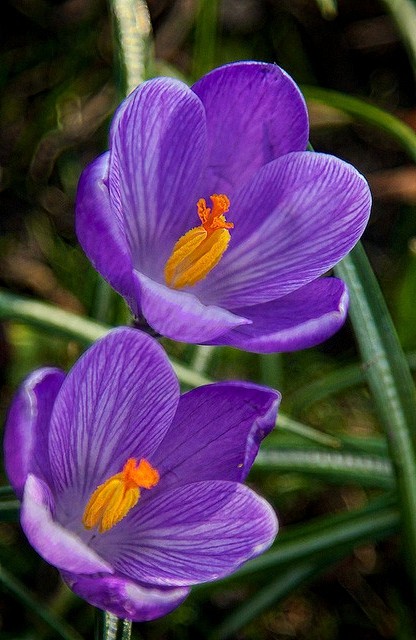}
    \end{subfigure}\\
    \begin{subfigure}[h]{\mylengthoop}
        \includegraphics[width=\linewidth]{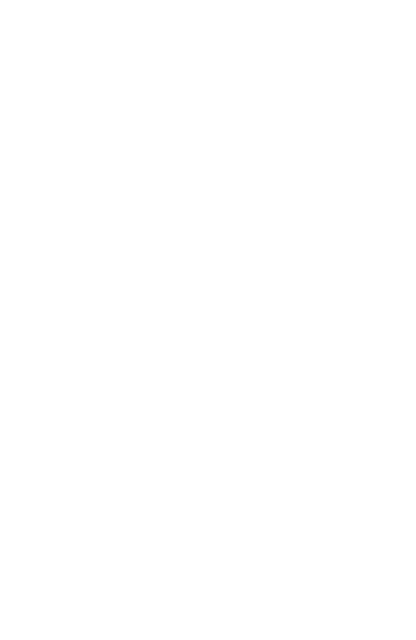}
    \end{subfigure}
    \begin{subfigure}[h]{\mylengthoop}
        \includegraphics[width=\linewidth]{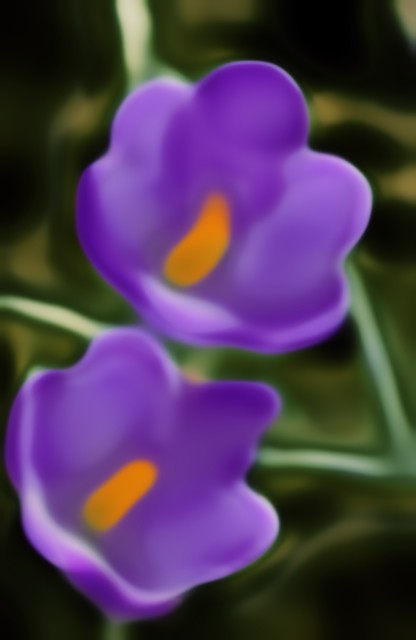}
        \caption{Iteration 250}
    \end{subfigure}
    \begin{subfigure}[h]{\mylengthoop}
        \includegraphics[width=\linewidth]{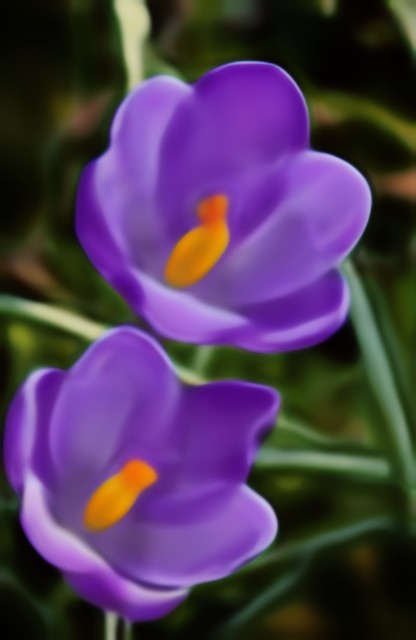}
        \caption{Iteration 500}
    \end{subfigure}
    \begin{subfigure}[h]{\mylengthoop}
        \includegraphics[width=\linewidth]{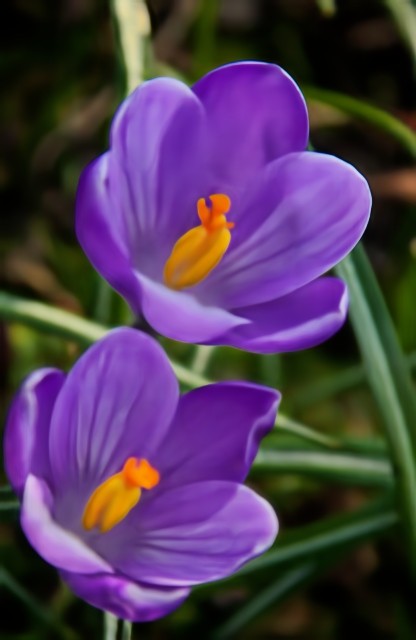}
        \caption{Iteration 2500}
    \end{subfigure}
    \begin{subfigure}[h]{\mylengthoop}
        \includegraphics[width=\linewidth]{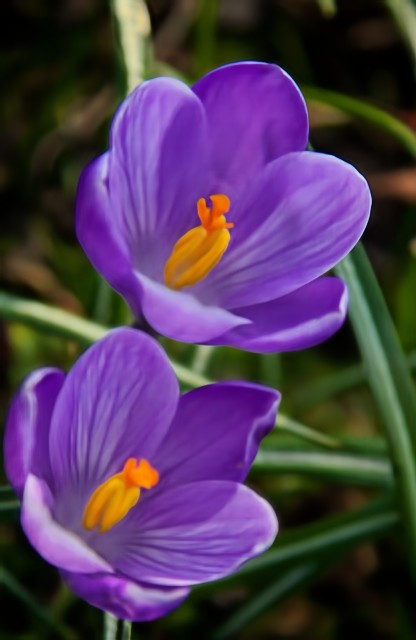}
        \caption{Iteration 7000}
    \end{subfigure}
    \caption{\textbf{Coarse and boosted images for different stopping points.} We obtain the coarse images (second row) running the optimization for reconstruction objective~\ref{eq:denoise2} for a certain number of iterations. We then subtract coarse version from the original image to get fine details and boost them (first row). Even for low iteration number the coarse approximation preserves edges for the large objects. The original image in shown the first column. }\label{fig:enchancement_process}
\end{figure*}

We also use the proposed deep image regularization to perform high frequency enhancement in an image. As demonstrated in~\cref{s:noise_impedance}, the noisy image is reconstructed starting from coarse low-frequency details and finishing with fine high frequency details and noise. To perform enhancement we use the objective~\eqref{eq:denoise2} setting the target image to be $x_0$. We stop the optimization process at a certain point, obtaining a coarse approximation $x_c$ of the image $x_0$. The fine details are then computed as
\begin{equation}
x_f = x_0 - x_c \,.
\end{equation}
We then construct an enhanced image by boosting the extracted fine details $x_f$:
\begin{equation}
x_e = x_0 + x_f \,.
\end{equation}

In~\cref{fig:enchancement_process} we present coarse and enhanced versions of the same image, running the optimization process for different number of iterations. At the start of the optimization process (corresponds to low number of iteration) the resulted approximation does not precisely recreates the shape of the objects (c.f.\ blue halo in the bottom row of ~\cref{fig:enchancement_process}). While the shapes become well-matched with the time, unwanted high frequency details also start to appear. Thus we need to stop the optimization process in time.

\begin{figure*}
    \centering
    \deflen{fourlennn}{0.245\linewidth}
    \renewcommand\makespy[1]{%
        \begin{tikzpicture}[spy using outlines={rectangle,magnification=2.7, height=4.05cm, width=8.9cm, every spy on node/.append style={line width=2mm}}]
                \node (nd1){\includegraphics{#1}};
                \RelativeSpy{nd1-spy1}{nd1}{(0.37,0.680)}{(0.167,-0.08)}{yellow}
                \RelativeSpy{nd1-spy4}{nd1}{(0.58,0.67)}{(0.5,-0.08)}{blue}
                \RelativeSpy{nd1-spy2}{nd1}{(0.78,0.695)}{(0.833,-0.08)}{green}
        \end{tikzpicture}
    }
    \begin{subfigure}[b]{\fourlennn}
        \resizebox{\linewidth}{!}{
            \makespy{{./cave01_flash.jpg}.jpg}
        }
        \caption{Flash}
    \end{subfigure}
    \begin{subfigure}[b]{\fourlennn}
        \resizebox{\linewidth}{!}{
            \makespy{{./cave01_noflash.jpg}.jpg}
        }
        \caption{No flash}
    \end{subfigure}
    \begin{subfigure}[b]{\fourlennn}
        \resizebox{\linewidth}{!}{
            \makespy{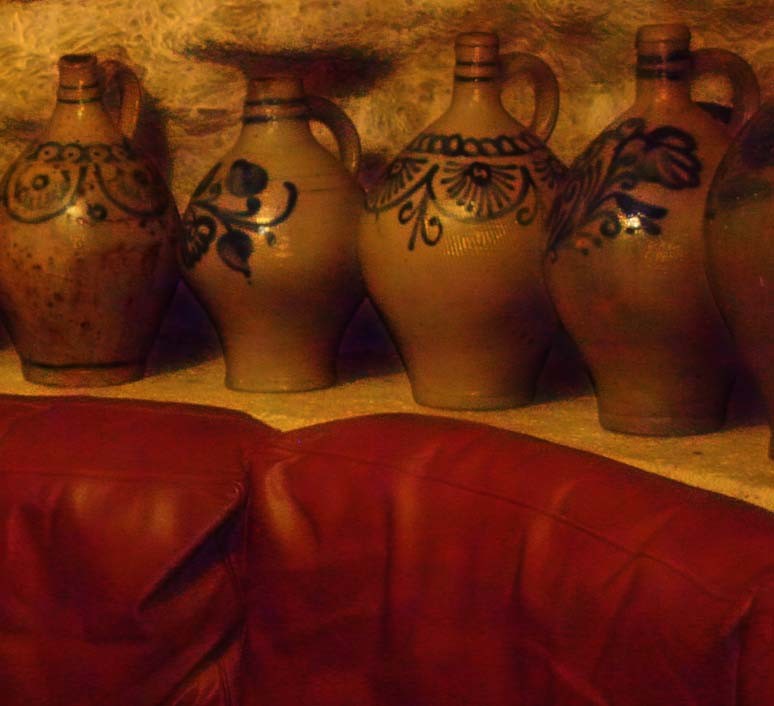}
        }
        \caption{Joint bilateral~\cite{PetschniggSACHT04}}
    \end{subfigure}
    \begin{subfigure}[b]{\fourlennn}
        \resizebox{\linewidth}{!}{
            \makespy{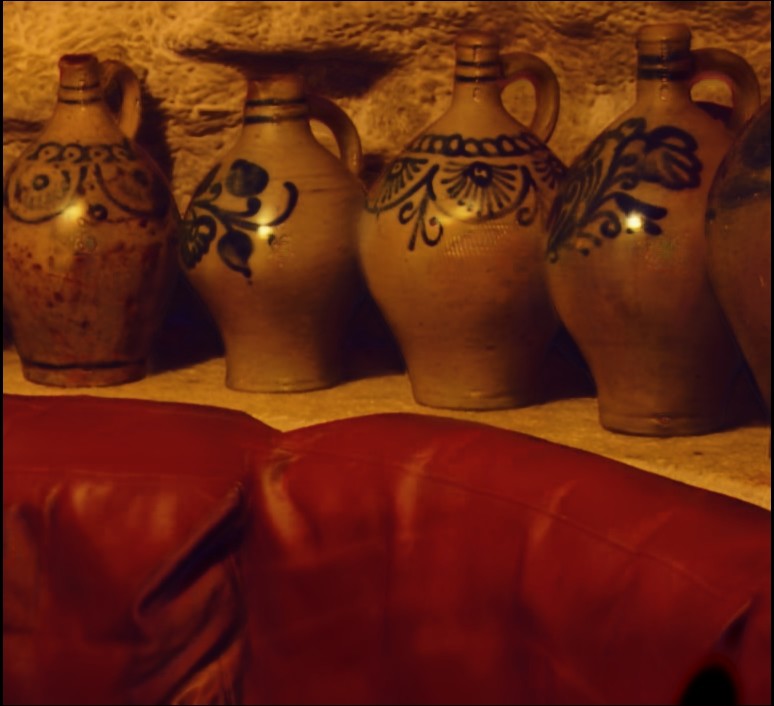}
        }
        \caption{Deep image prior}
    \end{subfigure}
    \caption{\textbf{Reconstruction based on flash and no-flash image pair.} The deep image prior allows to obtain low-noise reconstruction with the lighting very close to the no-flash image. It is more successful at avoiding ``leaks'' of the lighting patterns from the flash pair than joint bilateral filtering~\cite{PetschniggSACHT04} (c.f.~blue inset).} \label{fig:flash}
\end{figure*}

\subsection{Flash-no flash reconstruction}

While in this work we focus on single image restoration, the proposed approach can be extended to the tasks of the restoration of multiple images, e.g.\ for the task of video restoration. We therefore conclude the set of application examples with a qualitative example demonstrating how the method can be applied to perform restoration based on pairs of images. In particular, we consider flash-no flash image pair-based restoration~\cite{PetschniggSACHT04}, where the goal is to obtain an image of a scene with the lighting similar to a no-flash image, while using the flash image as a guide to reduce the noise level.

In general, extending the method to more than one image is likely to involve some coordinated optimization over the input codes $z$ that for single-image tasks in our approach was most often kept fixed and random. In the case of flash-no-flash restoration, we found that good restorations were obtained by using the denoising formulation \eqref{eq:denoise2}, while using flash image as an input (in place of the random vector $z$). The resulting approach can be seen as a non-linear generalization of guided image filtering~\cite{he2013guided}. The results of the restoration are given in the \cref{fig:flash}.

\section{Technical details}\label{s:tech_details}

\begin{figure*}
    \begin{subfigure}[b]{0.46\linewidth}
        \includegraphics[width=\linewidth]{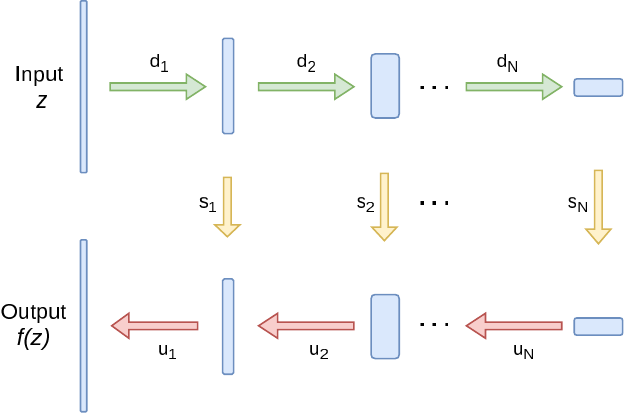}
    \end{subfigure}
    \begin{subfigure}[b]{0.46\linewidth}\hspace*{15mm}
        \includegraphics[width=\linewidth]{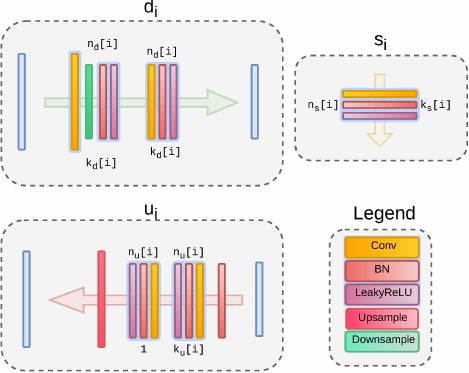}
    \end{subfigure}
    \caption{\textbf{The architecture used in the experiments.} We use ``hourglass'' (also known as ``decoder-encoder'') architecture. We sometimes add skip connections (yellow arrows). $n_u[i]$, $n_d[i]$, $n_s[i]$ correspond to the number of filters at depth $i$ for the upsampling, downsampling and skip-connections respectively. The values $k_u[i]$, $k_d[i]$, $k_s[i]$ correspond to the respective kernel sizes.}\label{fig:arch}
\end{figure*}

\begin{figure}
\includegraphics[width=\linewidth]{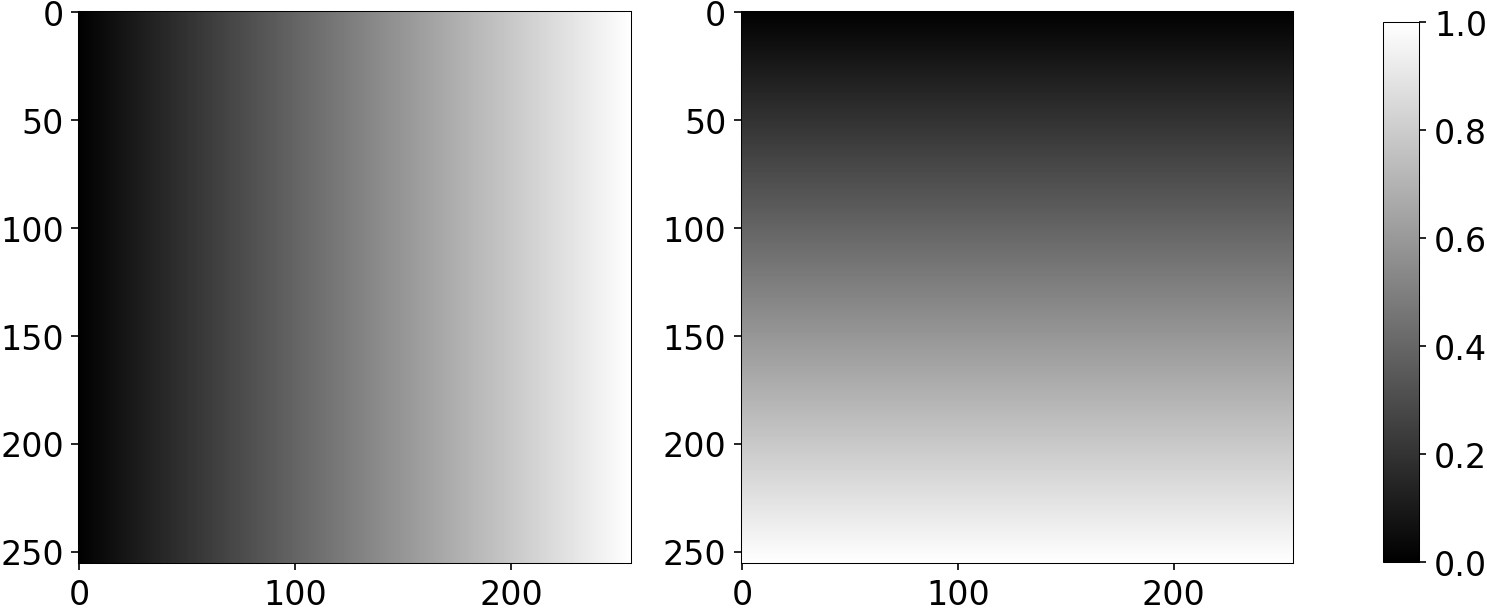}
\caption{\textbf{``Meshgrid'' input} $z$ used in some inpainting experiments. These are two channels of the input tensor; in BCHW layout: \texttt{z[0, 0, :, :], z[0, 1, :, :]} The intensity encodes the value: from zero (black) to one (white). Such type of input can be regarded as a part of the prior which enforces smoothness.}\label{fig:meshgrid}
\end{figure}

While other options are possible, we mainly experimented with fully-convolutional architectures, where the input $z \in \mathbb{R}^{C' \times W \times H}$ has the same spatial resolution as the the output of the network $f_\theta(z) \in \mathbb{R}^{3 \times W \times H}$.

We use encoder-decoder (``hourglass'') architecture  (possibly with skip-connections) for $f_\theta$ in all our experiments except noted otherwise (\cref{fig:arch}), varying a small number of hyper-parameters. Although the best results can be achieved by carefully tuning an architecture for a particular task (and potentially for a particular image), we found  that wide range of hyper-parameters and architectures give acceptable results.

We use LeakyReLU~\cite{he2015delving} as a non-linearity. As a downsampling technique we simply use strides implemented within convolution modules. We also tried average/max pooling and downsampling with Lanczos kernel, but did not find a consistent difference between any of them. As an upsampling operation we choose between bilinear upsampling and nearest neighbor upsampling. An alternative upsampling method could be to use transposed convolutions, but the results we obtained using them were worse. We use reflection padding instead of zero padding in convolution layers everywhere except for the feature inversion and activation maximization experiments.

We considered two ways to create the input $z$: 1. \textit{random}, where the $z$ is filled with uniform noise between zero and $0.1$, 2. \textit{meshgrid}, where we initialize $z \in \mathbb{R}^{2 \times W \times H}$ using \texttt{np.meshgrid} (see~\cref{fig:meshgrid}). Such initialization serves as an additional smoothness prior to the one imposed by the structure of $f_\theta$ itself. We found such input to be beneficial for large-hole inpainting, but not for other tasks.

During fitting of the networks we often use a \textit{noise-based regularization}. I.e.\ at each iteration we perturb the input $z$ with an additive normal noise with zero mean and standard deviation $\sigma_p$. While we have found such regularization to impede optimization process, we also observed that the network was able to eventually optimize its objective to zero no matter the variance of the additive noise (i.e.\ the network was always able to adapt to any reasonable variance for sufficiently large number of optimization steps).

We found the optimization process tends to destabilize as the loss goes down and approaches a certain value. Destabilization is observed as a significant loss increase and blur in generated image $f_{\theta}(z)$. From such destabilization point the loss goes down again till destabilized one more time. To remedy this issue we simply track the optimization loss and return to parameters from the previous iteration if the loss difference between two consecutive iterations is higher than a certain threshold.

Finally, we use \textit{ADAM} optimizer~\cite{Kingma14adam} in all our experiments and PyTorch as a framework. The proposed iterative optimization requires repeated forward and backward evaluation of a deep ConvNet and thus takes several minutes per image.

Below, we provide the remaining details of the network architectures. We use the notation introduced in~\cref{fig:arch}.

\medskip\noindent\textbf{Super-resolution (\textit{default architecture}).}
\\
\noindent\fbox{%
\parbox{\linewidth}{%
$z \in \mathbb{R}^{32 \times W \times H} \sim U(0,\frac{1}{10})$\\
$n_u = n_d = $ \texttt{[128, 128, 128, 128, 128]}\\
$k_u = k_d = $ \texttt{[3, 3, 3, 3, 3]}\\
$n_s = $ \texttt{[4, 4, 4, 4, 4]}\\
$k_s = $ \texttt{[1, 1, 1, 1, 1]}\\
$\sigma_p = \frac{1}{30}$\\
\texttt{num\_iter} $ = 2000$\\
\texttt{LR} $ = 0.01$\\
\texttt{upsampling} $ = $ \texttt{bilinear}
}}
\\
The decimation operator $d$ is composed of low pass filtering operation using Lanczos2 kernel (see~\cite{turkowski1990filters}) and resampling, all implemented as a single (fixed) convolutional layer.

For \texttt{8$\times$} super-resolution (\cref{fig:sr}) we have changed the standard deviation of the input noise to $\sigma_p = \frac{1}{20}$ and the number of iterations to $4000$.

\medskip\noindent\textbf{Text inpainting (\cref{fig:shepard}).} We used the same hyper-parameters as for super-resolution but optimized the objective for $6000$ iterations.

\medskip\noindent\textbf{Large hole inpainting (\cref{fig:inpainting_region}).} We used the same hyper-parameters as for super-resolution, but used \texttt{meshgrid} as an input, removed skip connections and optimized for $5000$ iterations.

\medskip\noindent\textbf{Large hole inpainting (\cref{fig:inpainting_comparison}).}
We used the following hyper-parameters:
\\
\noindent\fbox{%
\parbox{\linewidth}{%
$z \in \mathbb{R}^{32 \times W \times H} \sim U(0,\frac{1}{10})$\\
$n_u = n_d = $ \texttt{[16, 32, 64, 128, 128, 128]}\\
$k_d = $ \texttt{[3, 3, 3, 3, 3, 3]}\\
$k_u = $ \texttt{[5, 5, 5, 5, 5, 5]}\\
$n_s = $ \texttt{[0, 0, 0, 0, 0, 0]}\\
$k_s = $ \texttt{[NA, NA, NA, NA, NA, NA]}\\
$\sigma_p = 0$\\
\texttt{num\_iter = 5000}\\
\texttt{LR = 0.1}\\
\texttt{upsampling} $ = $ \texttt{nearest}
}}
\\
In\cref{fig:inpainting_comparison} (c), (d) we simply sliced off last layers to get smaller depth.

\medskip\noindent\textbf{Denoising (\cref{fig:denoising}).} Hyper-parameters were set to be the same as in the case of super-resolution with only difference in iteration number, which was set to $1800$.
We used the following implementations of referenced denoising methods: \cite{BM3Dcode} for CBM3D and~\cite{NLMcode} for NLM\@.
We used exponential sliding window with weight $\gamma=0.99$.

\medskip\noindent\textbf{JPEG artifacts removal (\cref{fig:jpeg}).}
Although we could use the same setup as in other denoising experiments, the hyper-parameters we used to generate the image in \cref{fig:jpeg} were the following:
\\
\noindent\fbox{%
\parbox{\linewidth}{%
$z \in \mathbb{R}^{3 \times W \times H} \sim U(0,\frac{1}{10})$\\
$n_u = n_d = $ \texttt{[8, 16, 32, 64, 128]}\\
$k_u = k_d = $ \texttt{[3, 3, 3, 3, 3]}\\
$n_s = $ \texttt{[0, 0, 0, 4, 4]}\\
$k_s = $ \texttt{[NA, NA, NA, 1, 1]}\\
$\sigma_p = \frac{1}{30}$\\
\texttt{num\_iter = 2400}\\
\texttt{LR = 0.01}\\
\texttt{upsampling} $ = $ \texttt{bilinear}
}}
\\

\medskip\noindent\textbf{Image reconstruction (\cref{fig:papyan}).} We used the same setup as in the case of super-resolution and denoising, but set \texttt{num\_iter = 11000}, \texttt{LR = 0.001}.

\medskip\noindent\textbf{Natural pre-image (\cref{fig:inv,fig:vgg_inv}).}
\\
\noindent\fbox{%
\parbox{\linewidth}{%
$z \in \mathbb{R}^{32 \times W \times H} \sim U(0,\frac{1}{10})$\\
$n_u = n_d = $ \texttt{[16, 32, 64, 128, 128, 128]}\\
$k_u = k_d = $ \texttt{[7, 7, 5, 5, 3, 3]}\\
$n_s = $ \texttt{[4, 4, 4, 4, 4]}\\
$k_s = $ \texttt{[1, 1, 1, 1, 1]}\\
\texttt{num\_iter = 3100}\\
\texttt{LR = 0.001}\\
\texttt{upsampling} $ = $ \texttt{nearest}
}}
\\
We used \texttt{num\_iter = 10000} for the VGG inversion experiment (\cref{fig:vgg_inv})

\medskip\noindent\textbf{Activation maximization (\cref{fig:actmax,fig:actmax_more}).}
In this experiment we used a very similar set of hyper-parameters to the ones in pre-image experiment.
\\
\noindent\fbox{%
\parbox{\linewidth}{%
$z \in \mathbb{R}^{32 \times W \times H} \sim U(0,\frac{1}{10})$\\
$n_u = n_d = $ \texttt{[16, 32, 64, 128, 128, 128]}\\
$k_u = k_d = $ \texttt{[5, 3, 5, 5, 3, 5]}\\
$n_s = $ \texttt{[0, 4, 4, 4, 4]}\\
$k_s = $ \texttt{[1, 1, 1, 1, 1]}\\
\texttt{num\_iter = 3100}\\
\texttt{LR = 0.001}\\
\texttt{upsampling} $ = $ \texttt{bilinear}\\
$\sigma_p = 0.03 $
}}

\medskip\noindent\textbf{Image enhancement (\cref{fig:enchancement_process}).} We used the same setup as in the case of super-resolution and denoising, but set $\sigma_p = 0$. 
\section{Related work}\label{s:related}

Our approach is related to image restoration and synthesis methods based on learnable ConvNets and referenced above. Here, we review other lines of work related to our approach.

Modelling ``translation-invariant'' statistics of natural images using filter responses has a very long history of research. The statistics of responses to various non-random filters (such as simple operators and higher-order wavelets) have been studied in seminal works~\cite{Field87,Mallat89,Simoncelli96,Zhu97}. Later, \cite{huang2000statistics} noted that image response distribution w.r.t.~random unlearned filters have very similar properties to the distributions of wavelet filter responses.

Our approach is closely related to a group of restoration methods that avoid training on the hold-out set and exploit the well-studied self-similarity properties of natural images~\cite{Ruderman94,Turiel98}. This group includes methods based on joint modeling of groups of similar patches inside corrupted image~\cite{buades2005non,dabov2007image,glasner2009super}, which are particularly useful when the corruption process is complex and highly variable (e.g.\ spatially-varying blur~\cite{bahat2017non}).

In this group, an interesting parallel work with clear links to our approach is the zero-shot super-resolution approach~\cite{Shocher18}, which trains a feed-forward super-resolution ConvNet based on synthetic dataset generated from the patches of a single image. While clearly related, the approach~\cite{Shocher18} is somewhat complementary as it exploit self-similarities across multiple scales of the same image, while our approach exploits self-similarities within the same scale (at multiple scales).

Several lines of work use dataset-based learning and modeling images using convolutional operations. Learning priors for natural images that facilitate restoration by enforcing filter responses for certain (learned) filters is behind an influential field-of-experts model~\cite{Roth09}. Also in this group are methods based on fitting dictionaries to the patches of the corrupted  image~\cite{mairal2010online,set14} as well as methods based on convolutional sparse coding~\cite{Grosse07,Bristow13}. The connections between convolutional sparse coding and ConvNets are investigated in~\cite{Papyan17jmlr} in the context of recognition tasks. More recently in~\cite{PapyanRSE17}, a fast single-layer convolutional sparse coding is proposed for reconstruction tasks. The comparison of our approach with~\cite{PapyanRSE17} (\cref{fig:shepard,tab:papyan}) however suggests that using deep ConvNet architectures popular in modern deep learning-based approaches may lead to more accurate restoration results.

Deeper convolutional models of natural images trained on large datasets have also been studied extensively. E.g.~deconvolutional networks~\cite{zeiler2010deconvolutional} are trained by fitting hierarchies of representations linked by convolutional operators to datasets of natural images. The recent work~\cite{lefkimmiatis2016non} investigates the model that combines ConvNet with a self-similarity based denoising and thus bridges learning on image datasets and exploiting within-image self-similarities.

Our approach is also related to inverse scale space denoising~\cite{Scherzer01,Burger05,Marquina09}. In this group of ``non-deep'' image processing methods, a sequence of solutions (a flow) that gradually progresses from a uniform image to the noisy image, while progressively finer scale details are recovered so that early stopping yields a denoised image. The inverse scale space approaches are however still driven by a simple total variation (TV) prior, which does not model self-similarity of images, and limits the ability to denoise parts of images with textures and gradual transitions. Note that our approach can also use the simple stopping criterion proposed in \cite{Burger05}, when the level of noise is known.

Finally, we note that this manuscript expands the conference version~\cite{Ulyanov18} in multiple ways: 1) It gives more intuition, provides more visualizations and explanation for the presented method altogether with extensive technical details. 2) It contains a more thorough experimental evaluation and shows an application to activation maximization and high frequency enhancement. Since the publication of the preliminary version of our approach, it has also been used by other groups in different ways. Thus,~\cite{Veen18CompressedSensing} proposes a novel method for compressed sensing recovery using deep image prior. The work~\cite{Athar18LCM} learns a latent variable model, where the latent space is parametrized by a convolutional neural network. The approach~\cite{Shedligeri18} aims to reconstruct an image from an event-based camera and utilizes deep image prior framework to estimate sensor's ego-motion. The method~\cite{Ilyas17} successively applies deep image prior to defend against adversarial attacks. Deep image prior is also used in~\cite{Boominathan18} to perform phase retrieval for Fourier ptychography.

\section{Discussion}\label{s:conc}

We have investigated the success of recent image generator neural networks, teasing apart the contribution of the prior imposed by the choice of architecture from the contribution of the information transferred from external images through learning. In particular, we have shown that fitting a randomly-initialized ConvNet to corrupted images works as a ``Swiss knife'' for restoration problems. This approach is probably too slow to be useful for most practical applications, and for each particular application, a feed-forward network trained for that particular application would do a better job and do so much faster. Thus, the slowness and the inability to match or exceed the results of problem specific methods are the two main limitations of our approach, when practical applications are considered. While of limited practicality, the good results of our approach across a wide variety of tasks demonstrate that an implicit prior inside deep convolutional network architectures is an important part of the success of such architectures for image restoration tasks. 

Why does this prior emerge, and, more importantly, why does it fit the structure of natural images so well? We speculate that generation by convolutional operations naturally tends impose self-similarity of the generated images (c.f.~\cref{fig:samples}), as convolutional filters are applied across the entire visual field thus imposing certain stationarity on the output of convolutional layers. Hourglass architectures with skip connections naturally impose self-similarity at multiple scales, making the corresponding priors suitable for the restoration of natural images.

We note that our results go partially against the common narrative that explain the success of deep learning in image restoration (and beyond) by the ability to learn rather than by hand-craft priors; instead, we show that properly \textit{hand-crafted} network architectures correspond to better \emph{hand-crafted} priors, and it seems that learning ConvNets builds on this basis. This observation also validates the importance of developing new deep learning architectures.


\paragraph{Acknowledgements.} DU and VL are supported by the Ministry of Education and Science of the Russian Federation (grant 14.756.31.0001) and AV is supported by ERC 638009-IDIU\@.

{\footnotesize\bibliographystyle{spmpsci}\bibliography{refs}}

\begin{thebibliography}{10}
\providecommand{\url}[1]{{#1}}
\providecommand{\urlprefix}{URL }
\expandafter\ifx\csname urlstyle\endcsname\relax
  \providecommand{\doi}[1]{DOI~\discretionary{}{}{}#1}\else
  \providecommand{\doi}{DOI~\discretionary{}{}{}\begingroup
  \urlstyle{rm}\Url}\fi

\bibitem{Shocher18}
Assaf~Shocher Nadav~Cohen, M.I.: "zero-shot" super-resolution using deep
  internal learning.
\newblock In: {CVPR}. {IEEE} Computer Society (2018)

\bibitem{Athar18LCM}
Athar, S., Burnaev, E., Lempitsky, V.S.: Latent convolutional models.
\newblock CoRR  (2018)

\bibitem{bahat2017non}
Bahat, Y., Efrat, N., Irani, M.: Non-uniform blind deblurring by reblurring.
\newblock In: Proc. {CVPR}, pp. 3286--3294. {IEEE} Computer Society (2017)

\bibitem{set5}
Bevilacqua, M., Roumy, A., Guillemot, C., Alberi{-}Morel, M.: Low-complexity
  single-image super-resolution based on nonnegative neighbor embedding.
\newblock In: {BMVC}, pp. 1--10 (2012)

\bibitem{Bojanowski17}
Bojanowski, P., Joulin, A., Lopez{-}Paz, D., Szlam, A.: Optimizing the latent
  space of generative networks.
\newblock CoRR  (2017)

\bibitem{Boominathan18}
Boominathan, L., Maniparambil, M., Gupta, H., Baburajan, R., Mitra, K.: Phase
  retrieval for fourier ptychography under varying amount of measurements.
\newblock CoRR  (2018)

\bibitem{Bristow13}
Bristow, H., Eriksson, A.P., Lucey, S.: Fast convolutional sparse coding.
\newblock In: {CVPR}, pp. 391--398. {IEEE} Computer Society (2013)

\bibitem{NLMcode}
Buades, A.: {NLM demo}.
\newblock \url{http://demo.ipol.im/demo/bcm_non_local_means_denoising/}

\bibitem{buades2005non}
Buades, A., Coll, B., Morel, J.M.: A non-local algorithm for image denoising.
\newblock In: Proc. {CVPR}, vol.~2, pp. 60--65. {IEEE} Computer Society (2005)

\bibitem{burger2012image}
Burger, H.C., Schuler, C.J., Harmeling, S.: Image denoising: Can plain neural
  networks compete with bm3d?
\newblock In: {CVPR}, pp. 2392--2399 (2012)

\bibitem{Burger05}
Burger, M., Osher, S.J., Xu, J., Gilboa, G.: Nonlinear inverse scale space
  methods for image restoration.
\newblock In: Variational, Geometric, and Level Set Methods in Computer Vision,
  Third International Workshop, {VLSM}, pp. 25--36 (2005)

\bibitem{dabov2007image}
Dabov, K., Foi, A., Katkovnik, V., Egiazarian, K.: Image denoising by sparse
  3-d transform-domain collaborative filtering.
\newblock IEEE Transactions on image processing \textbf{16}(8), 2080--2095
  (2007)

\bibitem{dong2014learning}
Dong, C., Loy, C.C., He, K., Tang, X.: Learning a deep convolutional network
  for image super-resolution.
\newblock In: Proc. {ECCV}, pp. 184--199 (2014)

\bibitem{DosovitskiyB16}
Dosovitskiy, A., Brox, T.: Generating images with perceptual similarity metrics
  based on deep networks.
\newblock In: {NIPS}, pp. 658--666 (2016)

\bibitem{dosovitskiy16inverting}
Dosovitskiy, A., Brox, T.: Inverting convolutional networks with convolutional
  networks.
\newblock In: {CVPR}. {IEEE} Computer Society (2016)

\bibitem{dosovitskiy2015learning}
Dosovitskiy, A., Tobias~Springenberg, J., Brox, T.: Learning to generate chairs
  with convolutional neural networks.
\newblock In: Proc. {CVPR}, pp. 1538--1546 (2015)

\bibitem{erhan09visualizing}
Erhan, D., Bengio, Y., Courville, A., Vincent, P.: Visualizing higher-layer
  features of a deep network.
\newblock Tech. Rep. Technical Report 1341, University of Montreal (2009)

\bibitem{Field87}
Field, D.J.: Relations between the statistics of natural images and the
  response properties of cortical cells.
\newblock Josa a \textbf{4}(12), 2379--2394 (1987)

\bibitem{glasner2009super}
Glasner, D., Bagon, S., Irani, M.: Super-resolution from a single image.
\newblock In: Proc. {ICCV}, pp. 349--356 (2009)

\bibitem{goodfellow2014generative}
Goodfellow, I., Pouget-Abadie, J., Mirza, M., Xu, B., Warde-Farley, D., Ozair,
  S., Courville, A., Bengio, Y.: Generative adversarial nets.
\newblock In: Proc. {NIPS}, pp. 2672--2680 (2014)

\bibitem{Grosse07}
Grosse, R.B., Raina, R., Kwong, H., Ng, A.Y.: Shift-invariance sparse coding
  for audio classification.
\newblock In: {UAI}, pp. 149--158. {AUAI} Press (2007)

\bibitem{GuZXMFZ15}
Gu, S., Zuo, W., Xie, Q., Meng, D., Feng, X., Zhang, L.: Convolutional sparse
  coding for image super-resolution.
\newblock In: {ICCV}, pp. 1823--1831. {IEEE} Computer Society (2015)

\bibitem{he2013guided}
He, K., Sun, J., Tang, X.: Guided image filtering.
\newblock {T-PAMI} \textbf{35}(6), 1397--1409 (2013)

\bibitem{he2015delving}
He, K., Zhang, X., Ren, S., Sun, J.: Delving deep into rectifiers: Surpassing
  human-level performance on imagenet classification.
\newblock In: {CVPR}, pp. 1026--1034. {IEEE} Computer Society (2015)

\bibitem{heide2015fast}
Heide, F., Heidrich, W., Wetzstein, G.: Fast and flexible convolutional sparse
  coding.
\newblock In: {CVPR}, pp. 5135--5143. {IEEE} Computer Society (2015)

\bibitem{huang2000statistics}
Huang, J., Mumford, D.: Statistics of natural images and models.
\newblock In: {CVPR}, pp. 1541--1547. {IEEE} Computer Society (1999)

\bibitem{Huang15sr}
Huang, J.B., Singh, A., Ahuja, N.: Single image super-resolution from
  transformed self-exemplars.
\newblock In: {CVPR}, pp. 5197--5206. {IEEE} Computer Society (2015)

\bibitem{IizukaSIGGRAPH2017}
Iizuka, S., Simo-Serra, E., Ishikawa, H.: {Globally and Locally Consistent
  Image Completion}.
\newblock ACM Transactions on Graphics (Proc. of SIGGRAPH) \textbf{36}(4),
  107:1--107:14 (2017)

\bibitem{Ilyas17}
Ilyas, A., Jalal, A., Asteri, E., Daskalakis, C., Dimakis, A.G.: The robust
  manifold defense: Adversarial training using generative models.
\newblock CoRR  (2017)

\bibitem{Kim16sr}
Kim, J., Lee, J.K., Lee, K.M.: Accurate image super-resolution using very deep
  convolutional networks.
\newblock In: {CVPR}, pp. 1646--1654. {IEEE} Computer Society (2016)

\bibitem{Kingma14adam}
Kingma, D.P., Ba, J.: Adam: {A} method for stochastic optimization.
\newblock CoRR  (2014)

\bibitem{kingma2013auto}
Kingma, D.P., Welling, M.: Auto-encoding variational bayes.
\newblock In: Proc. {ICLR} (2014)

\bibitem{AlexNet}
Krizhevsky, A., Sutskever, I., Hinton, G.E.: Imagenet classification with deep
  convolutional neural networks.
\newblock In: F.~Pereira, C.J.C. Burges, L.~Bottou, K.Q. Weinberger (eds.)
  Advances in Neural Information Processing Systems 25, pp. 1097--1105. Curran
  Associates, Inc. (2012)

\bibitem{Lai17sr}
Lai, W.S., Huang, J.B., Ahuja, N., Yang, M.H.: Deep laplacian pyramid networks
  for fast and accurate super-resolution.
\newblock In: {CVPR}. {IEEE} Computer Society (2017)

\bibitem{BM3Dcode}
Lebrun, M.: {BM3D code}.
\newblock \url{https://github.com/gfacciol/bm3d}

\bibitem{Ledig17sr}
Ledig, C., Theis, L., Huszar, F., Caballero, J., Cunningham, A., Acosta, A.,
  Aitken, A., Tejani, A., Totz, J., Wang, Z., Shi, W.: Photo-realistic single
  image super-resolution using a generative adversarial network.
\newblock In: {CVPR}. {IEEE} Computer Society (2017)

\bibitem{lefkimmiatis2016non}
Lefkimmiatis, S.: Non-local color image denoising with convolutional neural
  networks.
\newblock In: {CVPR}. {IEEE} Computer Society (2016)

\bibitem{mahendran15understanding}
Mahendran, A., Vedaldi, A.: Understanding deep image representations by
  inverting them.
\newblock In: {CVPR}. {IEEE} Computer Society (2015)

\bibitem{mahendran16visualizing}
Mahendran, A., Vedaldi, A.: Visualizing deep convolutional neural networks
  using natural pre-images.
\newblock {IJCV}  (2016)

\bibitem{mairal2010online}
Mairal, J., Bach, F., Ponce, J., Sapiro, G.: Online learning for matrix
  factorization and sparse coding.
\newblock Journal of Machine Learning Research \textbf{11}(Jan), 19--60 (2010)

\bibitem{Mallat89}
Mallat, S.G.: A theory for multiresolution signal decomposition: the wavelet
  representation.
\newblock {PAMI} \textbf{11}(7), 674--693 (1989)

\bibitem{Marquina09}
Marquina, A.: Nonlinear inverse scale space methods for total variation blind
  deconvolution.
\newblock {SIAM} J. Imaging Sciences \textbf{2}(1), 64--83 (2009)

\bibitem{Papyan17jmlr}
Papyan, V., Romano, Y., Elad, M.: Convolutional neural networks analyzed via
  convolutional sparse coding.
\newblock Journal of Machine Learning Research \textbf{18}(83), 1--52 (2017)

\bibitem{PapyanRSE17}
Papyan, V., Romano, Y., Sulam, J., Elad, M.: Convolutional dictionary learning
  via local processing.
\newblock In: {ICCV}. {IEEE} Computer Society (2017)

\bibitem{PetschniggSACHT04}
Petschnigg, G., Szeliski, R., Agrawala, M., Cohen, M.F., Hoppe, H., Toyama, K.:
  Digital photography with flash and no-flash image pairs.
\newblock {ACM} Trans. Graph. \textbf{23}(3), 664--672 (2004)

\bibitem{plotz2017benchmarking}
Plotz, T., Roth, S.: Benchmarking denoising algorithms with real photographs.
\newblock In: Proceedings of the IEEE Conference on Computer Vision and Pattern
  Recognition, pp. 1586--1595 (2017)

\bibitem{RenXYS15}
Ren, J.S.J., Xu, L., Yan, Q., Sun, W.: Shepard convolutional neural networks.
\newblock In: {NIPS}, pp. 901--909 (2015)

\bibitem{Roth09}
Roth, S., Black, M.J.: Fields of experts.
\newblock {CVPR} \textbf{82}(2), 205--229 (2009)

\bibitem{Ruderman94}
Ruderman, D.L., Bialek, W.: Statistics of natural images: Scaling in the woods.
\newblock In: {NIPS}, pp. 551--558. Morgan Kaufmann (1993)

\bibitem{rudin1992tv}
Rudin, L.I., Osher, S., Fatemi, E.: Nonlinear total variation based noise
  removal algorithms.
\newblock In: Proceedings of the Eleventh Annual International Conference of
  the Center for Nonlinear Studies on Experimental Mathematics : Computational
  Issues in Nonlinear Science: Computational Issues in Nonlinear Science, pp.
  259--268. Elsevier North-Holland, Inc., New York, NY, USA (1992)

\bibitem{Sajjadi17sr}
Sajjadi, M.S.M., Scholkopf, B., Hirsch, M.: Enhancenet: Single image
  super-resolution through automated texture synthesis.
\newblock In: The IEEE International Conference on Computer Vision (ICCV)
  (2017)

\bibitem{Scherzer01}
Scherzer, O., Groetsch, C.W.: Inverse scale space theory for inverse problems.
\newblock In: Scale-Space and Morphology in Computer Vision, Third
  International Conference, pp. 317--325 (2001)

\bibitem{Shedligeri18}
Shedligeri, P.A., Shah, K., Kumar, D., Mitra, K.: Photorealistic image
  reconstruction from hybrid intensity and event based sensor.
\newblock CoRR  (2018)

\bibitem{Simoncelli96}
Simoncelli, E.P., Adelson, E.H.: Noise removal via bayesian wavelet coring.
\newblock In: {ICIP} {(1)}, pp. 379--382. {IEEE} Computer Society (1996)

\bibitem{Simonyan14c}
Simonyan, K., Zisserman, A.: Very deep convolutional networks for large-scale
  image recognition.
\newblock CoRR  (2014)

\bibitem{Tai17sr}
Tai, Y., Yang, J., Liu, X.: Image super-resolution via deep recursive residual
  network.
\newblock In: {CVPR}. {IEEE} Computer Society (2017)

\bibitem{Turiel98}
Turiel, A., Mato, G., Parga, N., Nadal, J.: Self-similarity properties of
  natural images.
\newblock In: {NIPS}, pp. 836--842. The {MIT} Press (1997)

\bibitem{turkowski1990filters}
Turkowski, K.: Filters for common resampling-tasks.
\newblock Graphics gems pp. 147--165 (1990)

\bibitem{Ulyanov18}
Ulyanov, D., Vedaldi, A., Lempitsky, V.: Deep image prior.
\newblock In: {CVPR}. {IEEE} Computer Society (2018)

\bibitem{Veen18CompressedSensing}
Veen, D.V., Jalal, A., Price, E., Vishwanath, S., Dimakis, A.G.: Compressed
  sensing with deep image prior and learned regularization.
\newblock CoRR  (2018)

\bibitem{zeiler2010deconvolutional}
Zeiler, M.D., Krishnan, D., Taylor, G.W., Fergus, R.: Deconvolutional networks.
\newblock In: Proc. {CVPR}, pp. 2528--2535. {IEEE} Computer Society (2010)

\bibitem{set14}
Zeyde, R., Elad, M., Protter, M.: On single image scale-up using
  sparse-representations.
\newblock In: Curves and Surfaces, \emph{Lecture Notes in Computer Science},
  vol. 6920, pp. 711--730. Springer (2010)

\bibitem{zhang16understanding}
Zhang, C., Bengio, S., Hardt, M., Recht, B., Vinyals, O.: Understanding deep
  learning requires rethinking generalization.
\newblock In: {ICLR} (2017)

\bibitem{Zhu97}
Zhu, S.C., Mumford, D.: Prior learning and gibbs reaction-diffusion.
\newblock {IEEE} Trans. Pattern Anal. Mach. Intell. \textbf{19}(11), 1236--1250
  (1997)

\end{thebibliography}
\end{document}